%% file: main.tex
\documentclass[10pt,twocolumn,letterpaper]{article}

\usepackage{iccv}
\usepackage{times}
\usepackage{epsfig}
\usepackage{graphicx}
\usepackage{amsmath}
\usepackage{amssymb}
\usepackage{booktabs}
\usepackage{bbm}
\usepackage{caption}
\usepackage{subfig}
\usepackage{array}
\usepackage{tabularx}

\usepackage{hyperref}
\hypersetup{
    colorlinks=true,
    linkcolor=blue,
    urlcolor=blue,
    citecolor=cyan,
}
\iccvfinalcopy %

\usepackage[capitalize]{cleveref}
\crefname{section}{Sec.}{Secs.}
\Crefname{section}{Section}{Sections}
\Crefname{table}{Table}{Tables}
\crefname{table}{Tab.}{Tabs.}

\input{tex/shortcuts}

\ificcvfinal\fi

\begin{document}

\title{UrbanGIRAFFE: Representing Urban Scenes as Compositional \\ Generative Neural Feature Fields}

\author{
Yuanbo Yang,
~Yifei Yang,
~Hanlei Guo,
~Rong Xiong,
~Yue Wang,
~Yiyi Liao\thanks{Corresponding author.}
\vspace{0.8em}\\
Zhejiang University
}
\maketitle

\input{tex/sec_0_abstract}
\input{tex/sec_1_intro}
\input{tex/sec_2_related}
\input{tex/sec_3_method}
\input{tex/sec_4_results}

\input{tex/sec_5_conclusion}

{\small
\bibliographystyle{ieee_fullname}
\bibliography{bibliography_long,bibliography_custom,bibliography}
}

\newpage
\onecolumn
\begin{center}
\Large{\textbf{Supplementary Material for UrbanGIRAFFE: Representing Urban Scenes as  \\ Compositional Generative Neural Feature Fields}}
\end{center}
\appendix
\input{tex/sec_6_supp}

\end{document}

%% file: tex/shortcuts.tex
\newcommand{\bx}{\mathbf{x}}

\newcommand{\bI}{\mathbf{I}}

\newcommand{\bs}{\mathbf{s}}
\newcommand{\bM}{\mathbf{M}}

\newcommand{\bV}{\mathbf{V}}

\newcommand{\bR}{\mathbf{R}}

\newcommand{\bt}{\mathbf{t}}

\newcommand{\bz}{\mathbf{z}}
\newcommand{\bP}{\mathbf{P}}

\newcommand{\bA}{\mathbf{A}}

\newcommand{\bff}{\mathbf{f}}
\newcommand{\bF}{\mathbf{F}}
\newcommand{\bo}{\mathbf{o}}
\newcommand{\bO}{\mathbf{O}}

\newcommand{\bd}{\mathbf{d}}

\newcommand{\bPsi}{\boldsymbol{\Psi}}

\newcommand{\nR}{\mathbb{R}}

\newcommand{\nE}{\mathbb{E}}

\newcommand{\cD}{\mathcal{D}}
\newcommand{\cN}{\mathcal{N}}

\newcommand{\cL}{\mathcal{L}}

\newcommand{\cO}{\mathcal{O}}

\newcommand{\cV}{\mathcal{V}}

\makeatletter
\DeclareRobustCommand\onedot{\futurelet\@let@token\@onedot}
\def\@onedot{\ifx\@let@token.\else.\null\fi\xspace}
\def\eg{e.g\onedot} 
\def\ie{i.e\onedot}

\def\wrt{wrt\onedot}

\def\Fig{Fig\onedot}   
\makeatother

\newcommand{\figref}[1]{\Fig~\ref{#1}}
\newcommand{\secref}[1]{Section~\ref{#1}}

\renewcommand{\eqref}[1]{Eq.~\ref{#1}}
\newcommand{\tabref}[1]{Table~\ref{#1}}

\newcommand{\boldparagraph}[1]{\vspace{0.2cm}\noindent{\bf #1:} }

\newif\ifcomment
\commenttrue
\ifcomment
	\newcommand{\ag}[1]{ \noindent {\color{red} {\bf Andreas:} {#1}} }
	\newcommand{\yl}[1]{ \noindent {\color{cyan} {\bf Yiyi:} {#1}} }

\else
	\newcommand{\ag}[1]{}
	\newcommand{\yl}[1]{}
\fi

\newcolumntype{P}[1]{>{\arraybackslash}m{#1}}

%% file: tex/sec_0_abstract.tex
\begin{abstract}

Generating photorealistic images with controllable camera pose and scene contents is essential for many applications including AR/VR and simulation. Despite the fact that rapid progress has been made in 3D-aware generative models, most existing methods focus on object-centric images and are not applicable to generating urban scenes for free camera viewpoint control and scene editing.
To address this challenging task, we propose UrbanGIRAFFE, which uses a coarse 3D panoptic prior, including the layout distribution of uncountable stuff and countable objects, to guide a 3D-aware generative model. Our model is compositional and controllable as it breaks down the scene into stuff, objects, and sky. Using stuff prior in the form of semantic voxel grids, we build a conditioned stuff generator that effectively incorporates the coarse semantic and geometry information.  The object layout prior further allows us to learn an object generator from cluttered scenes. 
With proper loss functions, our approach facilitates photorealistic 3D-aware image synthesis with diverse controllability, including large camera movement, stuff editing, and object manipulation.
We validate the effectiveness of our model on both synthetic and real-world datasets, including the challenging KITTI-360 dataset. Project page:\urlstyle{same} \url{https://lv3d.github.io/urbanGIRAFFE}.

\end{abstract}

%% file: tex/sec_1_intro.tex
\section{Introduction}

\begin{figure}[t]
  \centering
     \includegraphics[width=\linewidth]{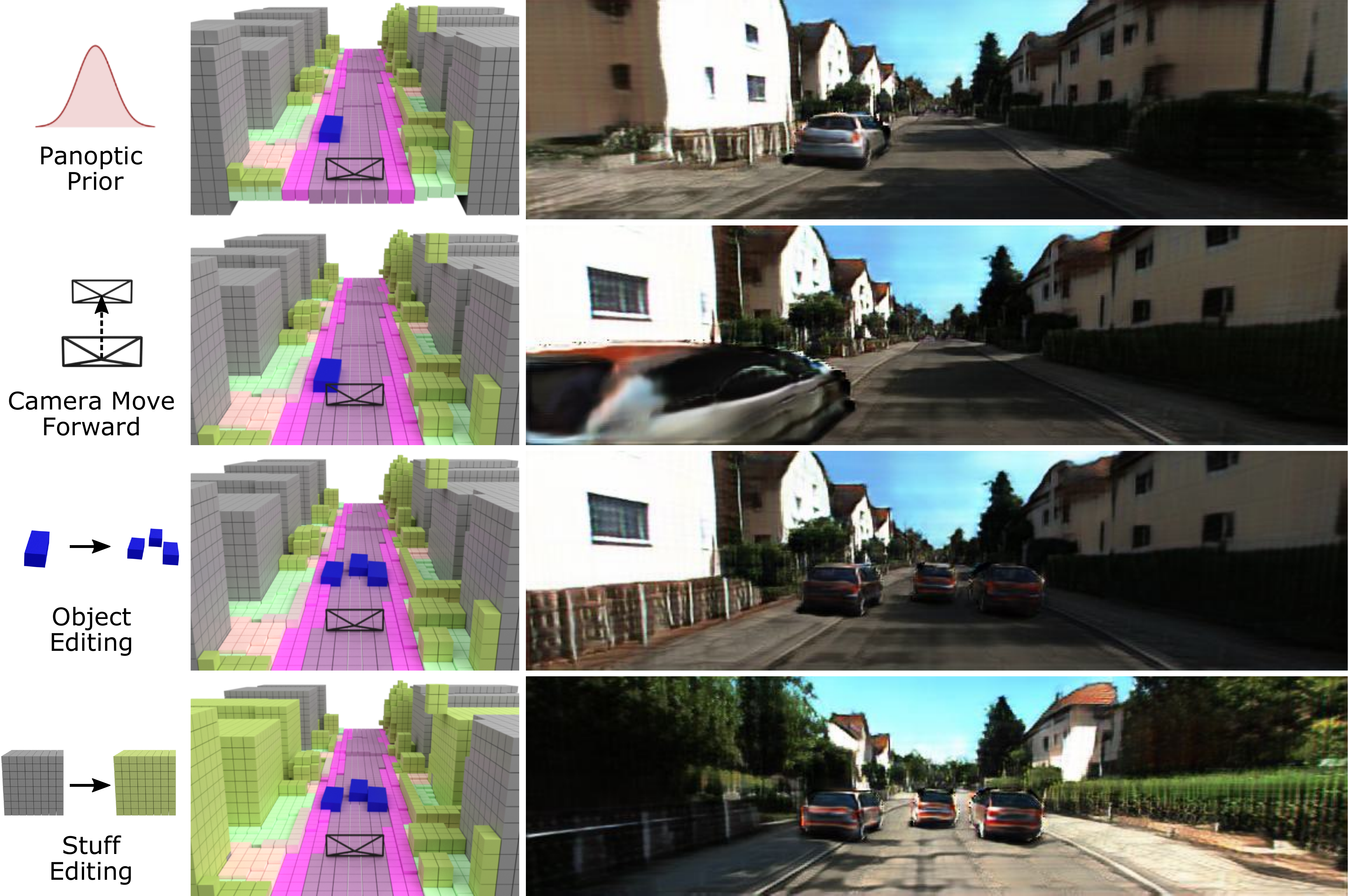}
   \caption{\textbf{Illustration.} UrbanGIRAFFE generates a photorealistic image given a sampled panoptic prior in the form of a semantic voxel grid and object layout. Our method enables diverse controllability regarding camera pose, instance, and stuff.}
   \label{fig:overview}
   \vspace{-0.2cm}
\end{figure}

Generating photorealistic urban scenes has many applications in simulation, gaming and virtual reality. 
Unfortunately, designing diverse urban scenes with novel 3D visual content is typically expensive and time-consuming as it requires the expertise of professional artists. 

Recent advances in generative models have demonstrated a promising direction to reduce the cost via learning to generate images from data.
Ideally, the generated scenes should be controllable in terms of camera pose and 3D content. For example, the camera should be able to move freely in the scene with six degrees of freedom. The poses of instantiated objects (\eg, cars) should be able to be manipulated independently. Furthermore, the layout of the scene should be controllable.

There are many attempts to generate photorealistic urban images. 
Several methods study semantic image synthesis to transfer a 2D semantic segmentation map to an RGB urban scene image~\cite{Isola2017CVPR,Park2019CVPRa,Schonfeld2021ICLR}. However, when changing the camera poses, the generated images across multiple frames may not be consistent using such 2D generative models.
Recently, 3D-aware generative models have witnessed a rapid progress by lifting the generation process to the 3D space. Despite achieving multi-view consistency, most existing 3D-aware generative models are limited to object-centric images, e.g., faces and cars~\cite{Schwarz2020NIPS,Chan2021CVPR,Chan2022CVPR}. 
There are a few attempts to generate scene images in a compositonal manner~\cite{Liao2020CVPRa, Niemeyer2021CVPR,Nguyen-Phuoc2020NEURIPS,Xu2022ARXIV}. However, all these methods struggle to learn a good geometry of the background and hence do not support large camera movement, e.g., moving the camera along the road. Another line of work enables camera movement but ignores the compositional nature of the scene, thus lacking controllability of the 3D content~\cite{DeVries2021ICCV,Bautista2022NEURIPS}.

In this paper, we propose UrbanGIRAFFE to address the challenging task of compositional and controllable 3D-aware image synthesis of urban scenes, see \figref{fig:overview}. Our key idea is to leverage scene-level but coarse 3D panoptic prior, simplifying the task of learning complex geometry through 2D supervision and incorporating semantic information for scene editing. The panoptic prior, including semantic voxel grids of uncountable stuff and bounding boxes of countable objects, can be obtained from existing datasets~\cite{Liao2022PAMI} or inferred from pre-trained models~\cite{Cao2022CVPR}.
Specifically, our model represents the scene as compositional neural feature fields consisting of stuff, objects, and sky.  %
We propose a semantic voxel-conditioned stuff generator, effectively preserving the semantic and geometry information provided by the prior. In terms of objects, we follow GIRAFFE~\cite{Niemeyer2021CVPR} to generate objects in canonical space by leveraging the object layout prior.  We further model the sky and far regions using a sky generator.
With all three generators, we render a composited feature map via volume rendering and upsample it to the target image using a neural renderer. 
For the complicated urban scenes, we observe that training with an adversarial loss on the full image alone is insufficient. We additionally employ an adversarial loss applied to objects and a reconstruction loss to the stuff image regions to improve the image fidelity.

Our contributions are as follows. i) We propose to study the challenging task of 3d-aware generative models for urban scenes with diverse controllability in terms of large camera movement, objects manipulation and stuff editing.  ii) We leverage coarse 3D panoptic prior to address this challenging task and design compositional generative radiance fields that leverages the prior information effectively.
iii) Our method demonstrates state-of-the-art performance compared to existing methods on both synthetic and real-world datasets, including the challenging KITTI-360 dataset. 

%% file: tex/sec_2_related.tex
\section{Related Work}
\label{sec:related_work}

\boldparagraph{Conditional Image synthesis}
In recent years, Generative Adversarial Networks~\cite{Goodfellow2014NIPS, Karras2019CVPR,Karras2020CVPRa,Karras2020NeurIPS,Karras2021NIPS,SauerS022SIGGRAPH} have achieved impressive results in photorealistic image synthesis. As it is not straightforward to control the generated images of unconditional GANs, many attempts have been made for conditional image synthesis. A line of works generates images conditioned on a 2D semantic segmentation map~\cite{Isola2017CVPR,Park2019CVPRa,Schonfeld2021ICLR}. Instead of requiring per-pixel semantic annotation, another line of methods generates images following an image layout in the form of 2D bounding boxes~\cite{Zhao2020IJCV,Yang2022CVPR,He2021CVPR} or learned  blobs~\cite{Epstein2022ECCV}. 
When changing the camera viewpoint, the generated images across different views are typically not multi-view consistent, as discussed in~\cite{Hao2021ICCV}. We instead learn a 3D-aware conditional generative model that leads to better consistency with the underlying 3D representation.

\boldparagraph{3D-Aware Image Synthesis}
3D-aware generative models have received growing attention recently. While early works learn to generate 3D voxel grids~\cite{Nguyen-Phuoc2019ICCV,Henzler2019ICCV}, recent methods achieve high-fidelity 3D-aware image synthesis leveraging neural radiance fields as the underlying 3D representation~\cite{Schwarz2020NIPS,Chan2022CVPR,Chan2021CVPR,Schwarz2022NEURIPS,Deng2022CVPR,Xu2022CVPR,gu2021stylenerf}. Empowered by 3D-aware generative models, many promising applications has been demonstrated, including semantic editing~\cite{Sun2022CVPR,Sun2022TOG}, relighting~\cite{Tan2022SIGGRAPH, Lee2022ARXIV}, single-view reconstruction~\cite{Cai2022CVPR,Muller2022CVPR} and articulated human generation~\cite{Zhang2022ECCV,Noguchi2022ECCV,Bergman2022ARXIV,Hong2022ARXIV}. However, all aforementioned methods focus on object-centric scenes and assume that the object lies in a canonical object coordinate system. Thus, it is non-trivial to extend these methods to complex, unaligned urban scenes. 
GSN~\cite{DeVries2021ICCV} and GAUDI~\cite{Bautista2022ARXIV} propose to generate unbounded indoor scenes. However, both ignore the compositionality of the scene, thus making it harder to achieve high visual fidelity and do not support editing of the scene content. 

A few works exploit the compositionality of 3D scenes to generate scenes containing multiple objects~\cite{Liao2020CVPRa,Xu2022ARXIV,Nguyen-Phuoc2020NEURIPS,Niemeyer2021CVPR,Xue2022CVPR}. 
As they consider the compositionality of foreground objects only, these methods are incapable of modeling complex background geometry in urban scenes. A concurrent work, DiscoScene~\cite{Xu2022ARXIV}, also study 3D-aware generative model of urban scenes. Despite achieving high-fidelity image synthesis, DiscoScene does not support camera control or stuff editing in urban scenes.

\boldparagraph{Neural Radiance Fields}
We proposed to present the scene as compositional neural feature fields. Exploiting implicit neural representations~\cite{Mescheder2019CVPR, Park2019CVPR}, NeRF~\cite{Mildenhall2020ECCV} has enabled impressive novel view synthesis by training a single model for each scene. Many exciting works have shown its potential in real-time rendering~\cite{Reiser2021ICCV, mueller2022instant}, geometric reconstruction~\cite{Wang2021NIPS, Wang2022ARXIV}, semantic segmentation~\cite{Zhi2021ICCV,Fu2022THREEDV,Kundu2022CVPR},  and view synthesis from sparse input~\cite{Wang2021ibrnet,yu2021pixelnerf,Chen2021ICCVmvs}.
It has been shown that NeRF can also be extended to model unbounded urban scenes ~\cite{Rematas2022CVPR, Ost2021CVPR} and scale to city level~\cite{Tancik2022CVPR,Xiangli2021ARXIV,Xiangli2022ECCV}. 
While all these methods focus on reconstructing existing scenes, we aim to learn a conditional generative model that can generate urban images conditioned on different panoptic layouts. A more related work, GANCraft~\cite{Hao2021ICCV}, aims to generate a scene based on semantic voxels, yet it also requires per-scene optimization. In contrast, our generative model allows for stuff editing by manipulating the semantic voxels.

\begin{figure*}[h]
    \centering
     \includegraphics[width=\linewidth]{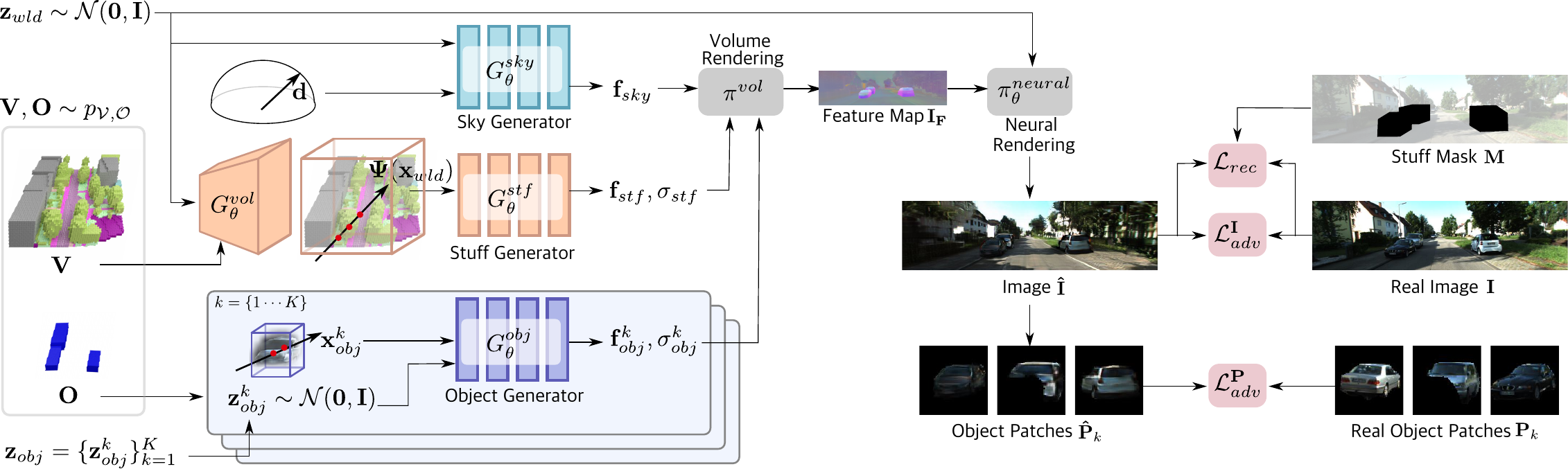}
  
     \caption{\textbf{Method Overview.} We leverage panoptic prior in the form of semantic voxel grids and instance object layout to build a 3D-aware generative model for urban scenes. Our model takes as input a global noise vector $\bz_{wld}$ for the entire scene, $K$ noise vectors $\{\bz^k_{obj}\}_{k=1}^K$ for objects, and a sampled panoptic prior $\bV,\bO\sim p_{\cV,\cO}$. We decompose the scene into sky, stuff, and objects. The stuff generator is conditioned on the semantic voxel grid $\bV$ to preserve its semantic and geometry information. The objects are generated in the canonical object coordinate system guided by $\bO$. Combined with the sky generator, a feature map $\bI_\bF$ is obtained via volume rendering. We further leverage neural rendering to output the RGB image $\hat{\bI}$ and object patches $\hat{\bP}_k$. The full model is optimized jointly with adversarial losses $\cL_{adv}^{\bI}$ and $\cL_{adv}^{\bP}$ applied to the full image and object patches, respectively, as well as a reconstruction loss $\cL_{rec}$ for stuff regions. }
     \label{fig:method}
  \end{figure*}

%% file: tex/sec_3_method.tex
\section{Method}

\label{sec:method}
In UrbanGIRAFFE, our goal is to build compositional generative feature fields of urban scenes with control over camera pose and scene contents.
To address this challenging task, we decompose the urban scene into three main components, including uncountable stuff, countable objects, and sky, see \figref{fig:method} for an overview. We assume prior distributions are provided for both stuff and objects in order to disentangle the complicated urban scenes. 
Given a camera pose, we render a composited feature map and generate the target image via neural rendering. Our model is trained end-to-end with adversarial and reconstruction losses. 

In this section, we first introduce the prior distributions of stuff and objects, respectively. Next, we introduce our compositional generative model for urban scene generation. Finally, we describe the sampling strategy, loss functions and implementation details.

\subsection{Panoptic Prior}
We assume a prior distribution of the scene layout is given in order to train our generative model, which we refer to as ``panoptic prior''.
The panoptic prior briefly describes the spatial distributions of countable objects and uncountable stuff within a certain region. Let $\bV,\bO\sim p_{\cV,\cO}$ denote a stuff layout $\bV$ and an object layout $\bO$ sampled from the joint distribution $p_{\cV,\cO}$. We now elaborate on the layout representation of $\bO$ and $\bV$, respectively.

\boldparagraph{Countable Object}
Following GIRAFFE, the layout distribution of countable objects (\eg, cars) is represented in the form of a set of 3D bounding boxes. A sample $\bO = \{ \bo_1, \bo_2,..\bo_K\}$ depicts a joint distribution of $K$ objects in one scene, where $K$ may vary for different scenes. Here, each object $\bo$ is represented by a 3D bounding box parameterized by its rotation $\bR\in SO(3)$, translation $\bt \in \nR^3$, and size $\bs \in \nR^3$:
\[
 \bo_k = \{\bR_k, \bt_k, \bs_k\}
\]
In this work, we leverage bounding boxes released by publicly available dataset~\cite{Liao2022PAMI} to form the distribution $p_\cO$. This distribution can be obtained from real-world images, e.g., by applying a 3D object detection method. 

\boldparagraph{Uncountable Stuff}
Unlike countable objects, there are many indispensable entities that are either uncountable (\eg, road and terrain) or sometimes too cluttered to be separated (\eg, trees). To address this problem, we represent uncountable stuff in the form of semantic voxel grids $\bV\in \nR^{H_v \times W_v \times D_v \times {L}}$, where each voxel stores a one-hot semantic label of length $L$.

\subsection{Compositional Urban Scene Generator}
Our generator follows the idea of GIRAFFE~\cite{Niemeyer2021CVPR} which represents the urban scene as a compositional neural feature fields. A key difference is that we model the background using a stuff generator conditioned on a semantic voxel grid, assisted with a sky generator to model sky and far regions. The stuff and the sky generator share a global latent code $\bz_{wld}\in\cN(\mathbf{0},\mathbf{I})$, whereas each object has its own latent code $\bz_{obj} = \{ \bz_{obj}^k\in\cN(\mathbf{0},\mathbf{I}) \}_{k=1}^{K}$ to ensure the diversity of object shape and appearance in a scene. We now describe each of these generators in detail.

\boldparagraph{Object Generator} %
For objects, we follow existing compositional methods to generate each object $k$ in a normalized object coordinate space~\cite{Liao2020CVPRa,Niemeyer2021CVPR}:
\begin{equation}
G_\theta^{obj}: (\gamma(\bx^k_{obj}), \bz^k_{obj}) \mapsto (\bff^k_{obj}, \sigma^k_{obj})
\end{equation}
where $G_\theta^{obj}$ denotes the object generator that maps a 3D point $\bx^k_{obj}$ encoded by positional encoding $\gamma(\cdot)$ and a noise vector $\bz^k_{obj}$ to a feature vector $\bff^k_{obj}\in\nR^{M_f}$ and density $\sigma^k_{obj}$. Here, $\bx^k_{obj}$ denotes a 3D point in the $k$th normalized object coordinate which is transformed to the world coordinate given the object transformation $\{\bR, \bt, \bs\}$. 
\begin{equation}
\label{eq:transform}
\bx_{wld} = \bR(\bs \odot \bx^k_{obj})+ \bt
\end{equation}
Generating objects in this canonical space enables information sharing across different objects, thus allowing for learning a complete shape from many single-view object images. With the learned complete shape, we can control the rotation, translation, and appearance of each individual object.

\begin{figure}[t]
  \centering
   \includegraphics[width=\linewidth]{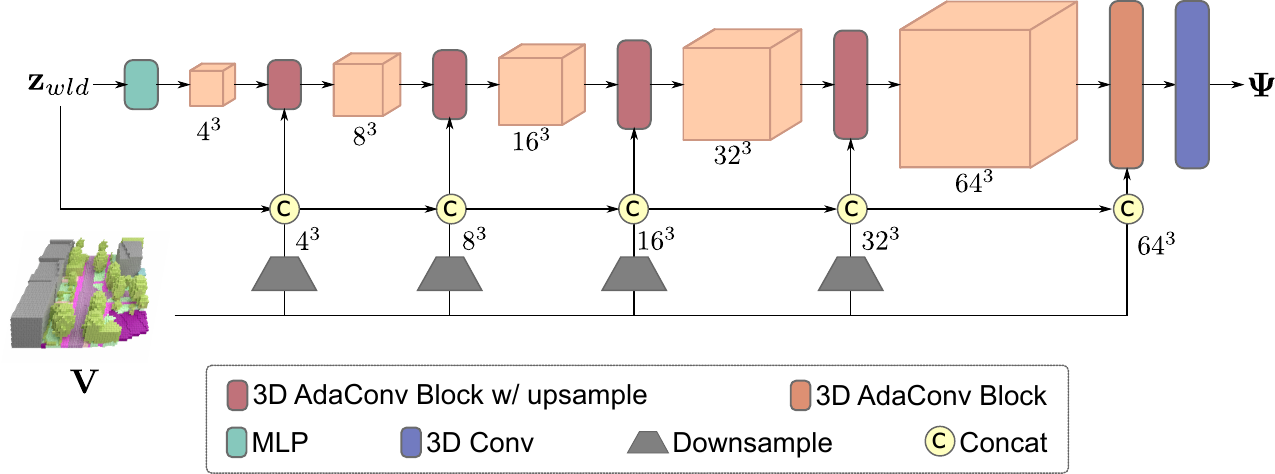}
   \caption{\textbf{Feature Grid Generator} $G_{\theta}^{vol}$ as a part of the stuff generator. We adopt spatially-adaptive normalization to inject the semantic condition $\bV$ and the noise vector $\bz_{wld}$  at multiple resolutions. }
   \label{fig:SPADE3D}
\end{figure}

\boldparagraph{Stuff Generator}
Our stuff generator generates feature fields for the uncountable stuff condition on the semantic voxel grid $\bV$. Inspired by 2D semantic image synthesis~\cite{Park2019CVPRa,Schonfeld2021ICLR}, we use the semantic voxel grid to modulate the stuff generation. More specifically, our stuff generator consists of a \textit{feature grid generator} $G^{vol}_\theta$ and a \textit{MLP head} $G^{stf}_\theta$. The feature grid generator first maps the noise vector $\bz_{wld}$ to a feature grid $\bPsi \in \nR^{H_v\times W_v\times D_v \times M_v}$ conditioned on the semantic voxel grid $\bV \in \nR^{H_v \times W_v \times D_v \times L}$:
\begin{align}
G_\theta^{vol}: (\bz_{wld}, \bV)  &\mapsto \bPsi  
\end{align}
In practice, $G^{vol}_\theta$ is a 3D convolutional neural network. The semantic condition $\bV$ is injected at multiple resolutions using spatially-adaptive normalization, see \figref{fig:SPADE3D} as an illustration.
Given a 3D point $\bx_{wld}$, we trilinearly interpolate a feature vector $\Psi(\bx_{wld})\in\nR^{M_v}$. Next, we map $\bx_{wld}$ and $\Psi(\bx_{wld})$ to the final stuff feature $\bff_{stf}\in\nR^{M_f}$ and density $\sigma_{stf}$ using the MLP head:
\begin{align}
G_\theta^{stf}: (\Psi(\bx_{wld}) , \gamma(\bx_{wld})) &\mapsto (\bff_{stf}, \sigma_{stf}) 
\end{align}
where $\gamma(\cdot)$ denotes positional encoding.

\boldparagraph{Sky Generator}
The stuff generator cannot model regions far from the semantic voxel grid, \eg, sky. Therefore, we model the sky and other far regions as an infinitely far away dome following \cite{Hao2021ICCV,Rematas2022CVPR}. Specifically, we use a sky generator $G_{\theta}^{sky}$ to map a ray direction $\bd$ to a sky feature vector $\bff_{sky}\in\nR^{M_f}$.
\[
 G_{\theta}^{sky}: ( \bz_{wld}, \bd)  \mapsto \bff_{sky}
\] 
Note that the global latent code $\bz_{wld}$ is used to ensure the style consistency between sky and other semantics within an urban scene. 

\boldparagraph{Compositional Volume Rendering}
We accumulate feature vectors of objects, stuff, and sky on each ray via compositional volume rendering.
We first sample points from the object and stuff generators independently (the sampling strategy will be elaborated in \secref{sec:sample}).
Next, we sort all points wrt. their distances to the camera center and accumulate their feature vectors via volume rendering. Finally, the sky feature is added to non-opaque regions. 

Formally, let $\{\bx_i\}_{i=1}^{M}$ denote $M$ sorted points on a ray, compositing of $\bx_{wld}$ sampled for the stuff generator and $\bx_{obj}^k$ sampled for the object generators (transformed to the world coordinate system via \eqref{eq:transform}). $\bff_i$ and $\sigma_i$ denote the corresponding feature vector and density at $\bx_i$. The volume rendering is 
  \begin{align}
    \pi^{vol}: \{  \bff_i, \sigma_i, \bff_{sky} \}_{i=1}^{M} &\mapsto \bF %
  \end{align}
Specifically, $\bF$ is obtained via numerical integration as
\begin{align}
    \bF = \sum_{i=1}^{N} T_i \alpha_i \bff_i + (1- \sum_{i=1}^{N} T_i \alpha_i )\bff_{sky} \\
    \alpha_i = 1-e^{ (-\sigma_i\delta_i)} \quad  T_i   = \prod_{j=1}^{i-1}\left(1-\alpha_j\right)
\end{align}
where $T_i$ and $\alpha_i$ denote transmittance and alpha value of a sample point $\bx_i$.

\boldparagraph{2D Neural Rendering}
  Following~\cite{Niemeyer2021CVPR}, we adopt a neural renderer to transform the rendered feature map to an output RGB image at the target resolution. This allows us to scale to a higher resolution without extensive computation burden.
 More specifically, our 2D neural renderer $\pi^{neural}_{\theta}$  maps the feature image $ \bI_\bF \in \nR^{H_f \times W_f \times M_f}$ and the noise vector $\bz_{wld}$ to the RGB image $\hat{\bI} \in \nR^{H \times W \times 3}$ at the target resolution. Here, $\bz_{wld}$ is adopted to enable content-aware upsampling.
  \begin{align}
    \pi_\theta^{neural}: (\bI_\bF, \bz_{wld} ) &\mapsto \hat{\bI} %
  \end{align}

\boldparagraph{Object Patch Rendering}
In addition to the full image $\hat{\bI}$, we further generate a set of object patches, see \figref{fig:method} as an illustration. We upsample object masks obtained from volume rendering to segment the objects after neural rendering. Please refer to the supplementary for more details.

\subsection{Sampling Strategy}
\label{sec:sample}
We use the panoptic prior to guide the sampling of volume rendering, effectively reducing the required number of sampling points and improving rendering efficiency. 

\boldparagraph{Ray-Voxel Intersection Sampling for Stuff} Inspired by existing methods~\cite{Hao2021ICCV,Liu2020NIPS}, we use the ray-voxel intersection sampling strategy to determine sampling locations for the stuff generator. For each ray, we find the first $4$ non-empty voxels that the ray hit, and then sample $M_{vol}$ points within these voxels. This effectively reduces the number of required sampling points by avoiding sampling in the empty space and occluded regions.

\boldparagraph{Ray-Box Intersection for Object}
For objects, we also leverage the 3D bounding boxes to reduce the number of samples in the empty space. Given a ray, we first calculate the ray-box intersections for each bounding box parameterized by $(\bR, \bt, \bs)$. Next, we sample $M_{obj}$ points within each bounding box by uniform sampling between the intersections. We use the stratified sampling strategy following~\cite{Mildenhall2020ECCV}, i.e., a random shift is added to the sampled points.

\subsection{Loss Functions}

We train the entire model end-to-end using adversarial training aided by a reconstruction loss for stuff regions.

\boldparagraph{Adversarial Loss}
We apply an adversarial to the composited image. Let $G_\theta$ denote the full conditional generator that maps the noise vectors and the panoptic prior to a full RGB image:
\begin{equation}
G_\theta: (\bz_{wld}, \bz_{obj} , \bV, \bO) \mapsto \hat{\bI}
\end{equation}
We apply the non-saturated adversarial loss with R1-regularization~\cite{Mescheder2018ICML}:
\begin{align}\nonumber
 & \cL_{adv}^{\bI}  =  \nE_{\bI \sim p_{\cD}}
\left[
f(-D^{\bI}_\phi(\bI))
\,-\, \lambda {\Vert \nabla D_\phi^\bI(\bI)\Vert}^2 \right] + \\ 
& \nE_{\bz_{wld},\bz_{obj}\sim \cN,\bV,\bO\sim p_{\cV,\cO}}
\left[f(D_\phi^{\bI}(G_\theta(\bz_{wld},\bz_{obj},\bV,\bO)))\right]
\end{align}
Note that the visual quality of objects like cars is essential for urban scenes. Unfortunately, objects do not always occupy a large area in urban images. Our experiments show that using scene-level adversarial training alone fails to generate photorealistic objects. Inspired by existing methods~\cite{Gadde2021ICCV, Xu2022ARXIV}, we adopt object-level discriminative training by feeding the object patches $\hat{\bP}$ to another object discriminator $D_\phi^\bP$, leading to the object-level adversarial loss $\cL_{adv}^{\bP}$ similar to $\cL_{adv}^{\bI}$.

\boldparagraph{Stuff Reconstruction Loss}
For our conditional stuff generator, we observe that using adversarial loss alone struggles to generate photorealistic results. One possible reason is that learning generative 3D feature fields for complex stuff regions is more challenging than the object-centric generation. 
To stabilize adversarial training and improve the quality of synthesized images, we further leverage reconstruction loss for stuff regions. 
Following~\cite{Hao2021ICCV}, our reconstruction loss is a combination of the MSE loss and perceptual loss $l_{vgg}$~\cite{Johnson2016ECCV}:
\begin{align*}
    \cL_{recon} = \mathbb{E} \big[\left\lVert \bM \odot( \bI - \hat{\bI}) \right\rVert_2^2 + \lambda_{vgg} l_{vgg}(\bM\odot \bI, \bM\odot\hat{\bI}) \big] 
\end{align*}
where $\bI$ and $\hat{\bI}$ are paired samples, and $\bM$ denotes a mask that filters out object regions based on the projected 3D projecting boxes $\bO$.
Since our stuff generator is a conditioned generative model depending on the semantic voxel grid, adding the reconstruction loss is reasonable as the appearance is highly relevant to the corresponding semantic label. This provides stronger supervision that $\bz_{wld}$ only needs to model the variation within the same semantic class. 

\subsection{Implementation Details}

We use 3D CNNs with 5 spatially-adaptive normalization blocks for the feature grid generator $G_\theta^{vol}$. We set $H_v=W_v=D_v=64$ for all experiments, i.e., the semantic voxel grids are at the resolution of $64^3$.  We use $M_v=16$ channels for the feature grid $\bPsi$ to avoid large memory consumption.  The MLP head $G_\theta^{stf}$ of the stuff generator is an 8-layer ReLU MLP with a hidden dimension of 256. The object generator $G_\theta^{obj}$ is also a 8-layer ReLU MLP with a hidden dimension of 128. In terms of the sky generator $G_\theta^{sky}$, a 5-layer MLP with a hidden dimension of 256 is adopted. All these MLP generators output feature vectors of dimension $M_f=32$.

During training, we sample camera poses along plausible driving trajectories given a semantic voxel grid. 
Regarding ray marching, we sample $M_{obj} = 12$ points within each object's bounding box and $M_{vol}=6$ within each voxel.
We use the Adam optimizer with a batch size of 16. The learning rates of the discriminator and the generator are $1 \times 10^{-4}$ and $2 \times 10^{-4}$, respectively. 
During inference, we generate images using a moving averaged model with an exponential decay of $0.999$ for the weights. 

%% file: tex/sec_4_results.tex
\input{gfx/results/comp_baseline_kitti360}

\input{gfx/results/comp_baseline_clevr}

\section{Experimental Results}
\label{sec:experimenmt}
In this section, we first compare our method to several 2D and 3D baselines on both synthetic and real-world datasets. Subsequently, we design a number of controllable urban scene editing experiments to evaluate the preferences of our synthesis model with regards to controllability and fidelity. We further conduct ablation studies to better understand the influence of different architectural components.

\input{gfx/results/tables/baseline_comp}

\boldparagraph{Datasets} We conduct experiments on two multi-object datasets with diverse backgrounds. %
\textbf{KITTI-360}~\cite{Liao2022PAMI} is an outdoor sub-urban dataset containing complex scene geometry.
Furthermore, scenes in KITTI-360 are replete with highlights and shadows, causing the appearance of the same object to vary greatly in different scenes. 
KITTI-360 provides coarse 3D bounding primitives in cuboids and spheres for both stuff and objects. We consider cars as objects since cars are important for driving scenarios. For stuff regions, we simply convert the coarse 3D bounding primitives to semantic voxel grids.
We further create an augmented \textbf{CLEVR-W} dataset following CLEVR~\cite{Johnson2017CVPR}. In contrast to existing methods~\cite{Niemeyer2021CVPR,Xu2022ARXIV} that places objects on a simple flat background in CLEVR, we introduce walls into the background. We consider the wall and the floor as stuff regions.
Please refer to the supplementary material for additional information regarding the CLEVR-W dataset.

\boldparagraph{Baselines} We compare our approach to two state-of-the-art models GIRAFFE~\cite{Niemeyer2021CVPR} and GSN~\cite{DeVries2021ICCV} for 3D-aware image synthesis. To further evaluate the fidelity of the synthesized image, we additionally compare our method with a state-of-the-art 2D method, StyleGAN2~\cite{Karras2020CVPRa}.

\input{gfx/results/control_full_kitti360}

\boldparagraph{Metrics}
 We report the FID~\cite{Heusel2017NIPS} and KID~\cite{Binkowski2018ICLR} scores to quantify image quality. We use 5k real and fake samples to calculate the FID and KID score.

\subsection{Comparison to the State of the Art}

\boldparagraph{Quantitative Comparison}
\tabref{tab:comp_baselines} shows the quantitative comparison on KITTI-360 and CLEVR-W. Note that GSN requires training on sequential frames, thus we omit GSN on the CLEVR-W dataset which does not contain sequential data. The quantitative comparison shows that our method greatly outperforms existing state-of-the-art 3D methods regarding image fidelity and is comparable to the 2D baseline.  

\boldparagraph{Qualitative Comparison}
We compare our method with GIRAFFE and GSN on KITTI-360 in \figref{fig:com_baseline_kitti360} with the camera moving forward. 
Note that GIRAFFE struggles to learn the complicated background geometry of urban scenes. This distracts the GAN training, thus leading to low-quality results even in a static scenario (the first row). Compared with GIRAFFE, GSN's scene representation is built upon a local 2D feature map, enabling it to model relatively complex 3D scenes. Therefore, GSN performs better in the static scenario, but the image quality drops dramatically as the camera moves forward. 
As a comparison, our method is conditioned on a 3D semantic voxel grid, thus enabling  photorealistic and consistent 3D-aware image synthesis even with a large camera moving distance. 

\figref{fig:com_baselines_clevr} shows the qualitative comparison with GIRAFFE on CLEVR-W. We conduct various experiments including stuff editing (\eg, editing the height of the wall or moving it closer to the objects), object rearrangement, and camera viewpoint manipulation. Note that GIRAFFE performs well on the foreground objects but still lags behind on the background. In contrast, our method can keep high fidelity and 3D consistency under these experiments, which clearly outperforms the baseline method.

\input{gfx/results/ablation_kitti360}

\input{gfx/results/comp_gt_kitti360}

\subsection{Controllable Urban Scene Generation} 
We now demonstrate the diverse controllability of our model in terms of stuff editing, object editing and camera viewpoint control.

\boldparagraph{Stuff Editing}
Our semantic-conditioned stuff generator enables fine-grained stuff editing by modifying the conditioning semantic voxel. As shown in \figref{fig:stuff_editing}. We can transfer stuff semantics like ``Road to Grass'' and ``Building to Tree''. It is also possible to edit the occupancy of the voxel grids, \eg, ``Lower building'' and ``Move tree''. All these stuff editings are achieved by modifying the semantic voxel grid without additional optimization.

It is worth mentioning that, in the ``Building To Tree'' example,  the shadow of the road also changes to a large degree after the editing. This suggests that our method not only allows for photorealistic and semantically-align urban scene generation but also learns the implicit relationship between the shadow condition and semantic layout. 

\boldparagraph{Object Editing}
Next, we conduct various experiments on object editing in \figref{fig:object_editing}. As in GIRAFFE~\cite{Niemeyer2020ARXIV}, we can add/delete objects, and control their appearance, rotation, and translation.
Our object experiments with object editing do not affect the appearance of other scene parts, suggesting that our method can disentangle objects from the complex background by leveraging the panoptic prior.

\boldparagraph{Camera Control}
Finally,  \figref{fig:camera_control} shows that our method also allows for large viewpoint control, including large rotation in azimuth and polar angles as well as in-plane rotation. 
We can also change the camera's focal length, successfully capturing a photorealistic wide-angle image.

\subsection{Ablation Study}

To verify our design choices, we conduct ablation studies on the KITTI-360 dataset, and evaluate both image-level and patch-level FID/KID scores in  \tabref{tab:ablation_study}.

\input{gfx/results/tables/ablation_comp}

\boldparagraph{Reconstruction Loss}
We first validate the role of the reconstruction loss. After removing the reconstruction loss, the FID and KID scores drop significantly  (w/o $\cL_{recon}$).  This is unsurprising as the reconstruction loss provides stronger supervision to align the generated scenes with the ground truth. \figref{fig:ablation} shows that removing the reconstruction loss can also lead to reasonable performance, but yields more artifacts. Moreover, reconstruction loss is particularly important for infrequently encountered semantic classes.  
For example, removing reconstruction loss results in the model rendering the ``rail track'' as grass, while the full model can render it with the corresponding semantic meaning faithfully (see 3rd row of \figref{fig:ablation}). Note that our full model can  maintain high fidelity while still exhibiting differences from the ground truth image, see \figref{fig:ablation_comp_gt}.
These findings suggest our full model can produce diverse results instead of simply remembering the entire dataset.

\boldparagraph{Object Discriminator}
Next, we exclude the adversarial loss $\cL_{adv}^{\bP}$ applied to object patches and train the object generator solely through the image adversarial loss $\cL_{adv}^{\bI}$. As shown in \tabref{tab:ablation_study}, removing $\cL_{adv}^{\bP}$ significantly increases the patch FID$_{\bP}$ and KID$_{\bP}$. This can also be seen from the qualitative results in \figref{fig:ablation}, where the cars are of lower quality when removing the object discriminator.
It is worth noting that FID$_\bI$ is less affected, indicating that in scenes where the proportion of objects pixels is not large, the global adversarial training cannot provide enough supervision to optimize objects which we actually care, and hence introducing  $\cL_{adv}^{\bP}$ is important to improve visual quality.

\boldparagraph{All Stuff} Lastly, we remove the object generator $G_\theta^{obj}$ and use the stuff generator to represent the full scene except for the sky (w/o $G_\theta^{obj}$), similar to the GSN approach. This can also be considered as a generative version of GANCraft~\cite{Hao2021ICCV}. As shown in \figref{fig:ablation}, the quality of objects drops significantly. This verifies the importance of decomposing stuff and objects to learn high-fidelity object generation. Note that $\cL_{adv}^{\bP}$ is also not applied in this experiment as there is no information of object instance.

%% file: gfx/results/comp_baseline_kitti360.tex
\begin{figure*}[htbp]
     \centering
     \setlength{\tabcolsep}{1pt}
     \def\mywidth{.33}
     \def\reduceheight{-3.pt}
     \begin{tabular}{ccc}

      \includegraphics[width=\mywidth\linewidth]{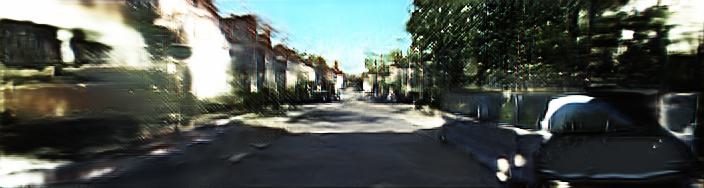} &
      \includegraphics[width=\mywidth\linewidth]{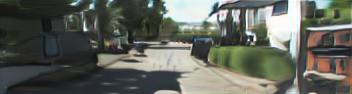} &
      \includegraphics[width=\mywidth\linewidth]{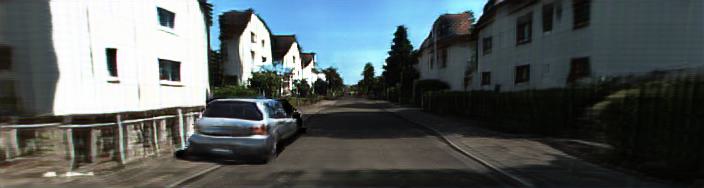} 
      \vspace{\reduceheight}\\
      \includegraphics[width=\mywidth\linewidth]{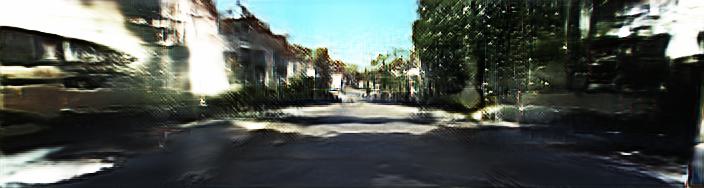} &
      \includegraphics[width=\mywidth\linewidth]{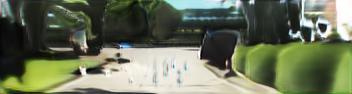} &
      \includegraphics[width=\mywidth\linewidth]{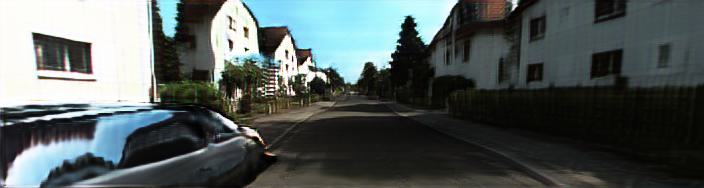} 
      \vspace{\reduceheight}\\
      \includegraphics[width=\mywidth\linewidth]{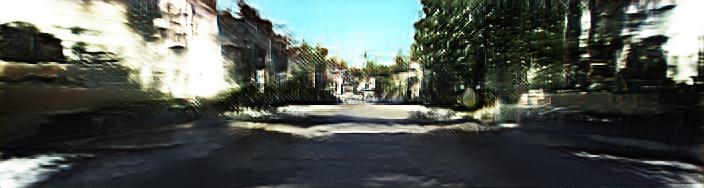} &
      \includegraphics[width=\mywidth\linewidth]{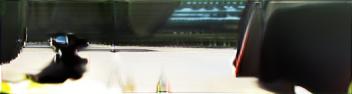} &
      \includegraphics[width=\mywidth\linewidth]{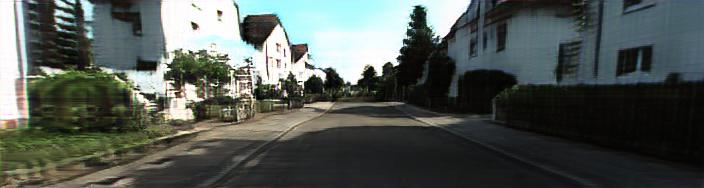} 
      \vspace{\reduceheight}\\

      \begin{small}GIRAFFE~\cite{Niemeyer2021CVPR}\end{small} &
      \begin{small}GSN~\cite{DeVries2021ICCV}\end{small} &
      \begin{small}UrbanGIRAFFE (Ours) \end{small} 
     \end{tabular}\vspace{-0.2cm}
     \caption{{\bf Qualitative Comparison on KITTI-360}. The 1st row shows images rendered at the default camera pose for each method. The camera moves forward in the remaining rows, with an accumulated moving distance of 10 meters.}
     \label{fig:com_baseline_kitti360}
     \vspace{-0.021cm}
    \end{figure*}

%% file: gfx/results/comp_baseline_clevr.tex
\begin{figure*}[htbp]
     \centering
     \setlength{\tabcolsep}{0pt}
     \def\mywidth{1.51cm}
     \def\largewidth{1.55cm}
     \def\reduceheight{-3.5pt}
     \begin{tabular}{P{0.4cm}P{\mywidth}P{\largewidth}P{\mywidth}P{\mywidth}P{\largewidth}P{\mywidth}P{\largewidth}P{\mywidth}P{\largewidth}P{\mywidth}P{\mywidth}}
      \rotatebox{90}{\scriptsize Ours} &
      \includegraphics[width=\mywidth]{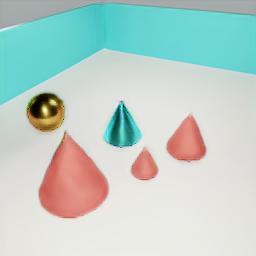} &
      \includegraphics[width=\mywidth]{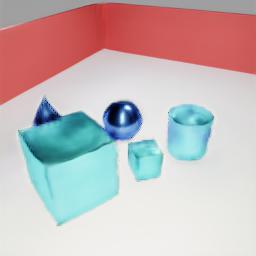} &
      \includegraphics[width=\mywidth]{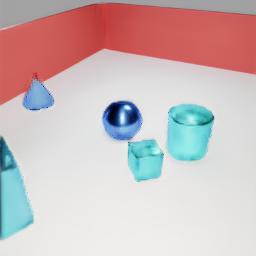} &
      \includegraphics[width=\mywidth]{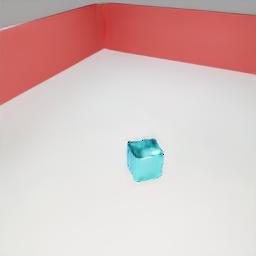} &
      \includegraphics[width=\mywidth]{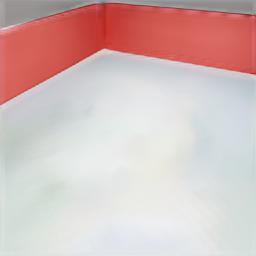} &
      \includegraphics[width=\mywidth]{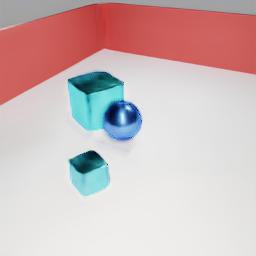} &
      \includegraphics[width=\mywidth]{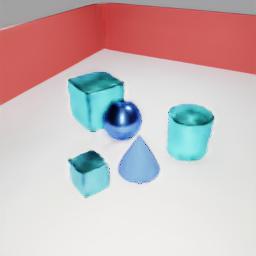} &
      \includegraphics[width=\mywidth]{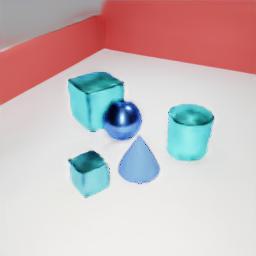} &
      \includegraphics[width=\mywidth]{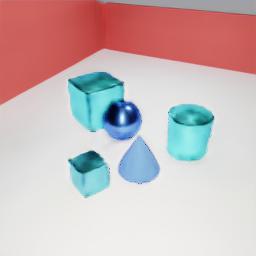} &
      \includegraphics[width=\mywidth]{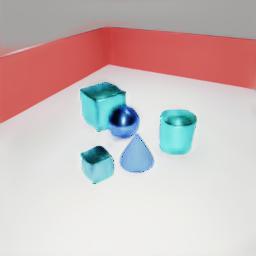} &
      \includegraphics[width=\mywidth]{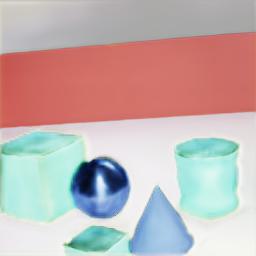} \\
      \vspace{\reduceheight}
      \rotatebox{90}{\scriptsize GIRAFFE~\cite{Niemeyer2021CVPR}} &
      \includegraphics[width=\mywidth]{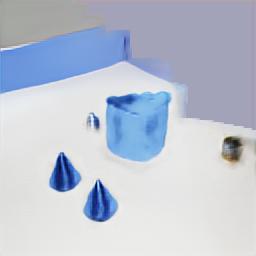} &
      \includegraphics[width=\mywidth]{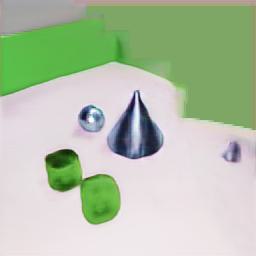} &
      \includegraphics[width=\mywidth]{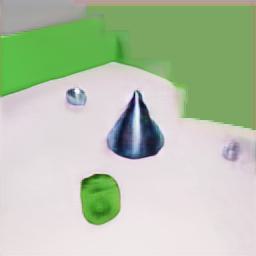} &
      \includegraphics[width=\mywidth]{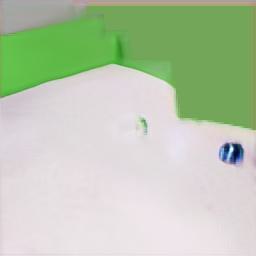} &
      \includegraphics[width=\mywidth]{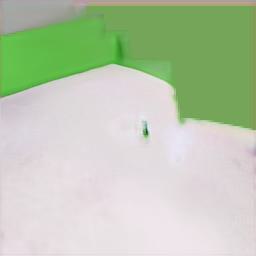} &
      \includegraphics[width=\mywidth]{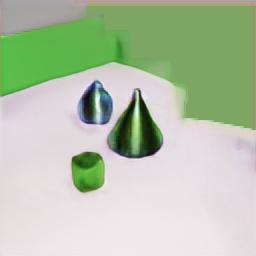} &
      \includegraphics[width=\mywidth]{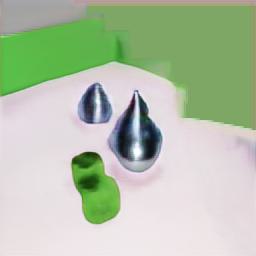} &
      \includegraphics[width=\mywidth]{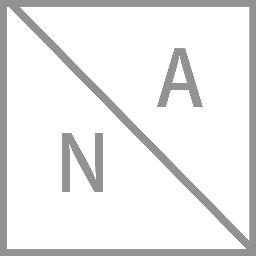} &
      \includegraphics[width=\mywidth]{gfx/results/place_holder_celvr.jpg}& 
      \includegraphics[width=\mywidth]{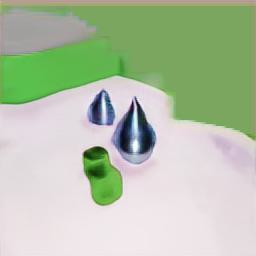} &
      \includegraphics[width=\mywidth]{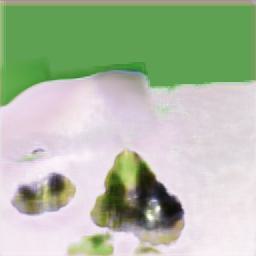} 
      \\
     & \multicolumn{2}{c}{\small Appearance} & \multicolumn{3}{c}{\small Object Removal} & \multicolumn{2}{c}{\small Object Insertion} & \multicolumn{2}{c}{\small Stuff Editing} & \multicolumn{2}{c}{\small Camera Control}
     \end{tabular}\vspace{-0.1cm}
     \caption{{\bf Qualitative Comparison on CLEVR-W}. We compare with GIRAFFE \wrt various controllable image synthesis tasks. Our method outperforms GIRAFFE in modeling the background, thus enabling stuff editing and better performance in camera viewpoint control.}
     \label{fig:com_baselines_clevr}
     \vspace{-0.021cm}
    \end{figure*}

%% file: gfx/results/tables/baseline_comp.tex
\begin{table}[]
    \centering
    \small
    \begin{tabularx}{\linewidth}{lXXXX}
    \toprule
                                 & \multicolumn{2}{c}{KITTI-360} & \multicolumn{2}{c}{CLEVR-W} \\ 

      \cmidrule(l){2-3} \cmidrule(l){4-5}
                       &  $\text{FID}_{\bI}\downarrow$          &  $\text{KID}_{\bI}\downarrow$       &  $\text{FID}_{\bI}\downarrow$          &  $\text{KID}_{\bI}\downarrow$         \\ \hline
    2D GAN~\cite{Karras2020CVPR}                   & 31.9          & 0.021          & 17.7           & 0.019         \\   \hline
    GSN~\cite{DeVries2021ICCV}                     & 160.0         & 0.114          & --              & --              \\ 
    GIRAFFE~\cite{Niemeyer2021CVPR}                  & 112.1         & 0.117         & 103.9          & 0.101          \\ 
    Ours                     & \textbf{39.6}         & \textbf{0.036}          & \textbf{25.7}          & \textbf{0.019}         \\ 
    \bottomrule
    \end{tabularx}
    \caption{{\bf Quantitative Comparison} on KITTI-360 and CLEVR-W. Our method outperforms 3D-aware baseline methods and is comparable to the 2D baseline.}
    \label{tab:comp_baselines}
    \end{table}

%% file: gfx/results/control_full_kitti360.tex
\begin{figure*}[htbp]
  \centering
  \setlength{\tabcolsep}{1pt}
  \def\mywidth{.245}
  \def\reduceheight{-3.5pt}
  \subfloat[][Stuff Editing\label{fig:stuff_editing}\vspace{-0.3cm}]{
  \begin{tabular}{cccc}
    \includegraphics[width=\mywidth\linewidth]{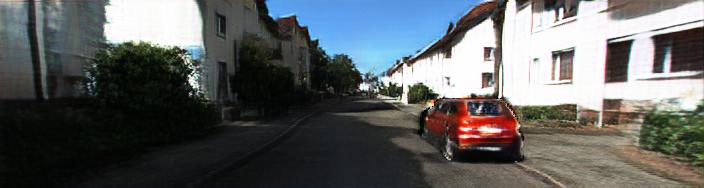} &
    \includegraphics[width=\mywidth\linewidth]{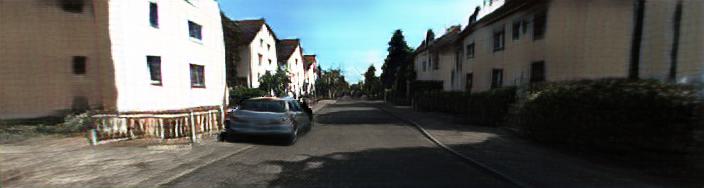}&
    \includegraphics[width=\mywidth\linewidth]{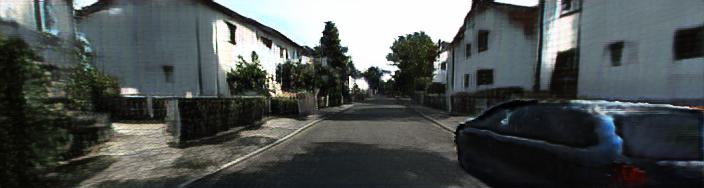} &
    \includegraphics[width=\mywidth\linewidth]{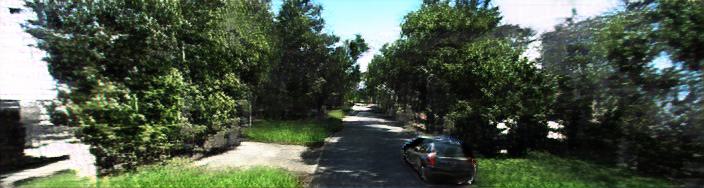} \vspace{\reduceheight}\\
    \includegraphics[width=\mywidth\linewidth]{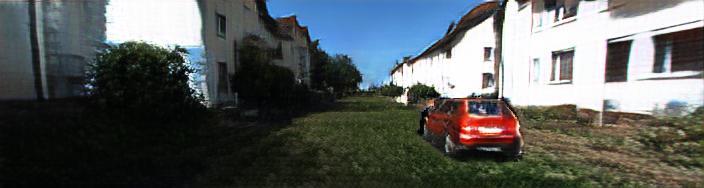} &
    \includegraphics[width=\mywidth\linewidth]{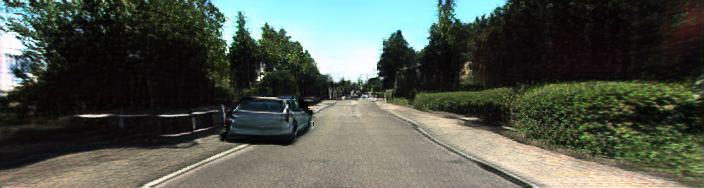}&
    \includegraphics[width=\mywidth\linewidth]{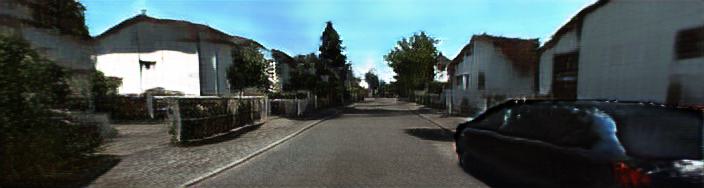} &
    \includegraphics[width=\mywidth\linewidth]{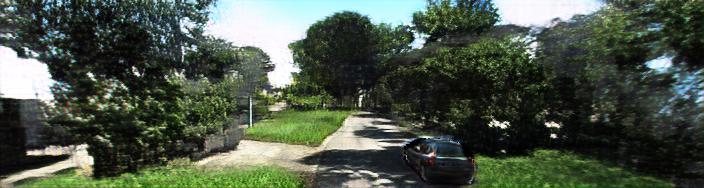} \vspace{\reduceheight}\\
   \begin{footnotesize}Road to grass\end{footnotesize} &
   \begin{footnotesize}Building to tree\end{footnotesize} &
   \begin{footnotesize}Lower building\end{footnotesize} &
    \begin{footnotesize}Moving Tree\end{footnotesize} \vspace{-0.25cm}\\
  \end{tabular}}\\
  \subfloat[][Object Editing\label{fig:object_editing}\vspace{-0.3cm}]{
       \begin{tabular}{cccc}
      \includegraphics[width=\mywidth\linewidth]{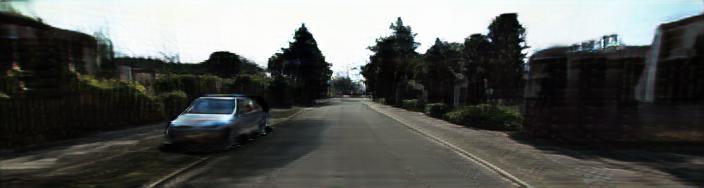} &
      \includegraphics[width=\mywidth\linewidth]{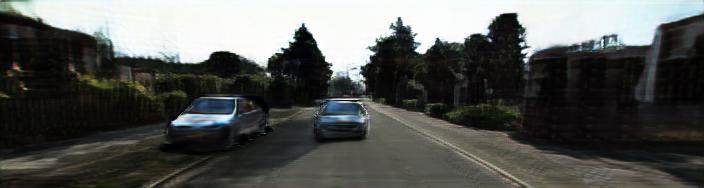} &
      \includegraphics[width=\mywidth\linewidth]{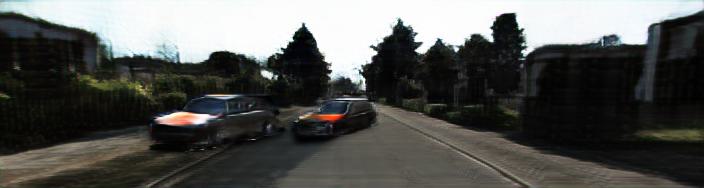} &
      \includegraphics[width=\mywidth\linewidth]{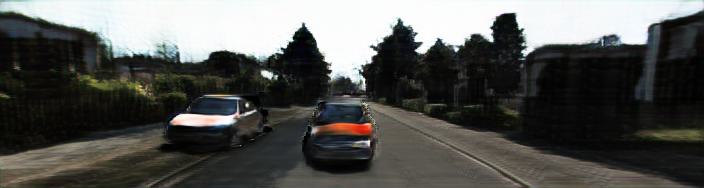} 
      \vspace{\reduceheight}\\
      \includegraphics[width=\mywidth\linewidth]{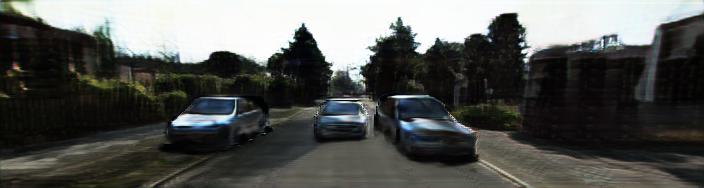} &
      \includegraphics[width=\mywidth\linewidth]{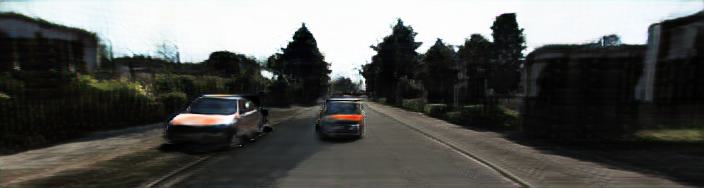} &
      \includegraphics[width=\mywidth\linewidth]{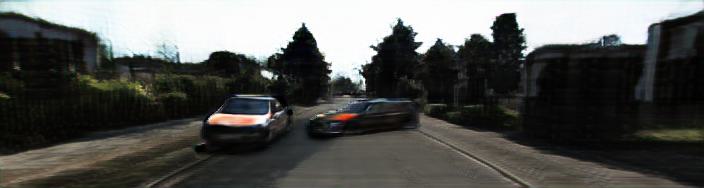} &
      \includegraphics[width=\mywidth\linewidth]{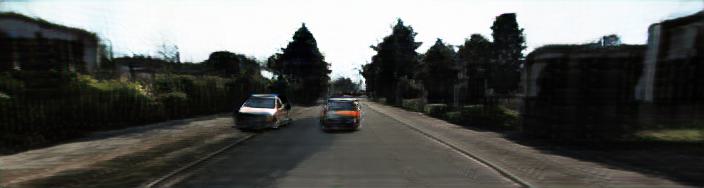} 
      \vspace{\reduceheight}\\
      \begin{footnotesize} Removal / Insertion\end{footnotesize} &
      \begin{footnotesize}Appearance \end{footnotesize} &
      \begin{footnotesize}Rotation\end{footnotesize} &
      \begin{footnotesize}Translation\end{footnotesize}  \vspace{-0.25cm}\\
     \end{tabular}\vspace{-0.15cm}}\\
  \subfloat[][Camera Control\label{fig:camera_control}\vspace{-0.2cm}]{
     \begin{tabular}{cccc}
      \includegraphics[width=\mywidth\linewidth]{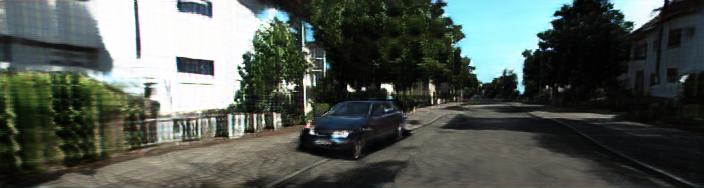} &
      \includegraphics[width=\mywidth\linewidth]{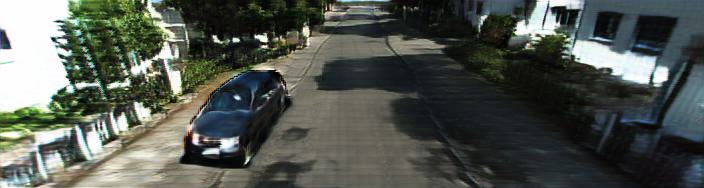} &
      \includegraphics[width=\mywidth\linewidth]{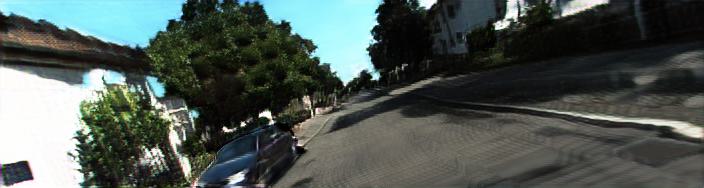} &
      \includegraphics[width=\mywidth\linewidth]{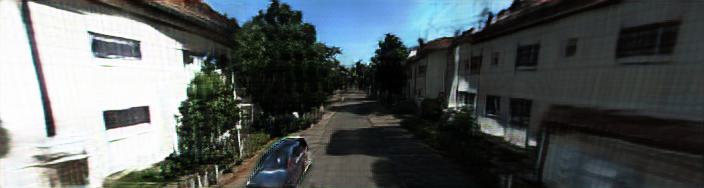} 
      \vspace{\reduceheight}\\
      \begin{footnotesize}Azimuth angle\end{footnotesize} &
      \begin{footnotesize}Polar angle \end{footnotesize} &
      \begin{footnotesize}In-plane rotation \end{footnotesize} &
      \begin{footnotesize}Changing FoV\end{footnotesize}  \vspace{-0.25cm}\\
     \end{tabular}\vspace{-0.15cm}}
  \caption{{\bf Controllable 3D-aware Image Synthesis} on KITTI-360.} 
  \label{fig:full_control}
  \vspace{-0.3cm}
 \end{figure*}

%% file: gfx/results/ablation_kitti360.tex
\begin{figure*}[htbp]
     \centering
     \setlength{\tabcolsep}{1pt}
     \def\mywidth{.25}
     \def\reduceheight{-3.5pt}
     \begin{tabular}{cccc}
      \includegraphics[width=\mywidth\linewidth]{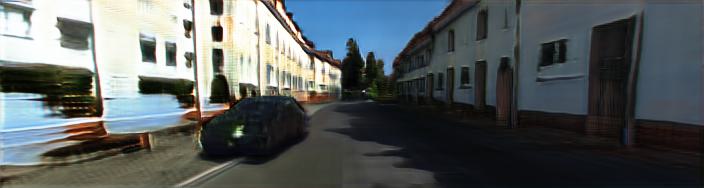} &
      \includegraphics[width=\mywidth\linewidth]{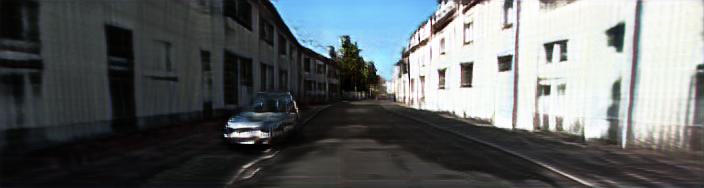} &
      \includegraphics[width=\mywidth\linewidth]{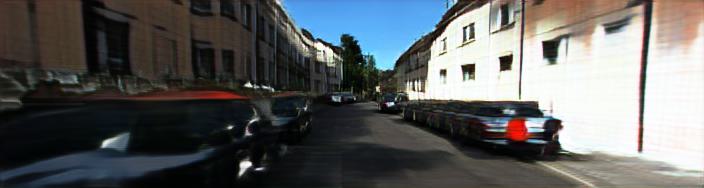} &
      \includegraphics[width=\mywidth\linewidth]{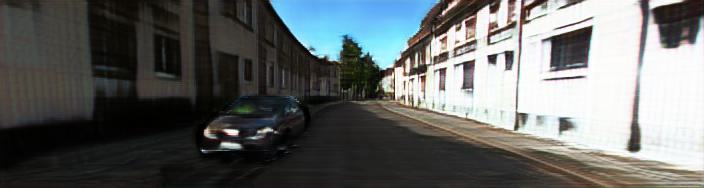} \vspace{\reduceheight}\\

      \includegraphics[width=\mywidth\linewidth]{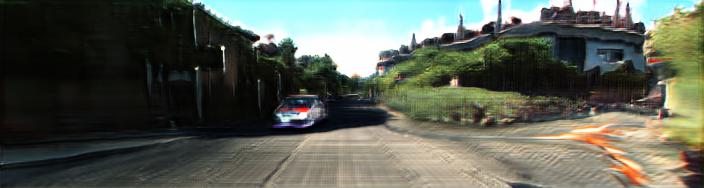} &
      \includegraphics[width=\mywidth\linewidth]{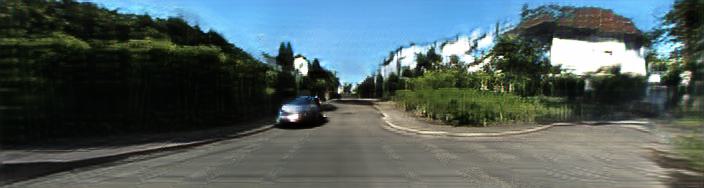} &
      \includegraphics[width=\mywidth\linewidth]{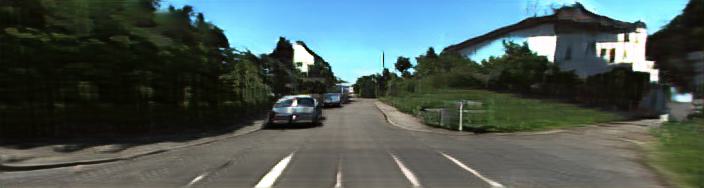} &
      \includegraphics[width=\mywidth\linewidth]{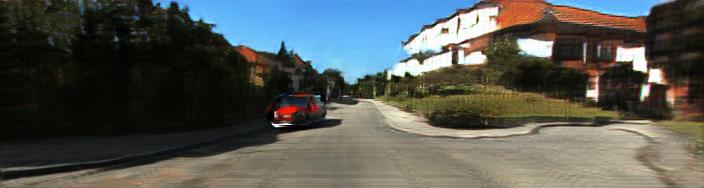} \vspace{\reduceheight}\\

      \includegraphics[width=\mywidth\linewidth]{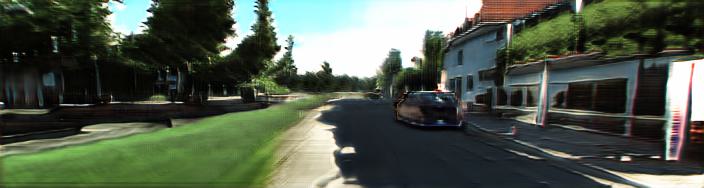} &
      \includegraphics[width=\mywidth\linewidth]{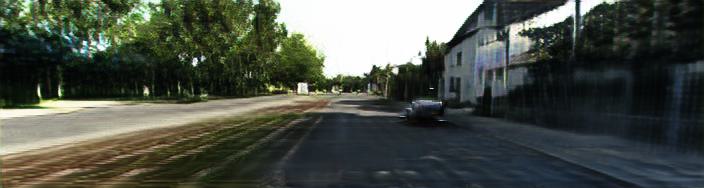} &
      \includegraphics[width=\mywidth\linewidth]{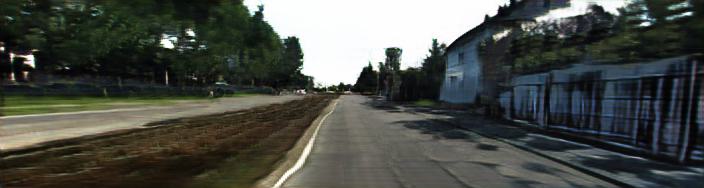} &
      \includegraphics[width=\mywidth\linewidth]{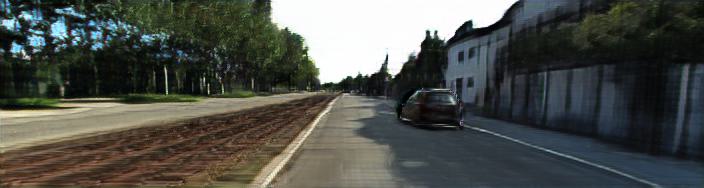} \\
      \begin{small}w/o $\cL_{recon}$\end{small} &
      \begin{small}w/o $\cL_{adv}^{\bP}$\end{small} &
      \begin{small}w/o $G_\theta^{obj}$\end{small} &
      \begin{small}Full Model\end{small} 
     \end{tabular}\vspace{-0.2cm}
     \caption{{{\bf Ablation Study}. Each row shows synthesized images given the same panoptic prior in KITTI-360 with different method variations. Removing the reconstruction loss (w/o $\cL_{recon}$) leads to more artifacts, and the semantic condition may not be preserved (3rd row). Removing the object discriminator (w/o $\cL_{adv}^{\bP}$) and modeling all objects as stuff (w/o $G_\theta^{obj}$) both impede the fidelity of objects (cars). }  
     }
     \vspace{-0.2cm}
     \label{fig:ablation}
    \end{figure*}

%% file: gfx/results/comp_gt_kitti360.tex
\begin{figure}[htbp!]
     \centering
     \setlength{\tabcolsep}{1pt}
     \def\mywidth{4cm}  
     \def\reduceheight{-4.5pt}
     \begin{tabular}{P{0.3cm}P{\mywidth}P{\mywidth}}
      \rotatebox{90}{\scriptsize Ours} & 
      \includegraphics[width=\mywidth]{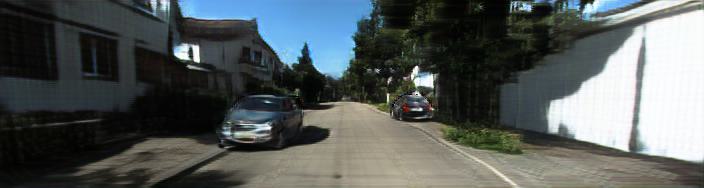} &
      \includegraphics[width=\mywidth]{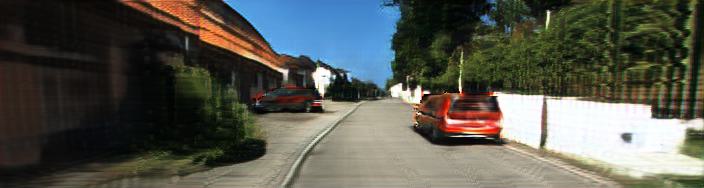}\\ 
      \rotatebox{90}{\scriptsize GT} &
      \includegraphics[width=\mywidth]{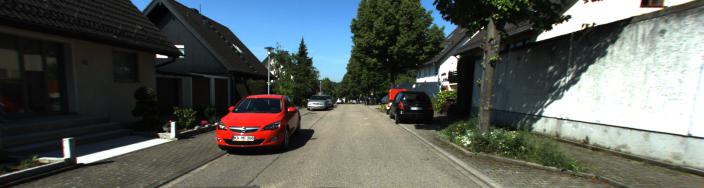} &
      \includegraphics[width=\mywidth]{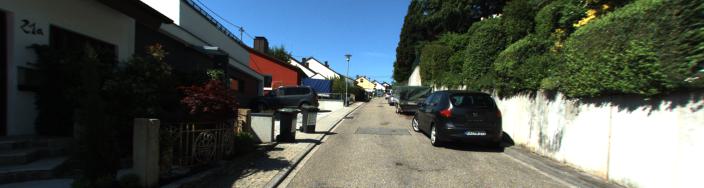} \\
     \end{tabular}\vspace{-0.1cm}
     \caption{{\bf Diversity}. We compare our synthesized images to the corresponding ground truth image with the same panoptic prior. Note that our method keeps the same layout but maintains diversity. }
     \label{fig:ablation_comp_gt}
     \vspace{-0.4cm}
    \end{figure}

%% file: gfx/results/tables/ablation_comp.tex
\begin{table}[]
    \centering
    \small
    \begin{tabularx}{\linewidth}{lXXXX}
    \toprule
    & $\text{FID}_{\bI}\downarrow$                  & $\text{KID}_{\bI}\downarrow$ & $\text{FID}_{\bP}\downarrow $ & $\text{KID}_{\bP}\downarrow$             \\ \hline
    w/o $\cL_{recon}$                        &    89.3                  &  0.067   &  77.6        &   0.062          \\ 
    w/o $\cL_{adv}^{\bP}$                       &  53.1                &  0.050   &    119.0      & 0.120                \\ 
    w/o $G_\theta^{obj}$                       &    44.7                  & 0.036    &   --       &       --               \\ 
    $\text{Full}$                            & \textbf{39.0}                  &\textbf{0.036}    &  \textbf{67.1}        &   \textbf{0.056}              \\
    \bottomrule   
    \end{tabularx}
    \caption{{\bf Ablation Study.} We report FID and KID of UrbanGIRAFFE on KITTI-360 without reconstruction loss (w/o $\cL_{recon}$), without object discriminator (w/o $\cL_{adv}^{\bP}$) and without object generator ( w/o $G_\theta^{obj}$)}
    \label{tab:ablation_study}
    \end{table}

%% file: tex/sec_5_conclusion.tex
\section{Conclusion}
\label{sec:conclusion}

We propose UrbanGIRAFFE to tackle controllable 3D-aware image synthesis for challenging urban scenes. By effectively incorporating 3D panoptic prior, our model decomposes the scene into stuff, objects, and sky. Our compositional generative model enables diverse controllability regarding large camera viewpoint change, semantic layout, and object manipulation. We believe that our method pushes the frontier of 3D-aware generative models for unbounded scenes with complex geometry. In future work, it can be augmented with a semantic voxel generator for sampling novel scenes. Further, our method does not disentangle light from ambient color, which is worth investigating to enable lighting control.

%% file: tex/sec_6_supp.tex
\section{Implementation Details}
\label{sec:implementation}
\subsection{Object Patch Rendering} 
We render object patches taking occlusions into consideration. First, this allows us to directly crop the object patches from the full composited image without additional computation to render the complete objects. Second, we can keep occluded object patches in the real images for training, as filtering out occluded objects is not trivial.  
More specifically, we first obtain the alpha value of a ray corresponding to the $k$th object via volume rendering:
\begin{gather}
\label{eq:obj_alpha}
    \mathrm{A}_{obj}^k =\sum_{i=1}^{N} T_i \alpha_i \mathbbm{1}[\bx_i \in \{\bR(\bs \odot \bx_{obj}^k)+ \bt\}]
\end{gather}
This means  $T_i \alpha_i$ is accumulated if the corresponding sampled point $\bx_i$ belongs to the $k$th object. 
Here, we use the transmittance $T_i$ of the composited scene obtained via Eq. 7 of the main paper, meaning that the alpha value is close to $0$ if the object is occluded. 
Let $\bA^k_{obj}\in\nR ^{H_f\times W_f}$ denote the alpha map consisting of object alpha values of all rays. Note that $\bA^k_{obj}$ is obtained via volume rendering, and thus its resolution is lower than the final output image $\hat{\bI}\in\nR^{H\times W \times 3}$. Thus, we upsample $\bA^k_{obj}$ via nearest neighbor sampling to obtain the object patches from $\hat{\bI}$:
\begin{equation}
\hat{\bP}_k=crop(\hat{\bI} \odot  up(\bA_{obj}^k)) 
\end{equation}
where $up(\cdot)$ denotes nearest neighbor upsampling and $crop(\cdot)$ denotes cropping the object based on its projected 3D bounding box.

\subsection{Network Architecture}

\boldparagraph{Generator Architecture} For the latent codes, we use a 256-dimension $\bz_{wld}$ for the entire scene and a 256-dimension $\bz^k_{obj}$ for each object.
Our \textit{stuff generator} consists of a semantic-conditioned feature grid generator and an MLP head. The feature grid generator $G_\theta^{vol}$ consists of 5 spatially-adaptive normalization blocks. Each block follows the structure of a ResNet~\cite{He2016CVPR} block with two convolutional layers, except that the batch normalization is replaced with spatial-adaptive normalization modulated by the semantic labels. As illustrated in Fig. 3 of the main paper, we inject the latent code  $\bz_{wld}$ at each block following~\cite{Schonfeld2021ICLR}. The MLP head  $G_\theta^{stf}$ is a 4-layer ReLU MLP with a hidden dimension of 256. The \textit{object generator} $G_\theta^{obj}$ is an 8-layer ReLU MLP but with a lower dimension of 128. We use skip connections at the fourth layer for $G_\theta^{obj}$. For the sky generator, we use a 5-layer ReLU MLP of hidden dimension 256 without a skip connection. As mentioned in the main paper, we apply positional encoding $\gamma(\cdot)$ to both $\bx^k_{obj}$ and $\bx_{wld}$:
\begin{equation}
\gamma(p) = (sin(2^0\pi p), cos(2^0\pi p), sin(2^1\pi p), cos(2^1\pi p),  \cdots, sin(2^{L-1}\pi p), cos(2^{L-1}\pi p) 
\end{equation}
where $\gamma(p)$ is applied to each element of the coordinate. 
We use $L=10$ for both $\bx^k_{obj}$ as input to the object generator and $\bx_{wld}$ as input to the stuff generator.

Our \textit{2D neural renderer} $\pi^{neural}_{\theta}$ consists of two blocks of StyleGAN2-modulated convolutional blocks and one upsampling layer. In practice, we render the feature maps $\bI_{\bF}\in\nR^{H_f\times W_f \times M_f}$ at half resolution and upsample it to image $\hat{\bI}\in\nR^{H\times W \times 3}$  at the target resolution, \ie, $H_f=H/2, W_f=W/2$. %

\boldparagraph{Discriminator Architecture} We adopt two independent discriminators to apply adversarial losses to the  composited images and the object patches, respectively. Both discriminators follow the design choice of the StyleGAN2 discriminator, while the object discriminator takes a lower-resolution image as input and thus has fewer parameters. More specifically, all object patches are rescaled to $128 \times 128$ pixels, and the object discriminator $D_\phi^{\bP}$ has 6 convolutional blocks with 5 downsampling layers. The discriminator of the full image $D_\phi^{\bI}$ has 8 convolutional blocks with 7 downsampling layers.

\subsection{Training and Inference}
We train our model on four Nvidia GeForce RTX 3090 with a batch size of 16 for 100k iterations, taking 4 days in total. For inference, our method can render an image at the resolution of $188\times 704$ at roughly 5 FPS.

\section{Baselines}
\label{sec:baseline}

\boldparagraph{GIRAFFE} We follow the original implementation\footnote{\url{https://github.com/autonomousvision/giraffe}} of GIRAFFE~\cite{Niemeyer2021CVPR}. For the KITTI-360 dataset, we sample objects following the same object layout prior as used in our method. We also sample points within the objects using the same ray-box intersection strategy for a fair comparison. We train GIRAFFE on four Nvidia GeForce RTX 3090 with a batch size of 16 for 140k iterations.

\boldparagraph{GSN} We use the official  implementation\footnote{\url{https://github.com/apple/ml-gsn}} of GSN~\cite{DeVries2021ICCV}. 
GSN generates a scene based on a 2D grid of local latent codes. Following the original implementation, the spatial resolution of the 2D grid is set to $32\times 32$. For the KITTI-360 dataset, we set the maximum sampling distance as $80m$, with each pixel in the 2D grid corresponding to a region of $2.5m \times 2.5m$ in reality. Because of the limitation of computational resources, we sample 32 points per ray instead of 64. We train GSN for 400k iterations with a batch size of 4 on two 3090 GPUs and perform gradient accumulation every two iterations. 

\boldparagraph{2D Baseline} We evaluate StyleGAN2 as a state-of-the-art 2D GAN following its PyTorch implementation\footnote{\url{https://github.com/rosinality/stylegan2-pytorch}}. We train the 2D baseline on four Nvidia GeForce RTX 3090 with a batch size of 32 for 120k iterations, taking 2 days in total.

\section{Dataset}
\label{sec:dataset}

\boldparagraph{KITTI-360} We adopt the KITTI-360 dataset to evaluate our method on urban scenes. For the real object (\ie, cars) patches, we filter out heavily occluded and far away objects based on the following criteria: 1) The pixel number of the object is larger than a given threshold; 2) The projected 3D bounding box has at least four visible vertices. Note that this does not filter out all occluded objects. We use instance masks provided by KITTI-360 to crop the object patches for training. These masks may be noisy as they are obtained via label transfer algorithms instead of manually labeled. Therefore, using instance masks predicted by 2D segmentation methods might be possible. In order to obtain training images containing enough objects, we keep one image only when it contains one or more than one valid object patches for training. This leads to a total of 20k training images. 

We train and evaluate our method at a half resolution of KITTI-360, \ie, $188\times 704$ pixels.  The same resolution is applied to the other baselines except for GSN, which requires large memory consumption and does not scale to the target resolution. Therefore, we train and evaluate GSN at the image resolution of $94 \times 352$ pixels. 

In all of our experiments, we use voxel grids at the resolution of $64 \times 64 \times 64$ voxels. Note that the sampling interval of the semantic voxel is $1 m$ horizontally and $0.25 m$ vertically. Therefore, each semantic voxel grid covers an area of $4096 m^2$ with a height of $16 m$ in the real world. 

\begin{figure}[t]
  \centering
   \includegraphics[width=0.6\linewidth]{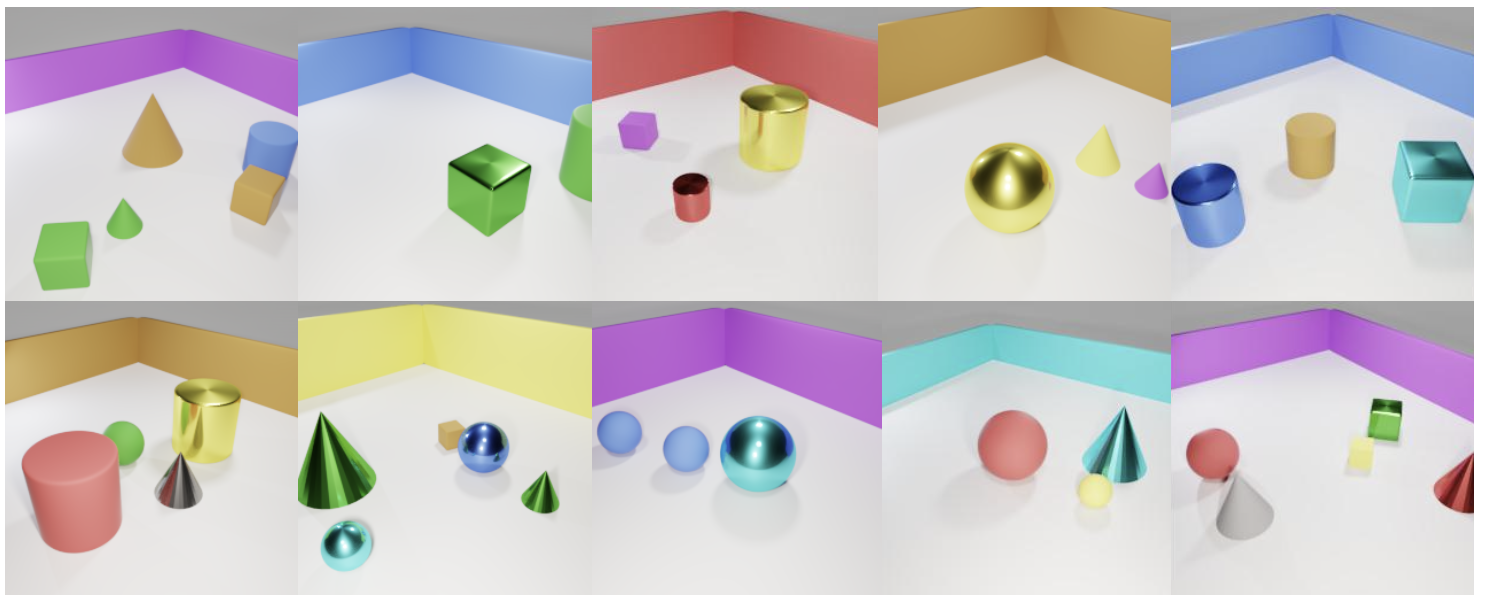}
   \caption{\textbf{Preview of CLEVR-W}. We add walls with randomly sampled colors and locations and render a set of images based on the rendering script of CLEVR~\cite{Johnson2017CVPR}.}
   \label{fig:Clevr}
\end{figure}

\boldparagraph{CLEVR-W} We further create a dataset CLEVR-W to facilitate comparison with GIRAFFE~\cite{Niemeyer2021CVPR} in more controlled environments. The dataset contains 10k images rendered at the resolution of $256\times 256$. In contrast to the dataset used in GIRAFFE, we add walls as stuff regions and thus increasing the difficulty of modeling the background. The walls are sampled at different locations with random colors, see \figref{fig:Clevr} for a preview.

\section{Additional Experimental Results}
\label{sec:result}
\subsection{Additional Comparison to Baselines}
\boldparagraph{KITTI-360}
In \figref{fig:comp_baselines_kitti360}, we show additional  comparison results to the baselines on KITTI-360, where each column shows a single scene with the camera consecutively moving forward for up to 20 meters. Note that the background regions of GIRAFFE barely change despite the large camera movement, since GIRAFFE models the background as a far-away planar structure in the challenging urban scenario. GSN can model the camera movement faithfully but has lower image fidelity, especially when synthesizing an image with a large camera moving distance. In contrast, our method is able to synthesize high-fidelity images along the large camera moving distance. %

\input{gfx_supp/comp_kitti360}

\boldparagraph{CLEVR-W}
In \figref{fig:comp_baselines_clevr}, we show additional comparisons to GIRAFFE on CLEVR-W. Our method enables control over objects, stuff, and the camera pose.
\input{gfx_supp/comp_clevr}

\input{gfx_supp/failure}

\subsection{Additional Results on Controllable Image Synthesis}

\figref{fig:editing} shows additional scene editing results on KITTI-360, including stuff editing (including lower building in \figref{fig:building2tree}, building to tree in \figref{fig:lowerbuilding}, road to grass in \figref{fig:road2grass}) and object editing (\figref{fig:object_editing}).

\input{gfx_supp/editing_kitti360}

\subsection{Limitations}
\figref{fig:fail} illustrates two limitations of our method. First, we observe in \figref{fig:z_fail} that changing $\bz_{wld}$ does not change the appearance of the stuff significantly. In contrast, we observe that editing the semantic layout can sometimes change the appearance of the scene. This is rational as the appearance is entangled with the semantic layout in the real world, \eg, to model shadows.  
\figref{fig:sky_fail} shows that our sky generator sometimes generates far-away buildings and thus yields artifacts. This is due to the fact that our semantic voxel grids can only model a region of $64 \times 64$ square meters.

\subsection{Random Samples}

We show randomly sampled, uncurated results on both KITTI-360 and CLEVR-W in \figref{fig:random}. 
\input{gfx_supp/random}

%% file: gfx_supp/comp_kitti360.tex
\begin{figure*}[htbp!]
     \centering
     \setlength{\tabcolsep}{0pt}
     \def\mywidth{.2}
    \def\reduceheight{-3.5pt}

     \subfloat[][GIRAFFE\label{fig:giraffe_kitti360}\vspace{-0.2cm}]{
     \begin{tabular}{ccccc}

          \includegraphics[width=\mywidth\linewidth]{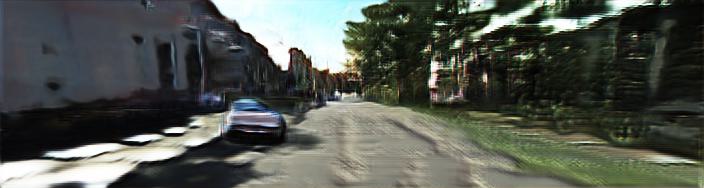}&
          \includegraphics[width=\mywidth\linewidth]{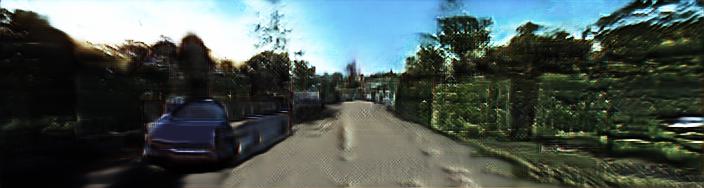}&
          \includegraphics[width=\mywidth\linewidth]{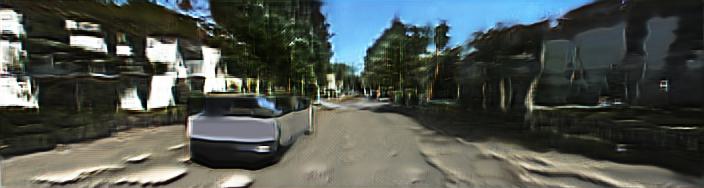}&
          \includegraphics[width=\mywidth\linewidth]{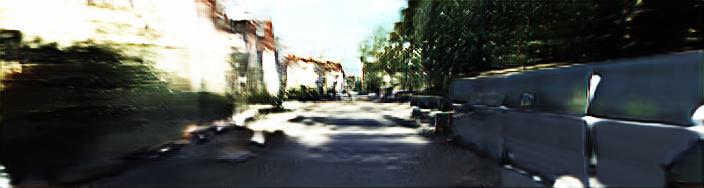}&
          \includegraphics[width=\mywidth\linewidth]{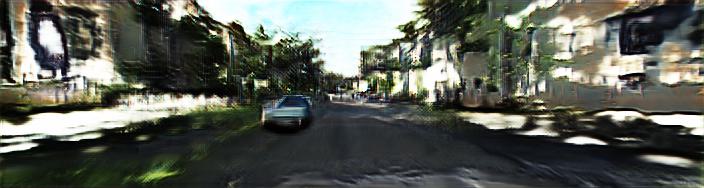}  \vspace{\reduceheight}\\
       
          \includegraphics[width=\mywidth\linewidth]{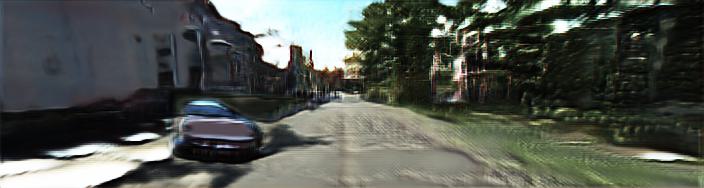}&
          \includegraphics[width=\mywidth\linewidth]{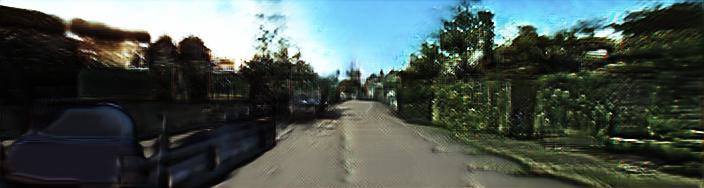}&
          \includegraphics[width=\mywidth\linewidth]{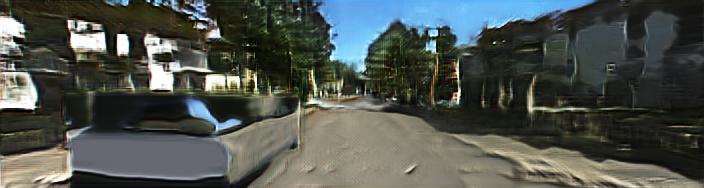}&
          \includegraphics[width=\mywidth\linewidth]{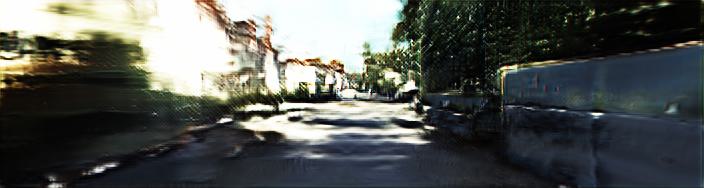}&
          \includegraphics[width=\mywidth\linewidth]{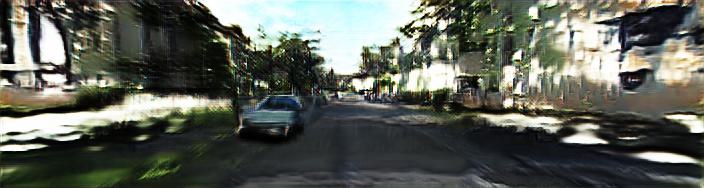} \vspace{\reduceheight}\\
       
          \includegraphics[width=\mywidth\linewidth]{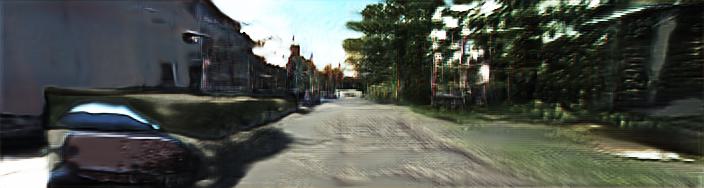}&
          \includegraphics[width=\mywidth\linewidth]{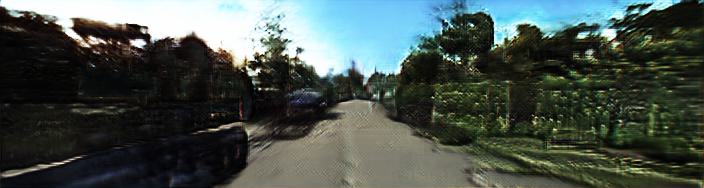}&
          \includegraphics[width=\mywidth\linewidth]{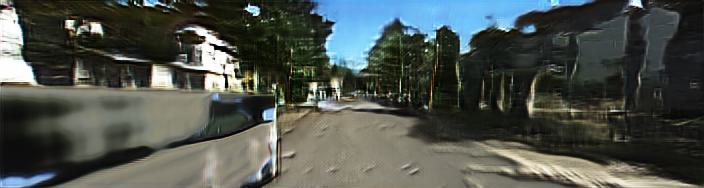}&
          \includegraphics[width=\mywidth\linewidth]{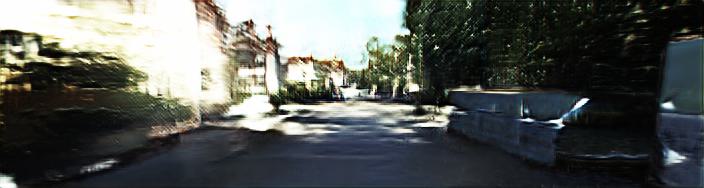}&
          \includegraphics[width=\mywidth\linewidth]{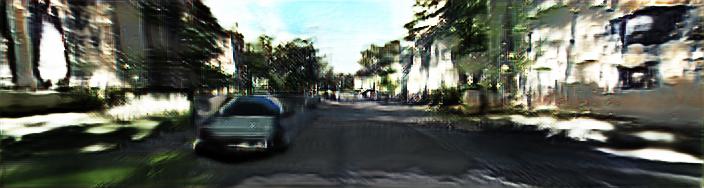} \vspace{\reduceheight}\\
       
          \includegraphics[width=\mywidth\linewidth]{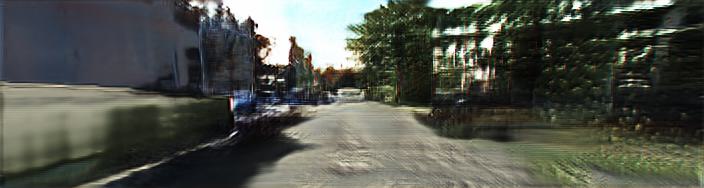}&
          \includegraphics[width=\mywidth\linewidth]{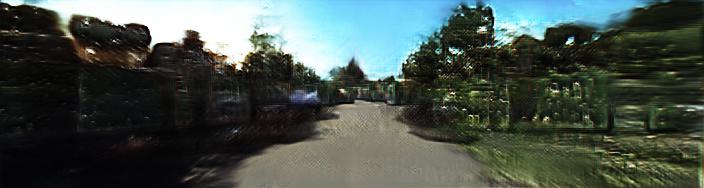}&
          \includegraphics[width=\mywidth\linewidth]{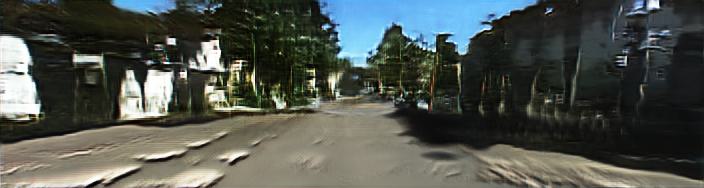}&
          \includegraphics[width=\mywidth\linewidth]{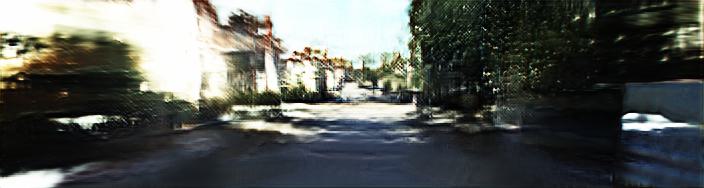}&
          \includegraphics[width=\mywidth\linewidth]{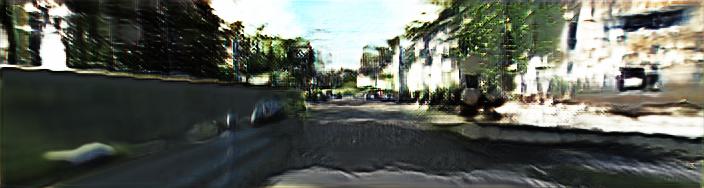} \vspace{\reduceheight}\\
       
          \includegraphics[width=\mywidth\linewidth]{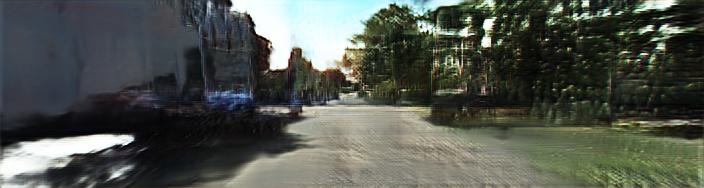}&
          \includegraphics[width=\mywidth\linewidth]{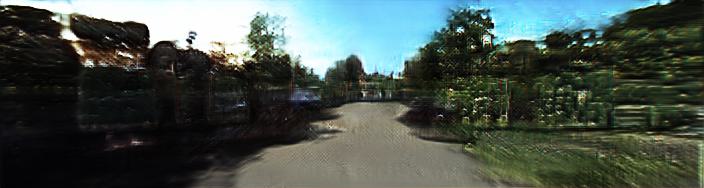}&
          \includegraphics[width=\mywidth\linewidth]{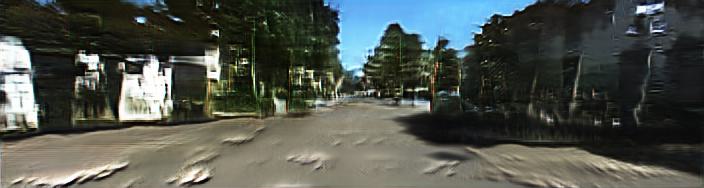}&
          \includegraphics[width=\mywidth\linewidth]{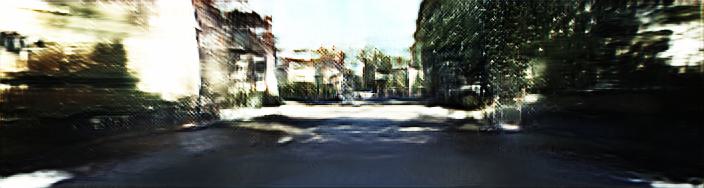}&
          \includegraphics[width=\mywidth\linewidth]{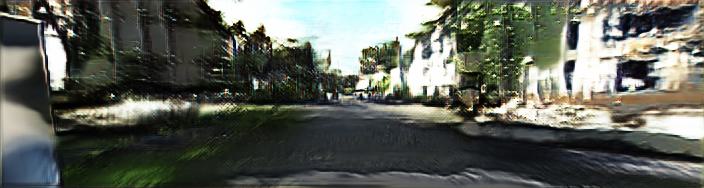} \vspace{\reduceheight}\\
       
          \includegraphics[width=\mywidth\linewidth]{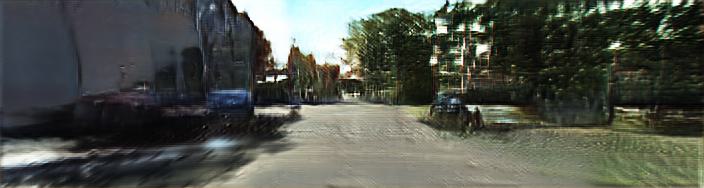}&
          \includegraphics[width=\mywidth\linewidth]{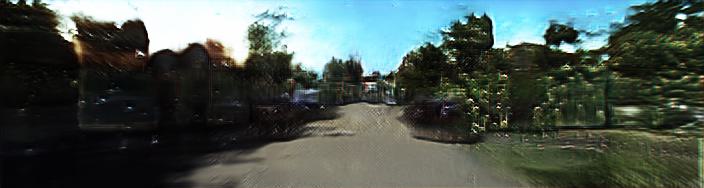}&
          \includegraphics[width=\mywidth\linewidth]{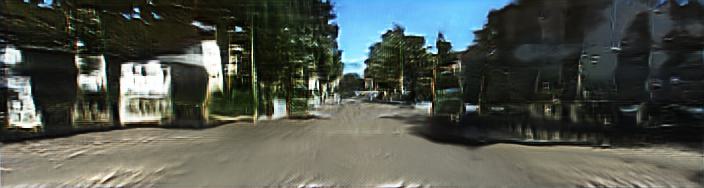}&
          \includegraphics[width=\mywidth\linewidth]{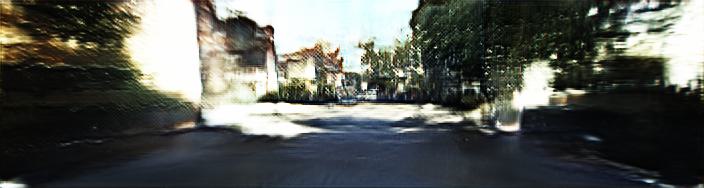}&
          \includegraphics[width=\mywidth\linewidth]{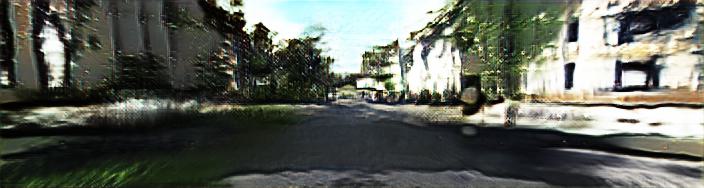}
          \vspace{\reduceheight}\\
     \end{tabular}\vspace{-0.1cm}}

     \subfloat[][GSN\label{fig:gsn_kitti360}\vspace{-0.2cm}]{
     \begin{tabular}{ccccc}

               \includegraphics[width=\mywidth\linewidth]{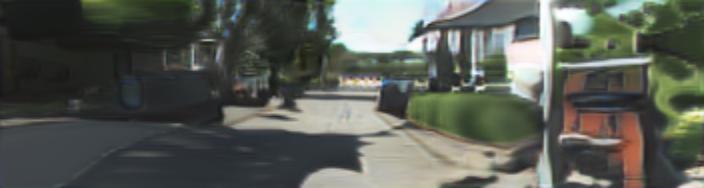}&
               \includegraphics[width=\mywidth\linewidth]{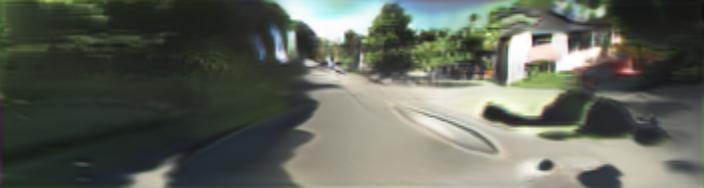}&
               \includegraphics[width=\mywidth\linewidth]{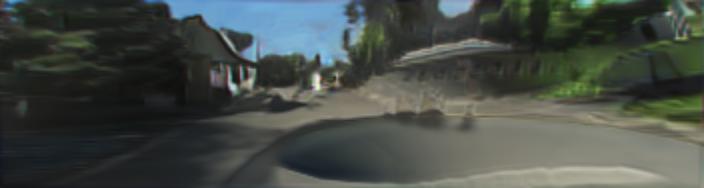}&
               \includegraphics[width=\mywidth\linewidth]{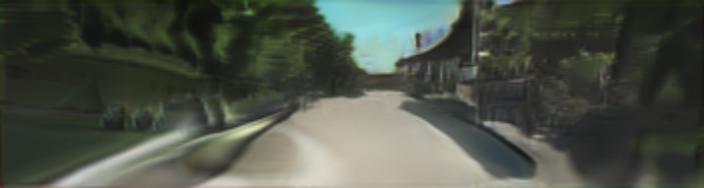}&
               \includegraphics[width=\mywidth\linewidth]{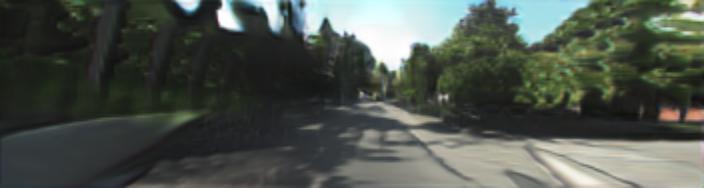}  \vspace{\reduceheight}\\
         
               \includegraphics[width=\mywidth\linewidth]{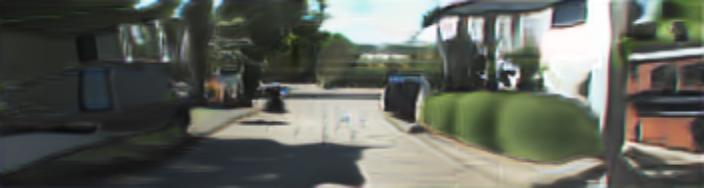}&
               \includegraphics[width=\mywidth\linewidth]{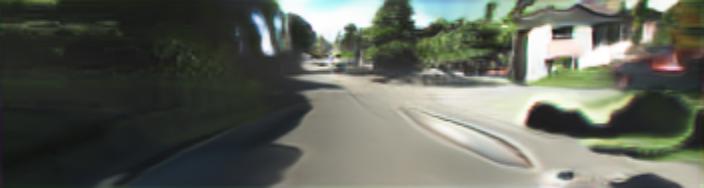}&
               \includegraphics[width=\mywidth\linewidth]{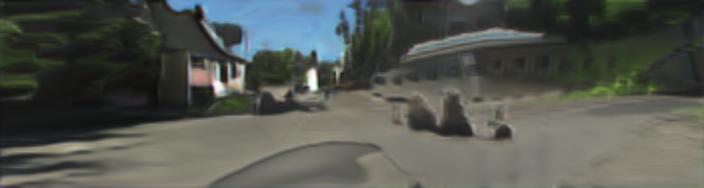}&
               \includegraphics[width=\mywidth\linewidth]{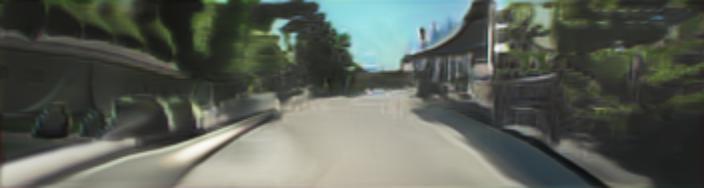}&
               \includegraphics[width=\mywidth\linewidth]{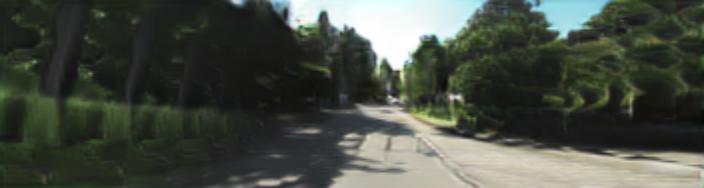} \vspace{\reduceheight}\\
         
               \includegraphics[width=\mywidth\linewidth]{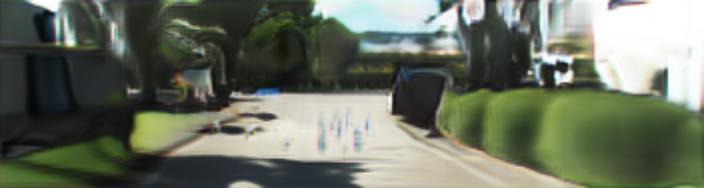}&
               \includegraphics[width=\mywidth\linewidth]{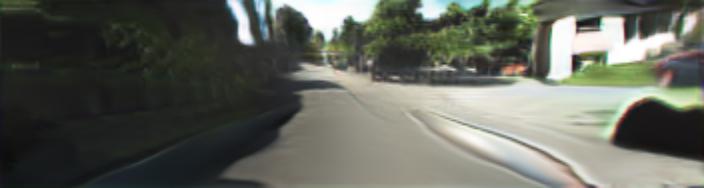}&
               \includegraphics[width=\mywidth\linewidth]{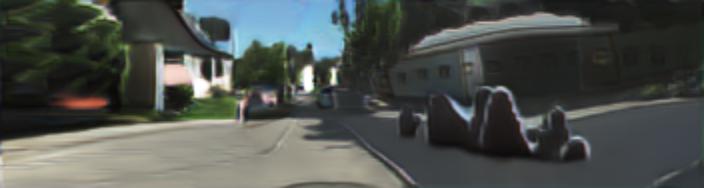}&
               \includegraphics[width=\mywidth\linewidth]{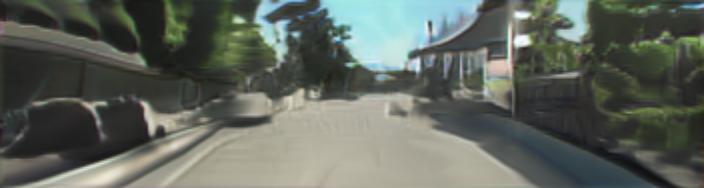}&
               \includegraphics[width=\mywidth\linewidth]{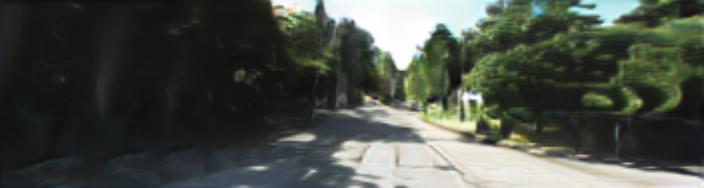} \vspace{\reduceheight}\\
         
               \includegraphics[width=\mywidth\linewidth]{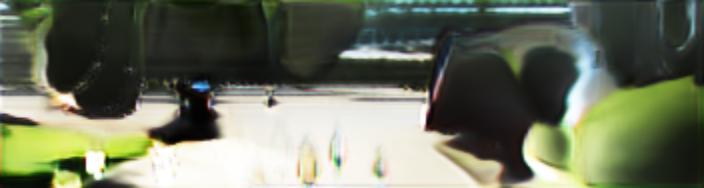}&
               \includegraphics[width=\mywidth\linewidth]{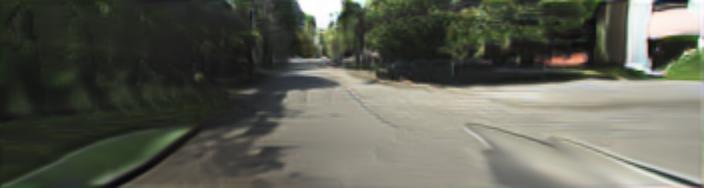}&
               \includegraphics[width=\mywidth\linewidth]{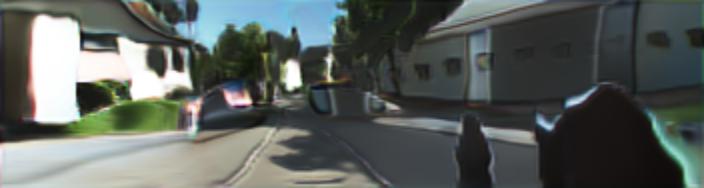}&
               \includegraphics[width=\mywidth\linewidth]{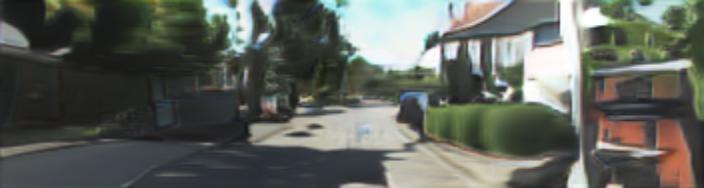}&
               \includegraphics[width=\mywidth\linewidth]{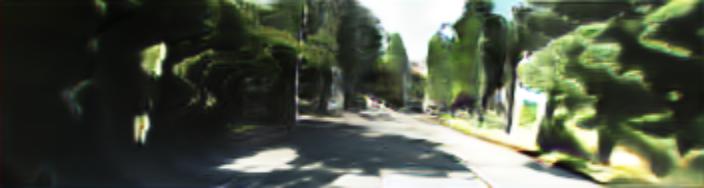} \vspace{\reduceheight}\\
         
               \includegraphics[width=\mywidth\linewidth]{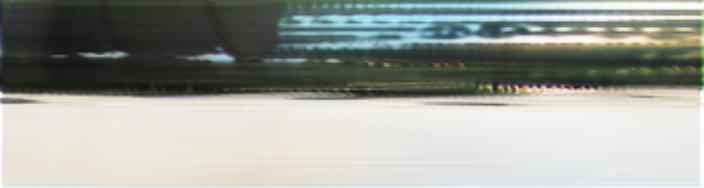}&
               \includegraphics[width=\mywidth\linewidth]{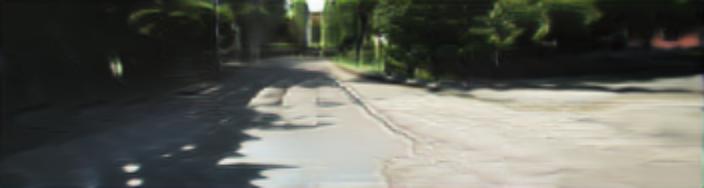}&
               \includegraphics[width=\mywidth\linewidth]{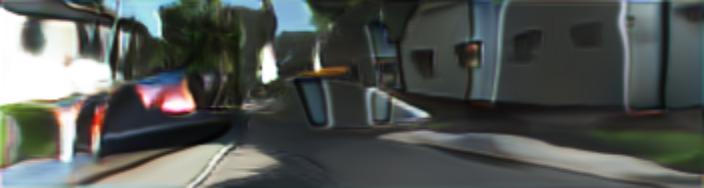}&
               \includegraphics[width=\mywidth\linewidth]{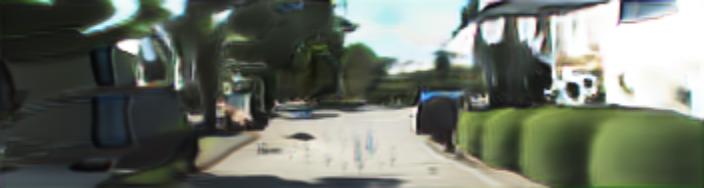}&
               \includegraphics[width=\mywidth\linewidth]{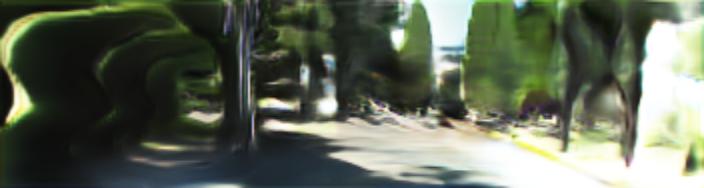} \vspace{\reduceheight}\\
         
               \includegraphics[width=\mywidth\linewidth]{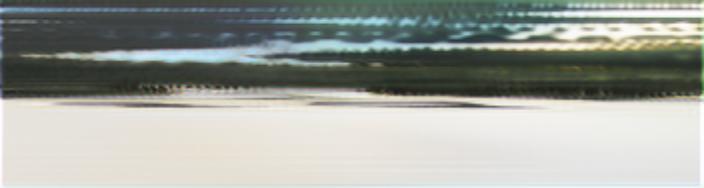}&
               \includegraphics[width=\mywidth\linewidth]{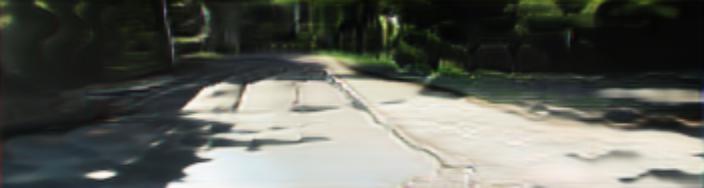}&
               \includegraphics[width=\mywidth\linewidth]{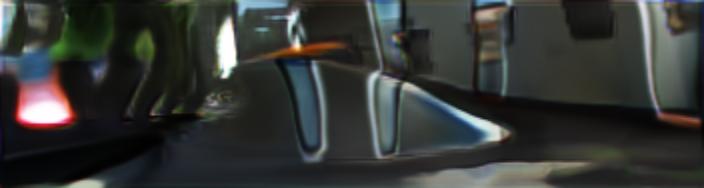}&
               \includegraphics[width=\mywidth\linewidth]{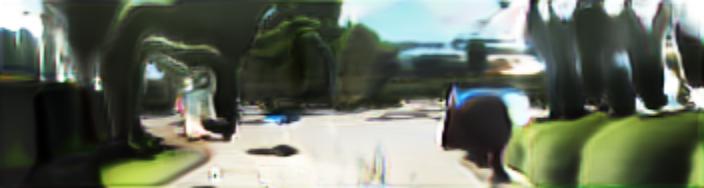}&
               \includegraphics[width=\mywidth\linewidth]{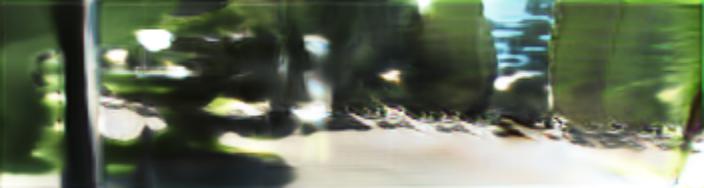}
               \vspace{\reduceheight}\\
     \end{tabular}\vspace{-0.1cm}
     }

     \subfloat[][UrbanGIRAFFE (Ours)\label{fig:ours_kitti360}\vspace{-0.2cm}]{
          \begin{tabular}{ccccc}
        
           \includegraphics[width=\mywidth\linewidth]{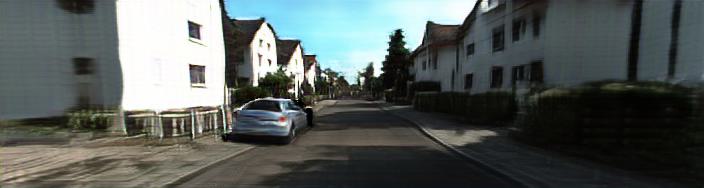}&
           \includegraphics[width=\mywidth\linewidth]{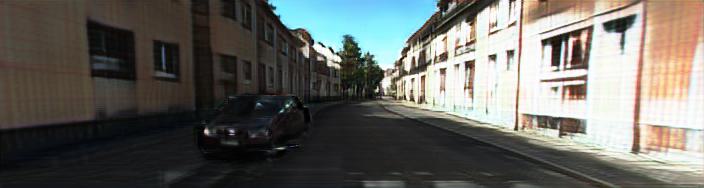}&
           \includegraphics[width=\mywidth\linewidth]{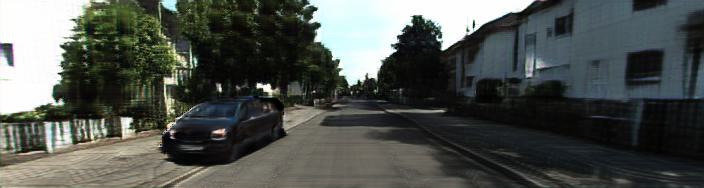}&
           \includegraphics[width=\mywidth\linewidth]{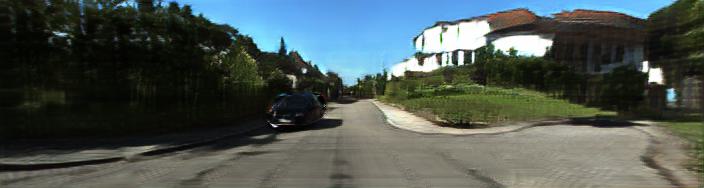}&
           \includegraphics[width=\mywidth\linewidth]{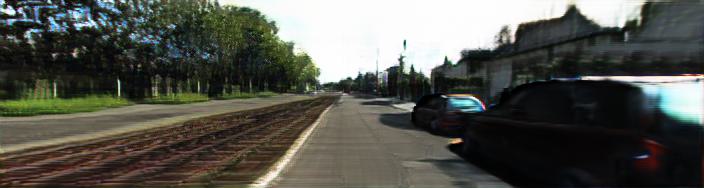}  \vspace{\reduceheight}\\
        
           \includegraphics[width=\mywidth\linewidth]{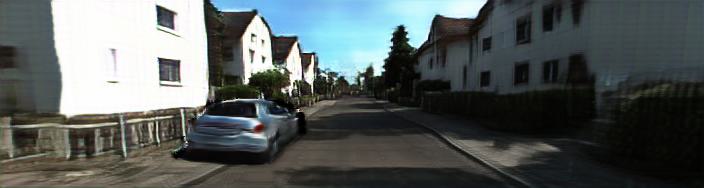}&
           \includegraphics[width=\mywidth\linewidth]{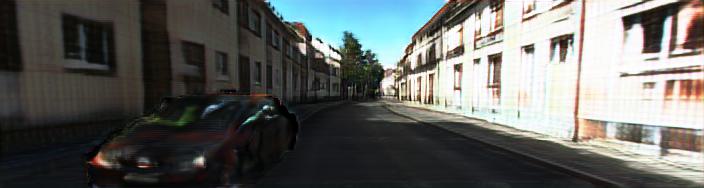}&
           \includegraphics[width=\mywidth\linewidth]{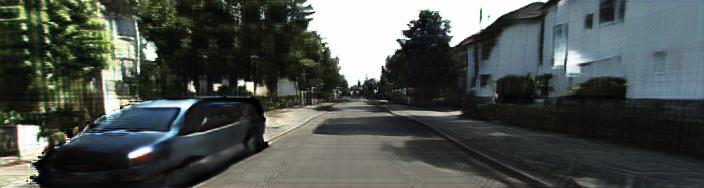}&
           \includegraphics[width=\mywidth\linewidth]{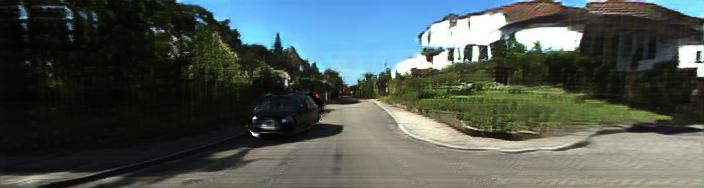}&
           \includegraphics[width=\mywidth\linewidth]{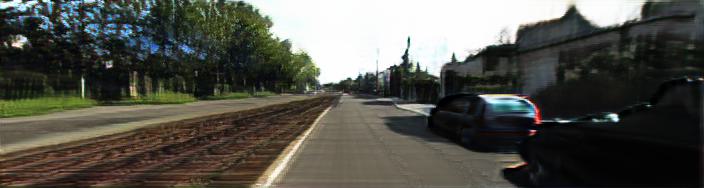} \vspace{\reduceheight}\\
        
           \includegraphics[width=\mywidth\linewidth]{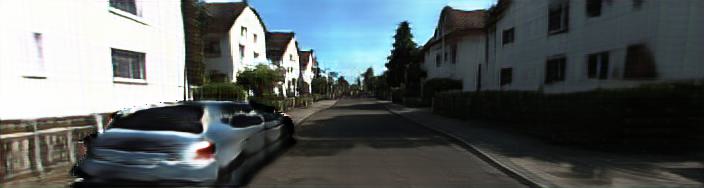}&
           \includegraphics[width=\mywidth\linewidth]{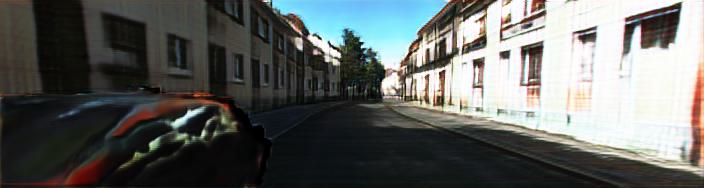}&
           \includegraphics[width=\mywidth\linewidth]{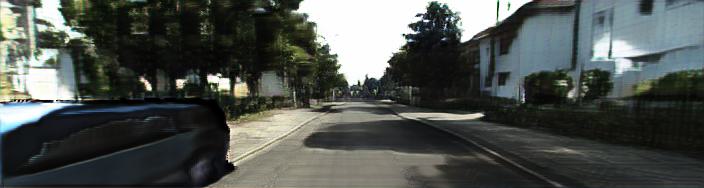}&
           \includegraphics[width=\mywidth\linewidth]{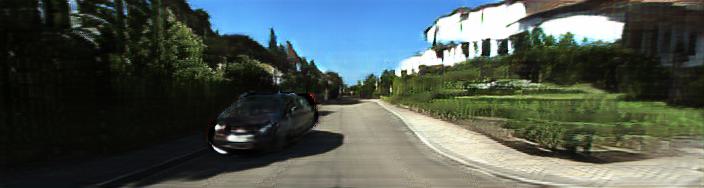}&
           \includegraphics[width=\mywidth\linewidth]{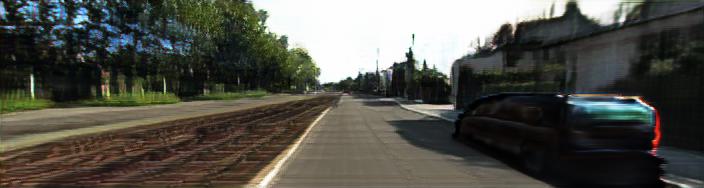} \vspace{\reduceheight}\\
        
           \includegraphics[width=\mywidth\linewidth]{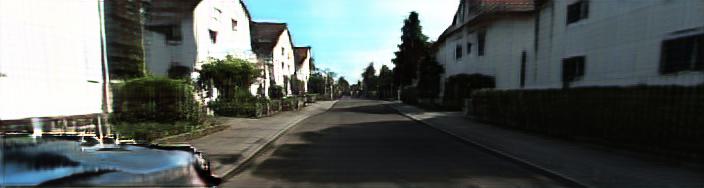}&
           \includegraphics[width=\mywidth\linewidth]{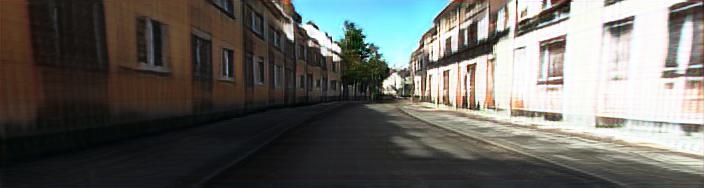}&
           \includegraphics[width=\mywidth\linewidth]{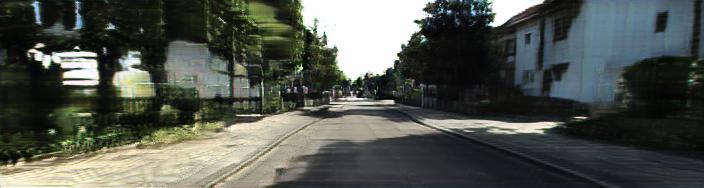}&
           \includegraphics[width=\mywidth\linewidth]{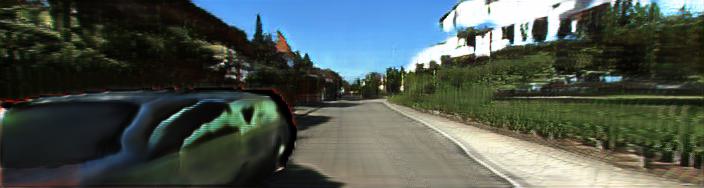}&
           \includegraphics[width=\mywidth\linewidth]{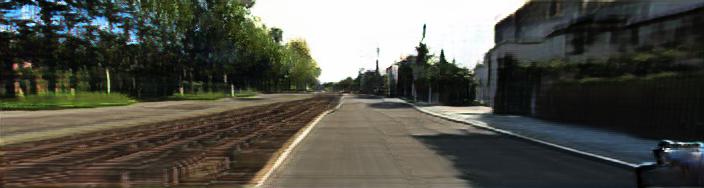} \vspace{\reduceheight}\\
        
           \includegraphics[width=\mywidth\linewidth]{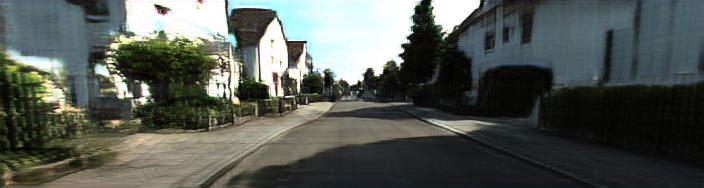}&
           \includegraphics[width=\mywidth\linewidth]{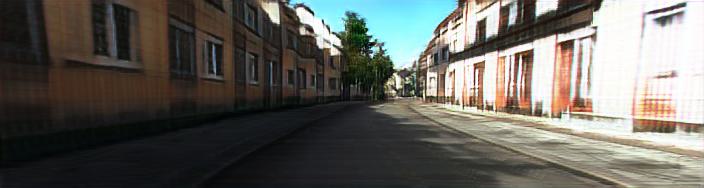}&
           \includegraphics[width=\mywidth\linewidth]{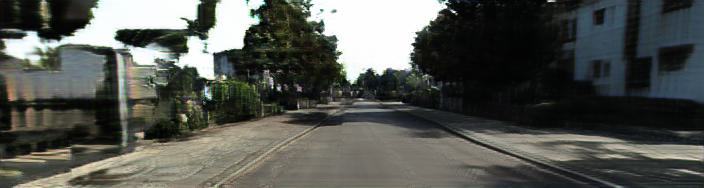}&
           \includegraphics[width=\mywidth\linewidth]{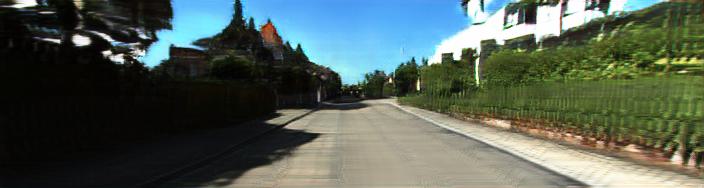}&
           \includegraphics[width=\mywidth\linewidth]{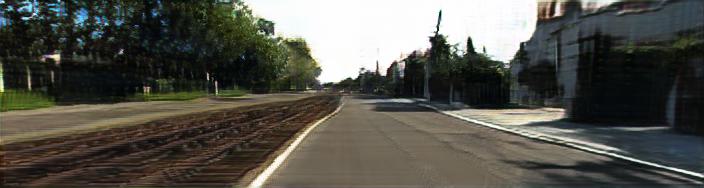} \vspace{\reduceheight}\\
        
           \includegraphics[width=\mywidth\linewidth]{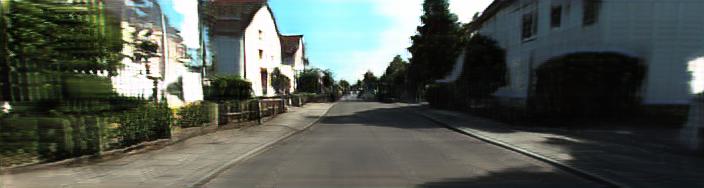}&
           \includegraphics[width=\mywidth\linewidth]{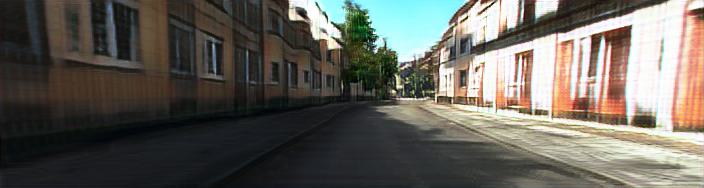}&
           \includegraphics[width=\mywidth\linewidth]{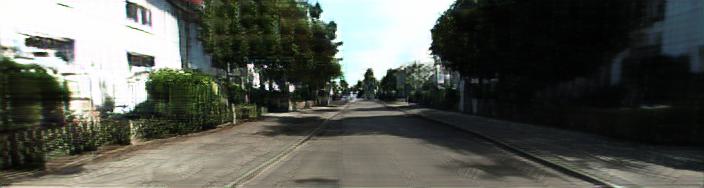}&
           \includegraphics[width=\mywidth\linewidth]{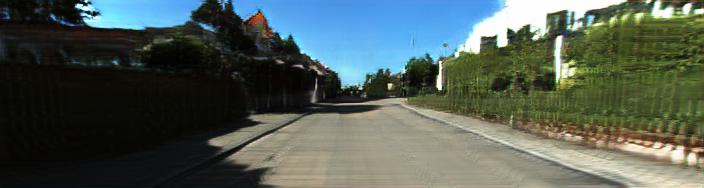}&
           \includegraphics[width=\mywidth\linewidth]{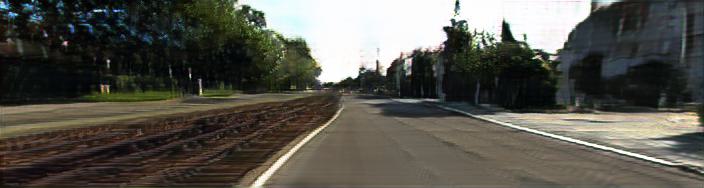}
           \vspace{\reduceheight}\\
          \end{tabular}\vspace{-0.1cm}}
     \caption{{\bf Additional Qualitative Comparison on KITTI-360}. The camera moves up to 20 meters in each scene.}
     \label{fig:comp_baselines_kitti360}
     \vspace{-0.3cm}
    \end{figure*}

%% file: gfx_supp/comp_clevr.tex
\begin{figure*}[htbp]
     \centering
     \setlength{\tabcolsep}{0pt}
     \def\mywidth{1.51cm}
     \def\largewidth{1.55cm}
     \def\reduceheight{-3.5pt}

     \subfloat[][GIRAFFE\label{fig:giraffe_clevr}\vspace{-0.2cm}]
     {\begin{tabular}{P{0.4cm}P{\mywidth}P{\largewidth}P{\mywidth}P{\mywidth}P{\largewidth}P{\mywidth}P{\largewidth}P{\mywidth}P{\largewidth}P{\mywidth}P{\mywidth}}
          &
          \includegraphics[width=\mywidth]{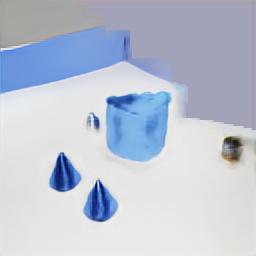} &
          \includegraphics[width=\mywidth]{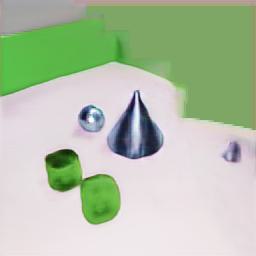} &
          \includegraphics[width=\mywidth]{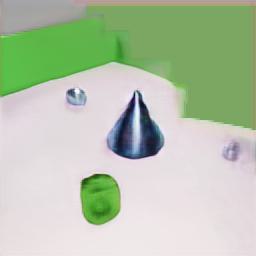} &
          \includegraphics[width=\mywidth]{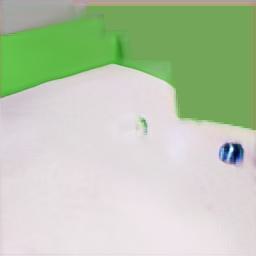} &
          \includegraphics[width=\mywidth]{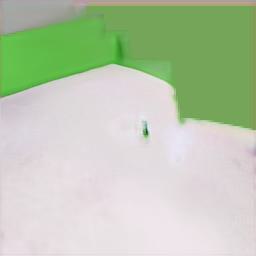} &
          \includegraphics[width=\mywidth]{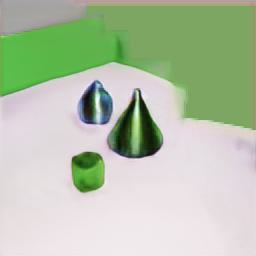} &
          \includegraphics[width=\mywidth]{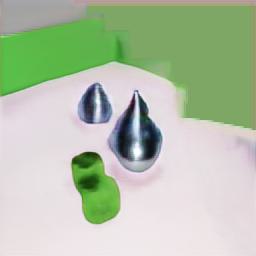} &
          \includegraphics[width=\mywidth]{gfx/results/place_holder_celvr.jpg} &
          \includegraphics[width=\mywidth]{gfx/results/place_holder_celvr.jpg} &
          \includegraphics[width=\mywidth]{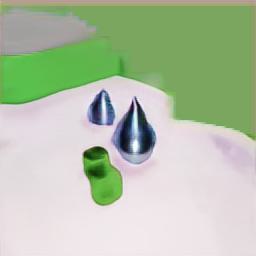} &
          \includegraphics[width=\mywidth]{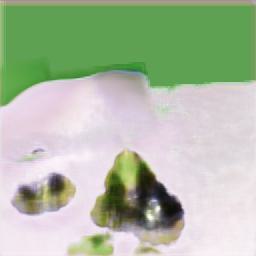} \\
          \vspace{\reduceheight}
          &
          \includegraphics[width=\mywidth]{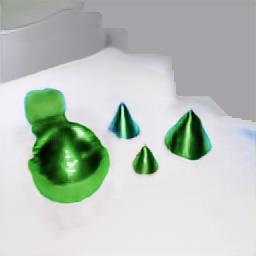} &
          \includegraphics[width=\mywidth]{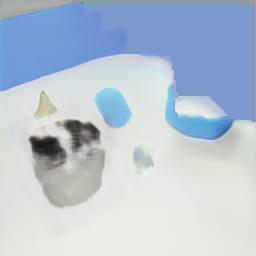} &
          \includegraphics[width=\mywidth]{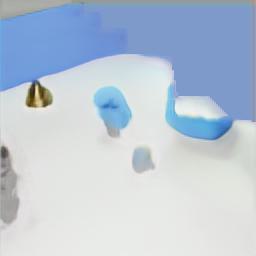} &
          \includegraphics[width=\mywidth]{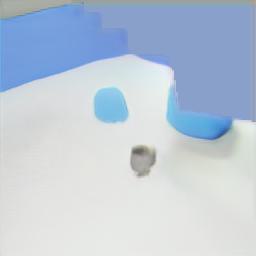} &
          \includegraphics[width=\mywidth]{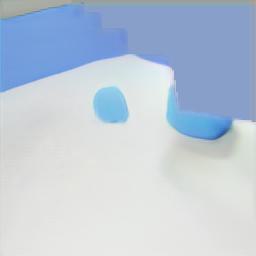} &
          \includegraphics[width=\mywidth]{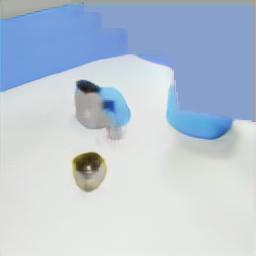} &
          \includegraphics[width=\mywidth]{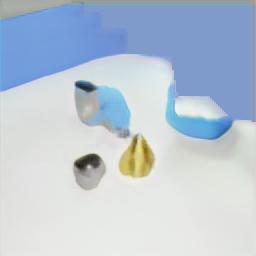} &
          \includegraphics[width=\mywidth]{gfx/results/place_holder_celvr.jpg} &
          \includegraphics[width=\mywidth]{gfx/results/place_holder_celvr.jpg} &
          \includegraphics[width=\mywidth]{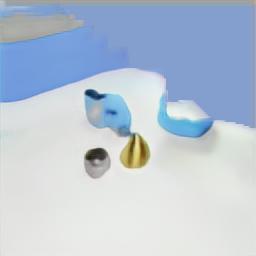} &
          \includegraphics[width=\mywidth]{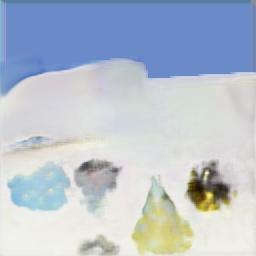} \\
          \vspace{\reduceheight}
          &
          \includegraphics[width=\mywidth]{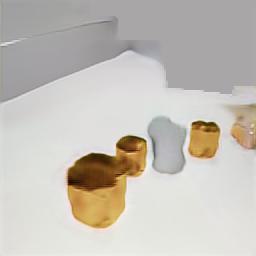} &
          \includegraphics[width=\mywidth]{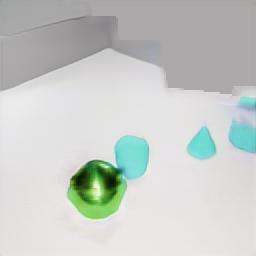} &
          \includegraphics[width=\mywidth]{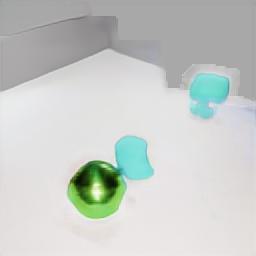} &
          \includegraphics[width=\mywidth]{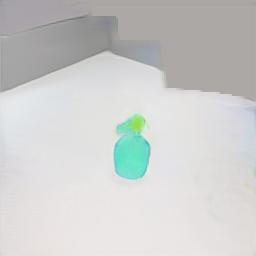} &
          \includegraphics[width=\mywidth]{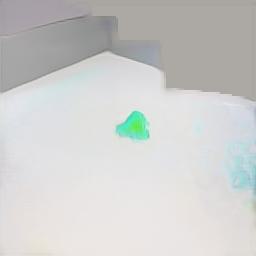} &
          \includegraphics[width=\mywidth]{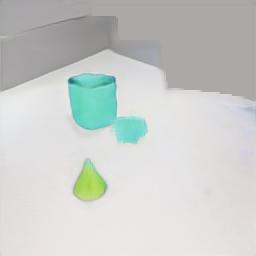} &
          \includegraphics[width=\mywidth]{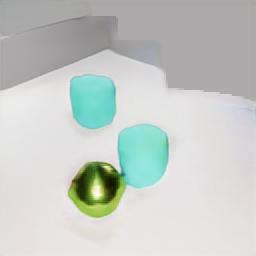} &
          \includegraphics[width=\mywidth]{gfx/results/place_holder_celvr.jpg} &
          \includegraphics[width=\mywidth]{gfx/results/place_holder_celvr.jpg} &
          \includegraphics[width=\mywidth]{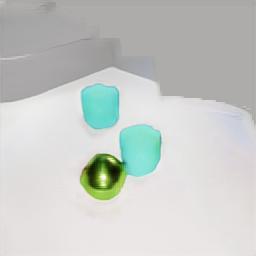} &
          \includegraphics[width=\mywidth]{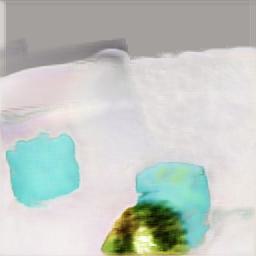} \\
          \vspace{\reduceheight}
         & \multicolumn{2}{c}{\small Appearance} & \multicolumn{3}{c}{\small Object Removal} & \multicolumn{2}{c}{\small Object Insertion} & \multicolumn{2}{c}{\small Stuff Editing} & \multicolumn{2}{c}{\small Camera Control}
         \end{tabular}\vspace{-0.1cm}

}

          \subfloat[][UrbanGIRAFFE (Ours)\label{fig:ours_clevr}\vspace{-0.2cm}]
          {\begin{tabular}{P{0.4cm}P{\mywidth}P{\largewidth}P{\mywidth}P{\mywidth}P{\largewidth}P{\mywidth}P{\largewidth}P{\mywidth}P{\largewidth}P{\mywidth}P{\mywidth}}
               &
               \includegraphics[width=\mywidth]{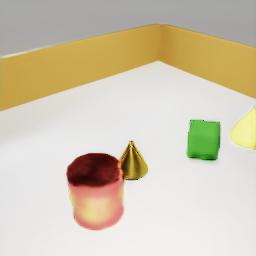} &
               \includegraphics[width=\mywidth]{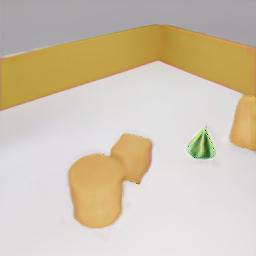} &
               \includegraphics[width=\mywidth]{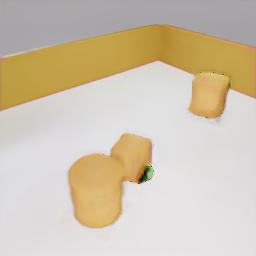} &
               \includegraphics[width=\mywidth]{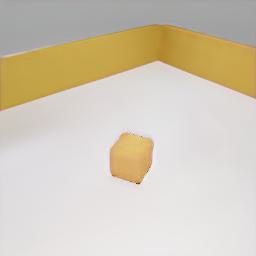} &
               \includegraphics[width=\mywidth]{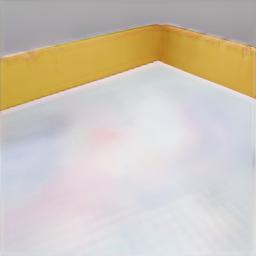} &
               \includegraphics[width=\mywidth]{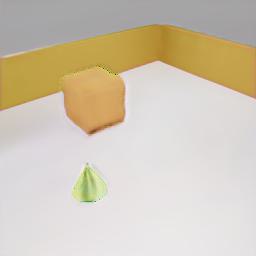} &
               \includegraphics[width=\mywidth]{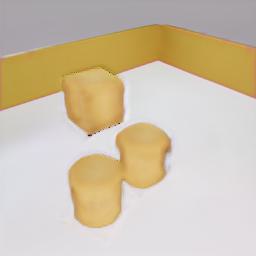} &
               \includegraphics[width=\mywidth]{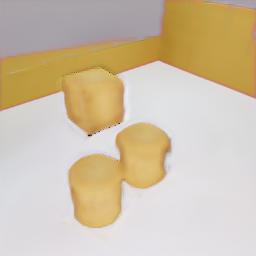} &
               \includegraphics[width=\mywidth]{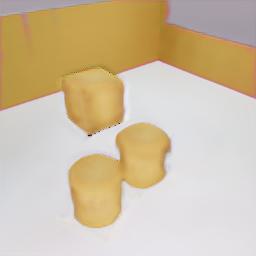} &
               \includegraphics[width=\mywidth]{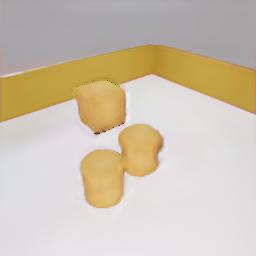} &
               \includegraphics[width=\mywidth]{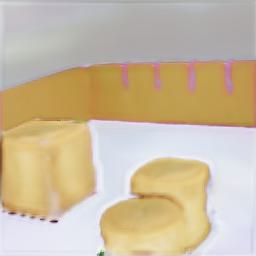} \\
               \vspace{\reduceheight}
               &
               \includegraphics[width=\mywidth]{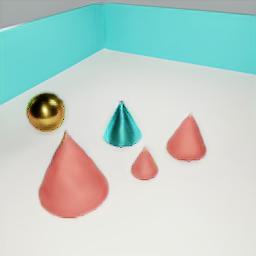} &
               \includegraphics[width=\mywidth]{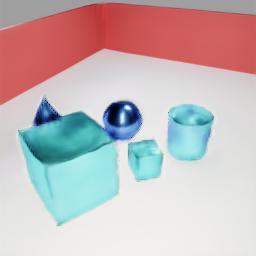} &
               \includegraphics[width=\mywidth]{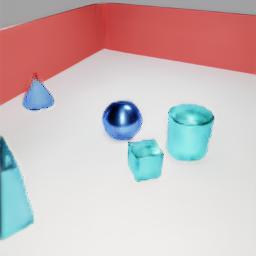} &
               \includegraphics[width=\mywidth]{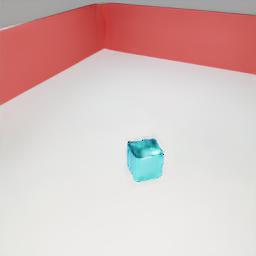} &
               \includegraphics[width=\mywidth]{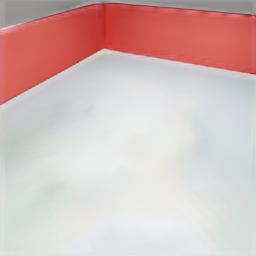} &
               \includegraphics[width=\mywidth]{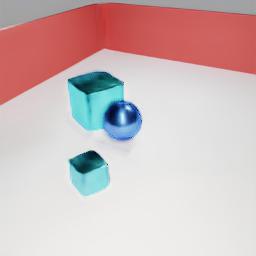} &
               \includegraphics[width=\mywidth]{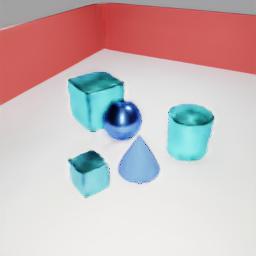} &
               \includegraphics[width=\mywidth]{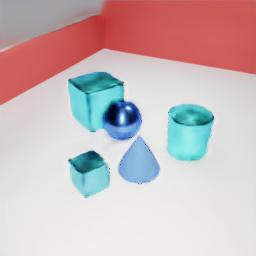} &
               \includegraphics[width=\mywidth]{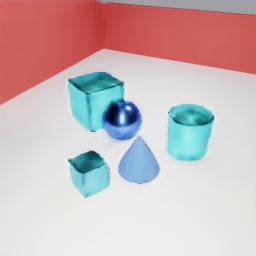} &
               \includegraphics[width=\mywidth]{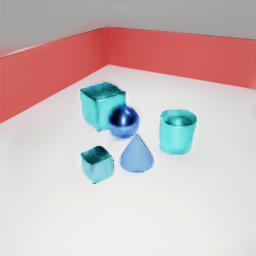} &
               \includegraphics[width=\mywidth]{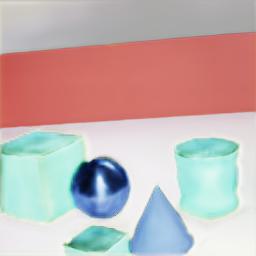} \\
               \vspace{\reduceheight}
               &
               \includegraphics[width=\mywidth]{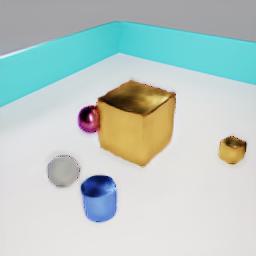} &
               \includegraphics[width=\mywidth]{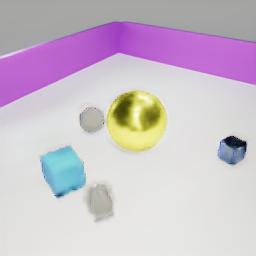} &
               \includegraphics[width=\mywidth]{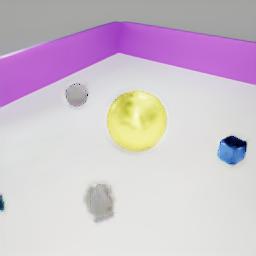} &
               \includegraphics[width=\mywidth]{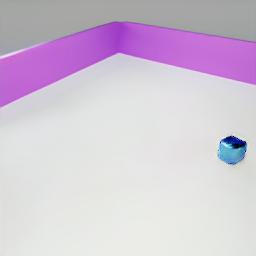} &
               \includegraphics[width=\mywidth]{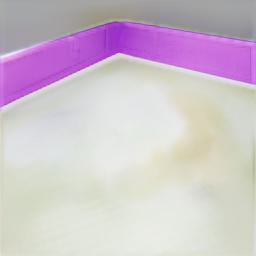} &
               \includegraphics[width=\mywidth]{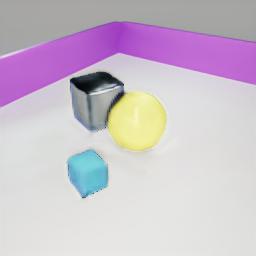} &
               \includegraphics[width=\mywidth]{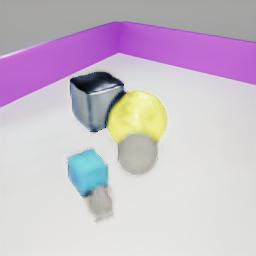} &
               \includegraphics[width=\mywidth]{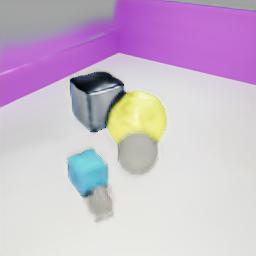} &
               \includegraphics[width=\mywidth]{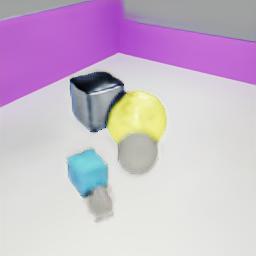} &
               \includegraphics[width=\mywidth]{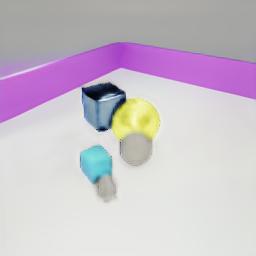} &
               \includegraphics[width=\mywidth]{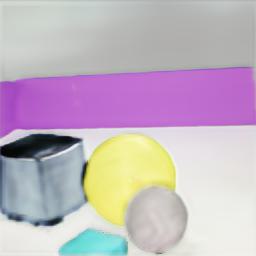} \\
               \vspace{\reduceheight}
              & \multicolumn{2}{c}{\small Appearance} & \multicolumn{3}{c}{\small Object Removal} & \multicolumn{2}{c}{\small Object Insertion} & \multicolumn{2}{c}{\small Stuff Editing} & \multicolumn{2}{c}{\small Camera Control}
              \end{tabular}\vspace{-0.1cm}

     }
     \caption{{\bf Additional Qualitative Comparison on CLEVR-W}. Our method enables stuff editing and shows more convincing results in camera viewpoint control.}
     \label{fig:comp_baselines_clevr}
     \vspace{-0.021cm}
    \end{figure*}

%% file: gfx_supp/failure.tex
\begin{figure*}[htbp!]
     \centering
     \setlength{\tabcolsep}{0pt}
     \def\mywidth{.25}
    \def\reduceheight{-3.5pt}

     \subfloat[][Sampling $\bz_{wld}$ can only slightly adjust the stuff appearance\label{fig:z_fail}\vspace{-0.2cm}]{
     \begin{tabular}{cccc}
   
      \includegraphics[width=\mywidth\linewidth]{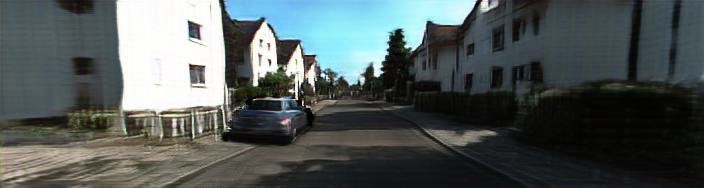}&
      \includegraphics[width=\mywidth\linewidth]{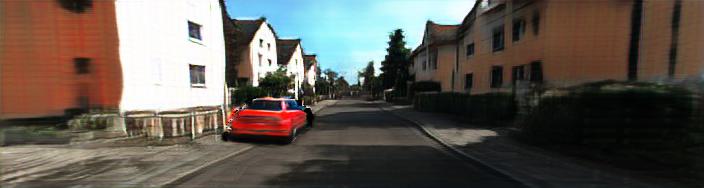}&
      \includegraphics[width=\mywidth\linewidth]{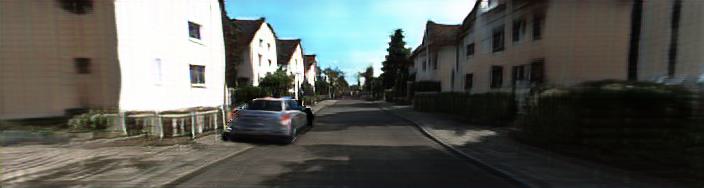}&
      \includegraphics[width=\mywidth\linewidth]{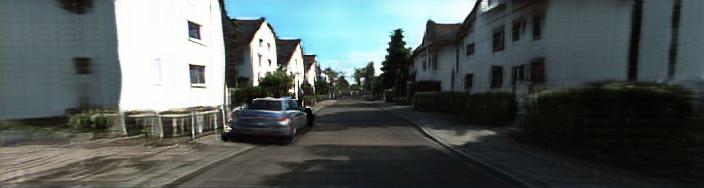}  \vspace{\reduceheight}\\
   
     \end{tabular}\vspace{-0.2cm}}\\
     \subfloat[][Sky occasionally models far buildings\label{fig:sky_fail}\vspace{-0.2cm}]{
     \begin{tabular}{cccc}

          \includegraphics[width=\mywidth\linewidth]{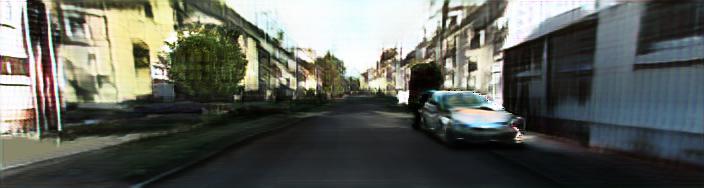}&
          \includegraphics[width=\mywidth\linewidth]{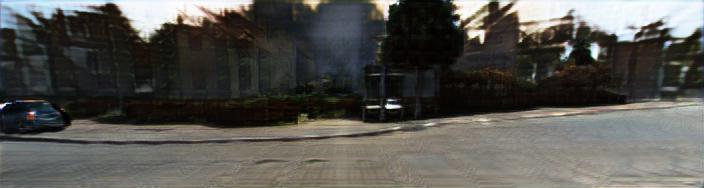}&
          \includegraphics[width=\mywidth\linewidth]{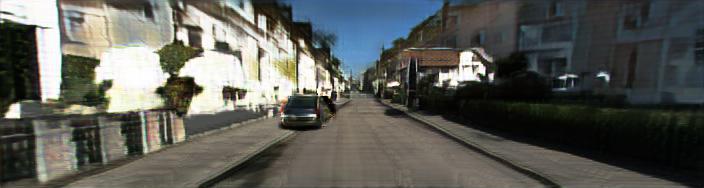}&
          \includegraphics[width=\mywidth\linewidth]{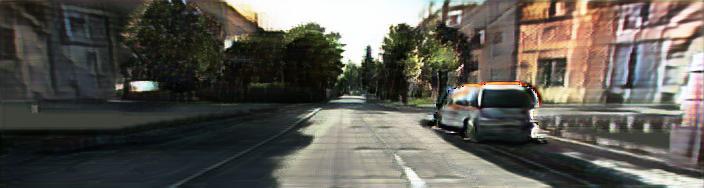}  
          \vspace{\reduceheight}\\

          \vspace{\reduceheight}\\
     \end{tabular}\vspace{-0.1cm}}
     \caption{{\bf Limitations} of UrbanGIRAFFE.}
     \label{fig:fail}
     \vspace{-0.3cm}
    \end{figure*}

%% file: gfx_supp/editing_kitti360.tex
\begin{figure*}[htbp!]
     \centering
     \setlength{\tabcolsep}{0pt}
     \def\mywidth{.2}
    \def\reduceheight{-3.5pt}

     \subfloat[][Lower building\label{fig:building2tree}\vspace{-0.2cm}]{
     \begin{tabular}{ccccc}
   
      \includegraphics[width=\mywidth\linewidth]{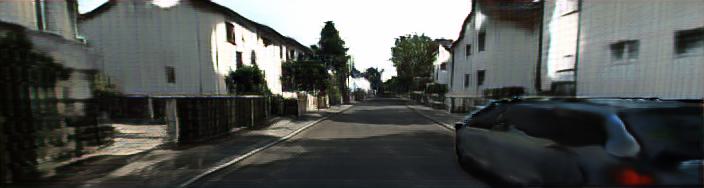}&
      \includegraphics[width=\mywidth\linewidth]{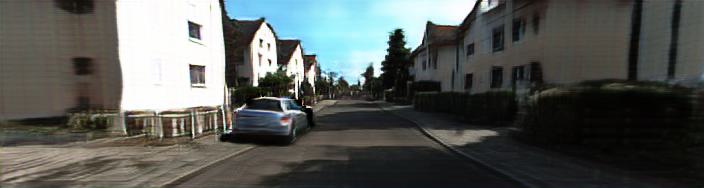}&
      \includegraphics[width=\mywidth\linewidth]{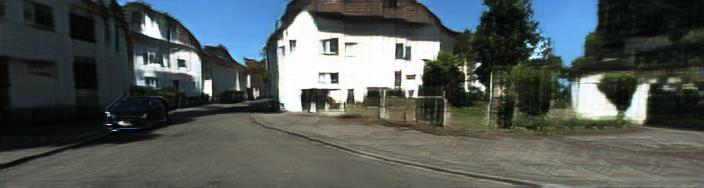}&
      \includegraphics[width=\mywidth\linewidth]{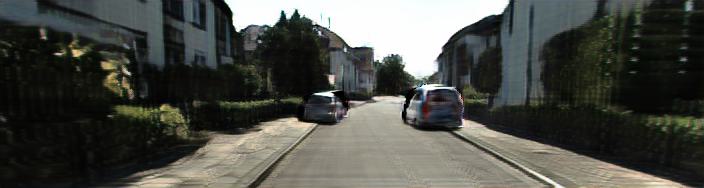}&
      \includegraphics[width=\mywidth\linewidth]{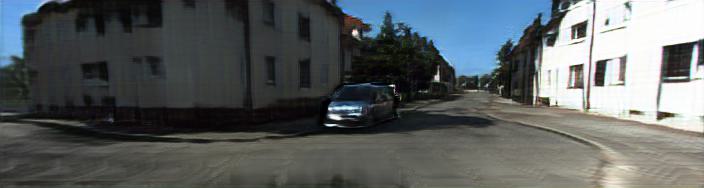} \vspace{\reduceheight}\\
   
      \includegraphics[width=\mywidth\linewidth]{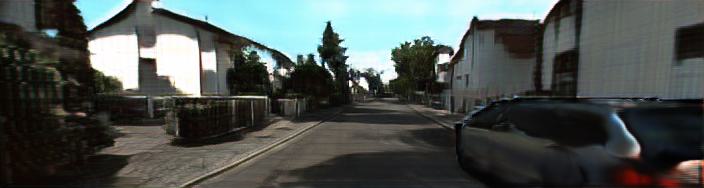}&
      \includegraphics[width=\mywidth\linewidth]{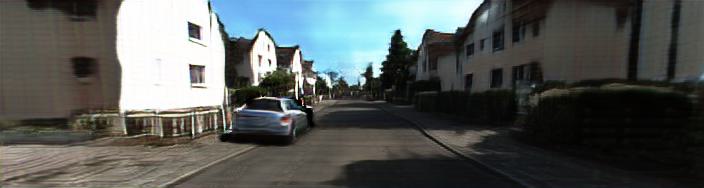}&
      \includegraphics[width=\mywidth\linewidth]{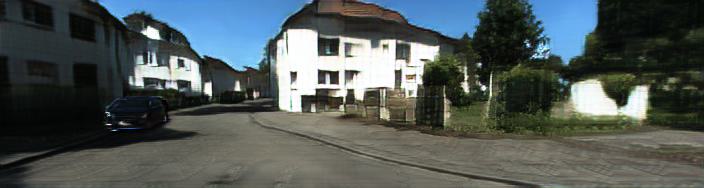}&
      \includegraphics[width=\mywidth\linewidth]{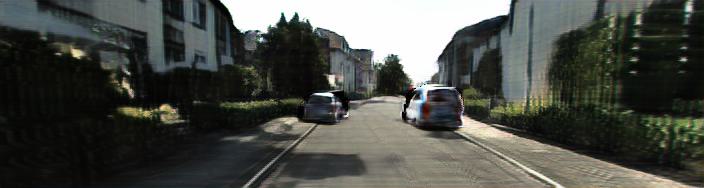}&
      \includegraphics[width=\mywidth\linewidth]{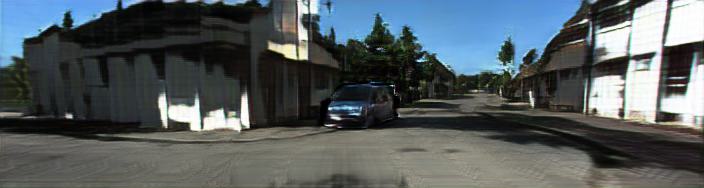} \vspace{\reduceheight}\\
   
      \includegraphics[width=\mywidth\linewidth]{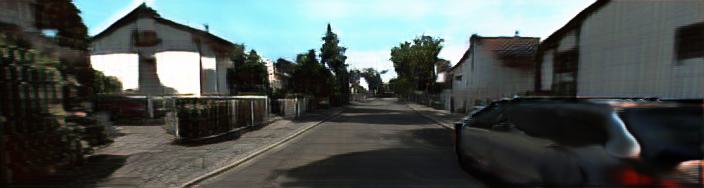}&
      \includegraphics[width=\mywidth\linewidth]{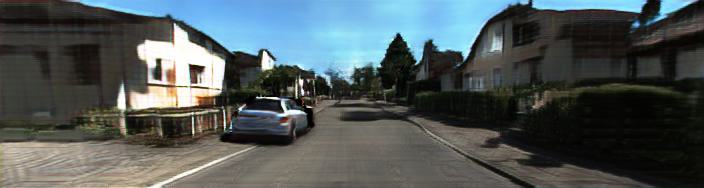}&
      \includegraphics[width=\mywidth\linewidth]{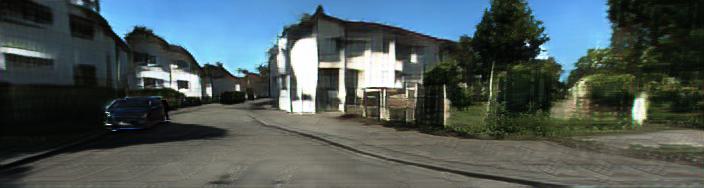}&
      \includegraphics[width=\mywidth\linewidth]{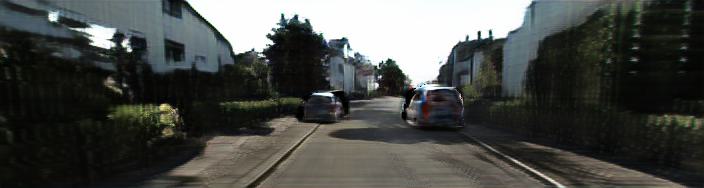}&
      \includegraphics[width=\mywidth\linewidth]{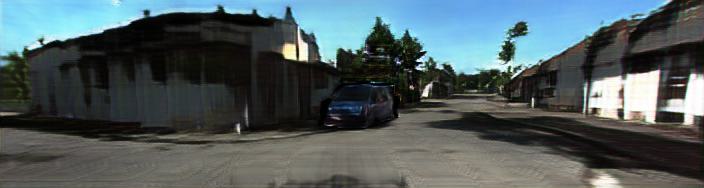} \vspace{\reduceheight}\\
   
      \includegraphics[width=\mywidth\linewidth]{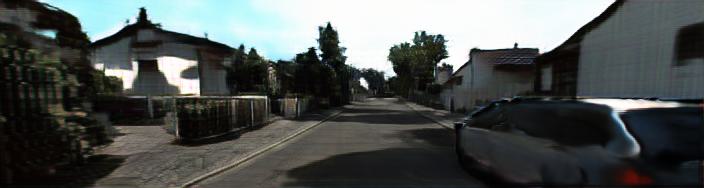}&
      \includegraphics[width=\mywidth\linewidth]{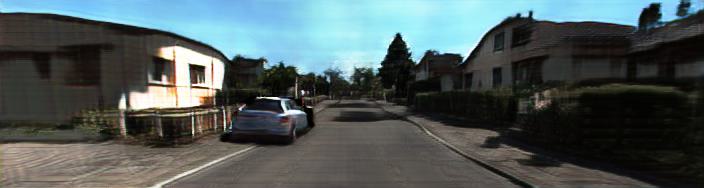}&
      \includegraphics[width=\mywidth\linewidth]{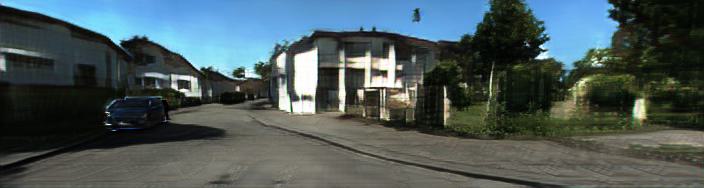}&
      \includegraphics[width=\mywidth\linewidth]{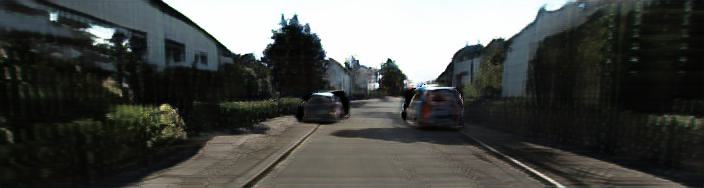}&
      \includegraphics[width=\mywidth\linewidth]{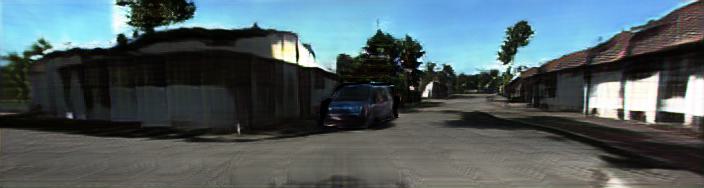} \vspace{\reduceheight}\\

      \includegraphics[width=\mywidth\linewidth]{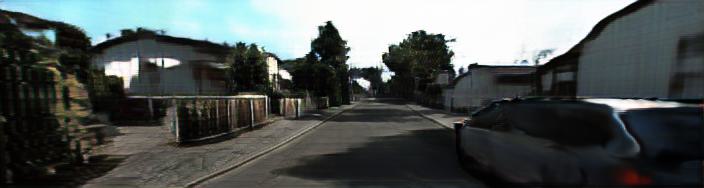}&
      \includegraphics[width=\mywidth\linewidth]{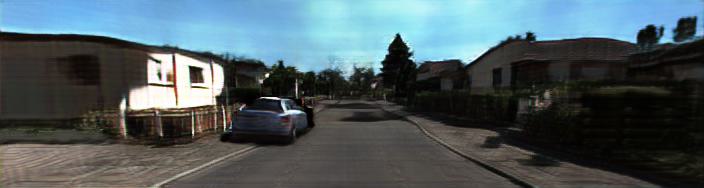}&
      \includegraphics[width=\mywidth\linewidth]{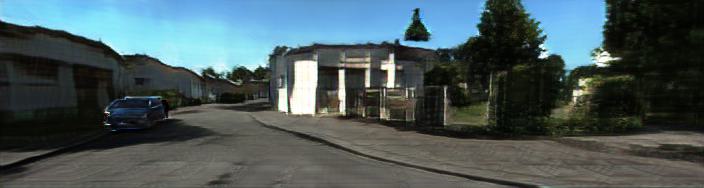}&
      \includegraphics[width=\mywidth\linewidth]{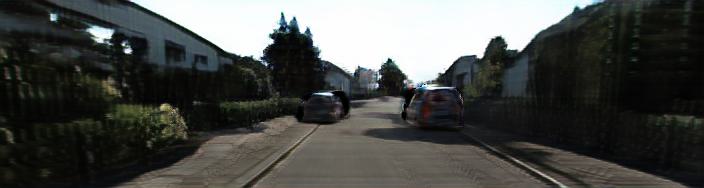}&
      \includegraphics[width=\mywidth\linewidth]{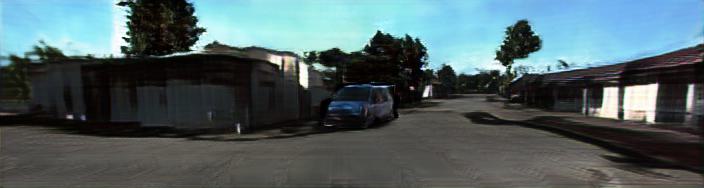} \vspace{\reduceheight}\\

     \end{tabular}\vspace{-0.1cm}}

     \subfloat[][Building to tree\label{fig:lowerbuilding}\vspace{-0.2cm}]{
          \begin{tabular}{ccccc}
   
               \includegraphics[width=\mywidth\linewidth]{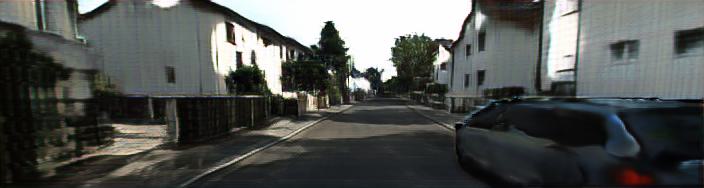}&
               \includegraphics[width=\mywidth\linewidth]{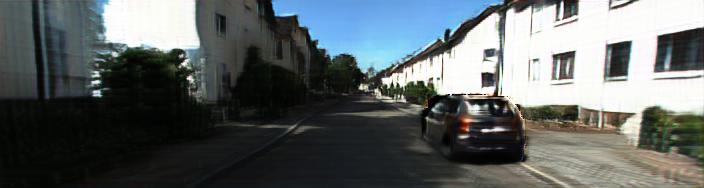}&
               \includegraphics[width=\mywidth\linewidth]{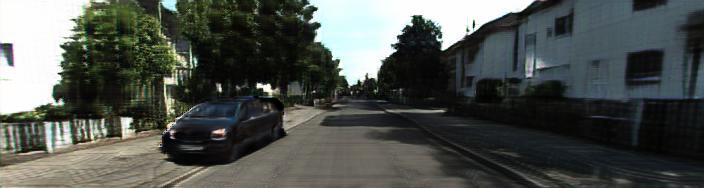}&
               \includegraphics[width=\mywidth\linewidth]{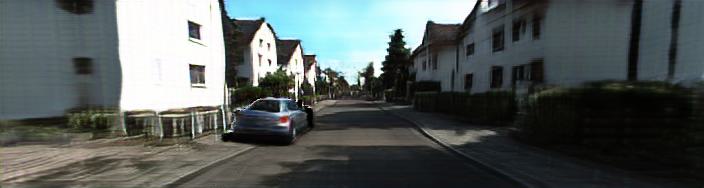}&
               \includegraphics[width=\mywidth\linewidth]{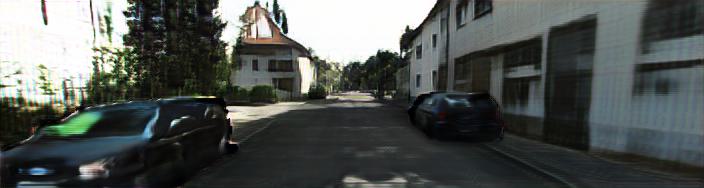} \vspace{\reduceheight}\\
            
               \includegraphics[width=\mywidth\linewidth]{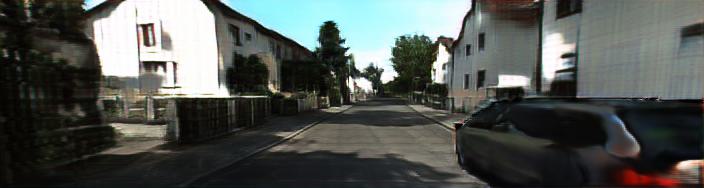}&
               \includegraphics[width=\mywidth\linewidth]{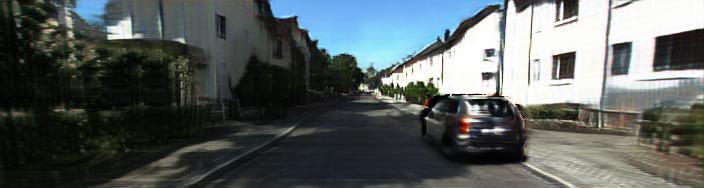}&
               \includegraphics[width=\mywidth\linewidth]{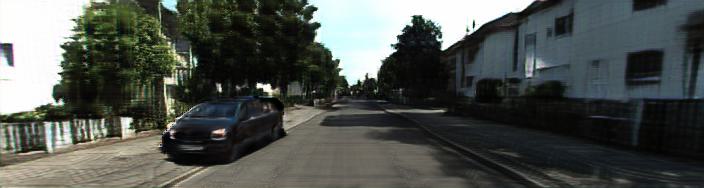}&
               \includegraphics[width=\mywidth\linewidth]{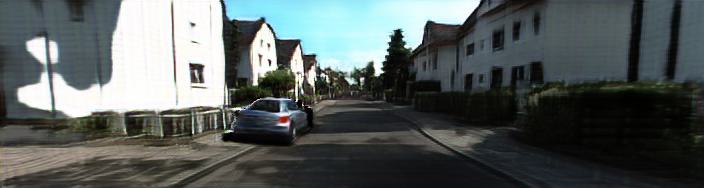}&
               \includegraphics[width=\mywidth\linewidth]{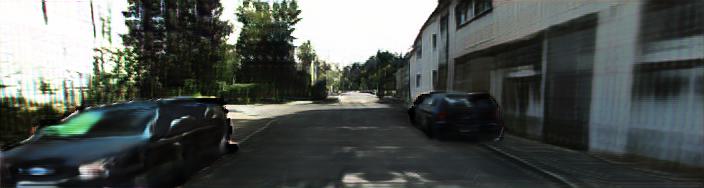} \vspace{\reduceheight}\\
            
               \includegraphics[width=\mywidth\linewidth]{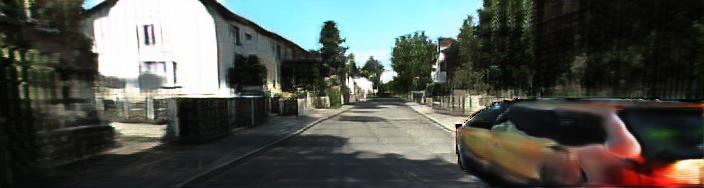}&
               \includegraphics[width=\mywidth\linewidth]{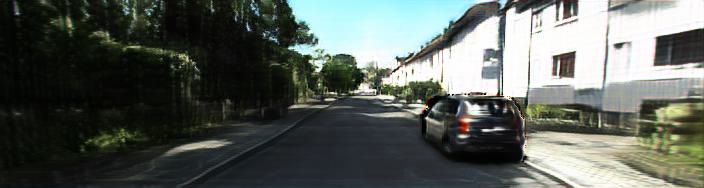}&
               \includegraphics[width=\mywidth\linewidth]{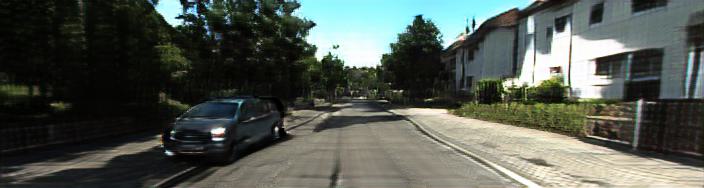}&
               \includegraphics[width=\mywidth\linewidth]{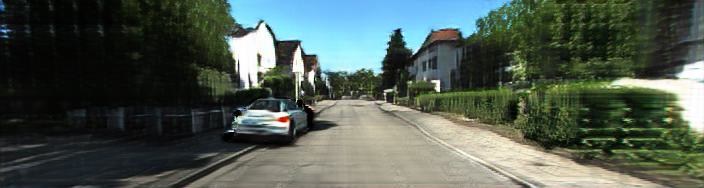}&
               \includegraphics[width=\mywidth\linewidth]{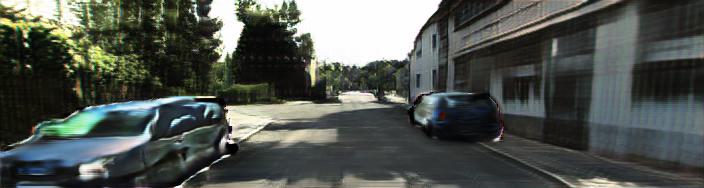} \vspace{\reduceheight}\\
            
               \includegraphics[width=\mywidth\linewidth]{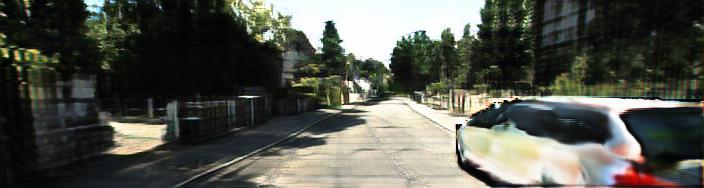}&
               \includegraphics[width=\mywidth\linewidth]{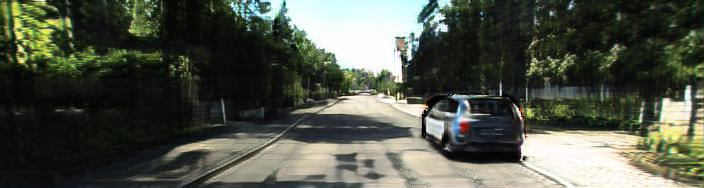}&
               \includegraphics[width=\mywidth\linewidth]{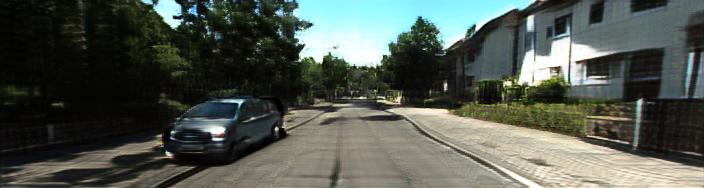}&
               \includegraphics[width=\mywidth\linewidth]{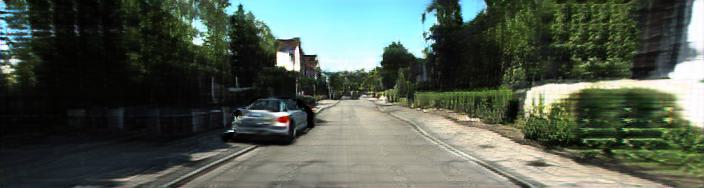}&
               \includegraphics[width=\mywidth\linewidth]{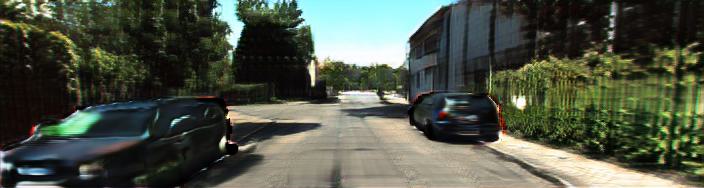} \vspace{\reduceheight}\\

               \includegraphics[width=\mywidth\linewidth]{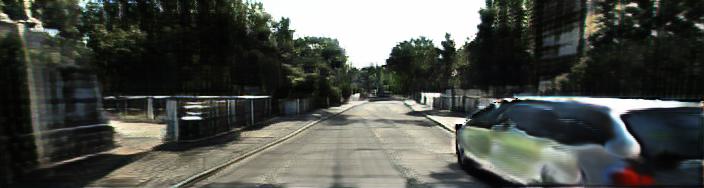}&
               \includegraphics[width=\mywidth\linewidth]{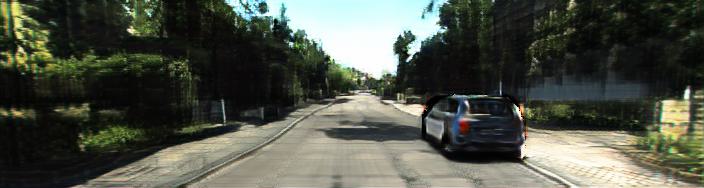}&
               \includegraphics[width=\mywidth\linewidth]{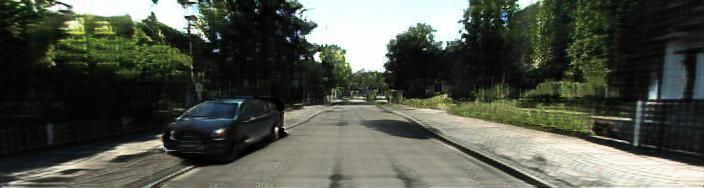}&
               \includegraphics[width=\mywidth\linewidth]{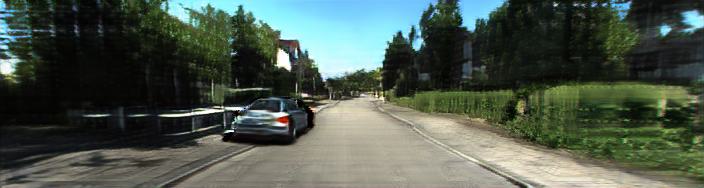}&
               \includegraphics[width=\mywidth\linewidth]{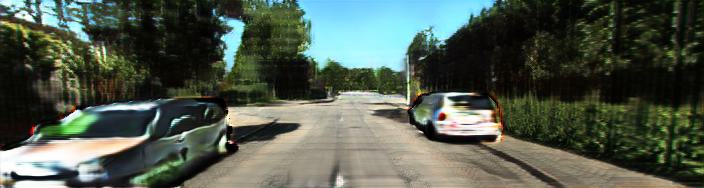} \vspace{\reduceheight}\\
         
              \end{tabular}\vspace{-0.1cm}}

     \subfloat[][Road to grass\label{fig:road2grass}\vspace{-0.2cm}]{
          \begin{tabular}{ccccc}

               \includegraphics[width=\mywidth\linewidth]{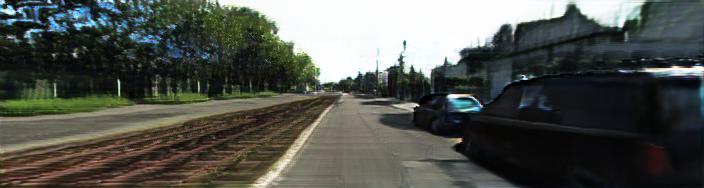}&
               \includegraphics[width=\mywidth\linewidth]{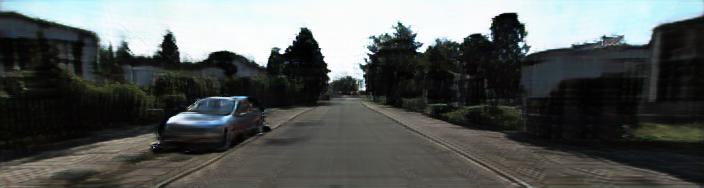}&
               \includegraphics[width=\mywidth\linewidth]{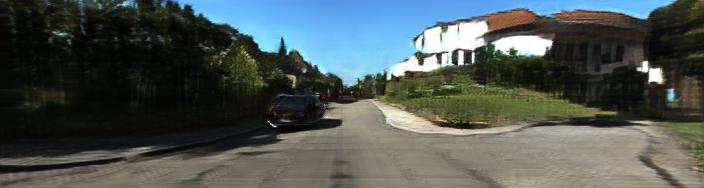}&
               \includegraphics[width=\mywidth\linewidth]{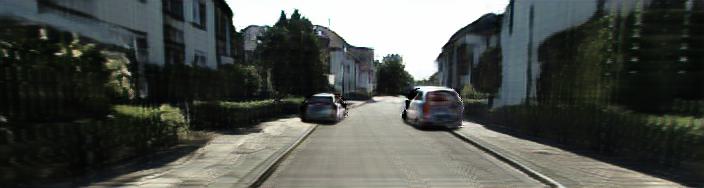}&
               \includegraphics[width=\mywidth\linewidth]{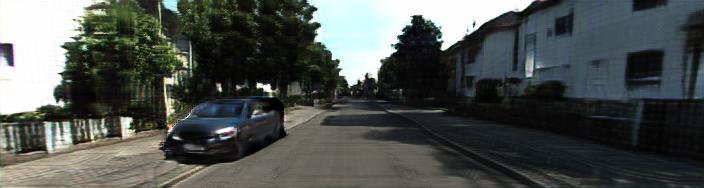} \vspace{\reduceheight}\\
            
               \includegraphics[width=\mywidth\linewidth]{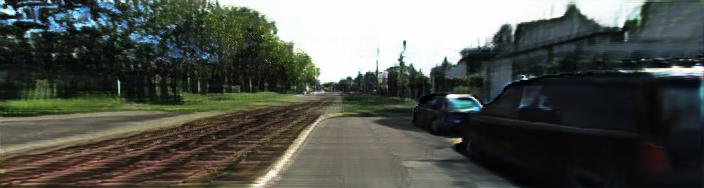}&
               \includegraphics[width=\mywidth\linewidth]{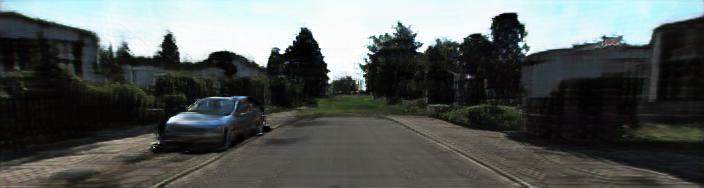}&
               \includegraphics[width=\mywidth\linewidth]{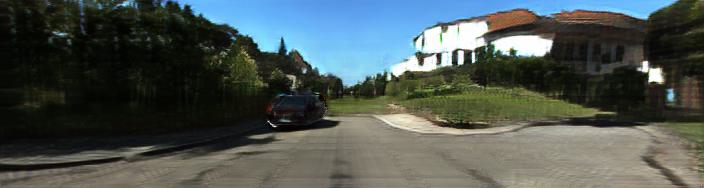}&
               \includegraphics[width=\mywidth\linewidth]{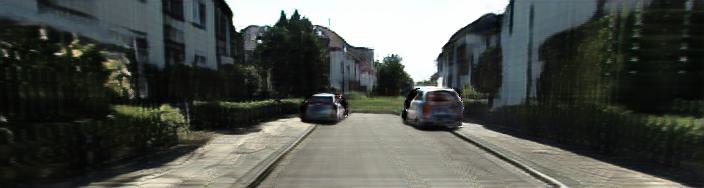}&
               \includegraphics[width=\mywidth\linewidth]{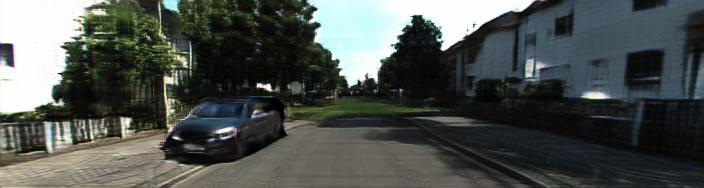} \vspace{\reduceheight}\\
            
               \includegraphics[width=\mywidth\linewidth]{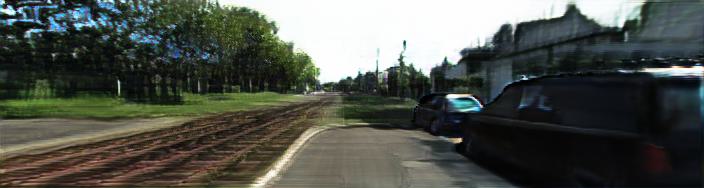}&
               \includegraphics[width=\mywidth\linewidth]{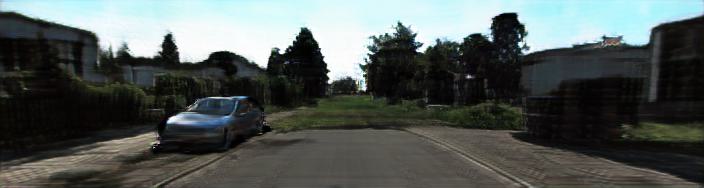}&
               \includegraphics[width=\mywidth\linewidth]{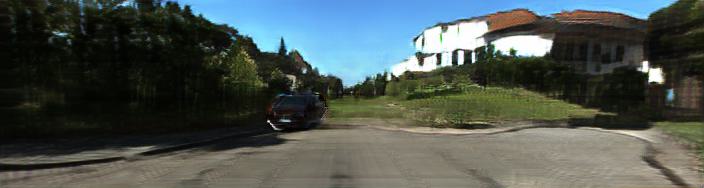}&
               \includegraphics[width=\mywidth\linewidth]{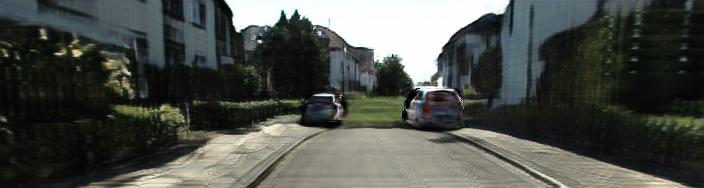}&
               \includegraphics[width=\mywidth\linewidth]{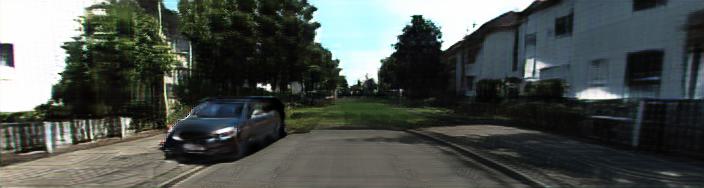} \vspace{\reduceheight}\\
            
               \includegraphics[width=\mywidth\linewidth]{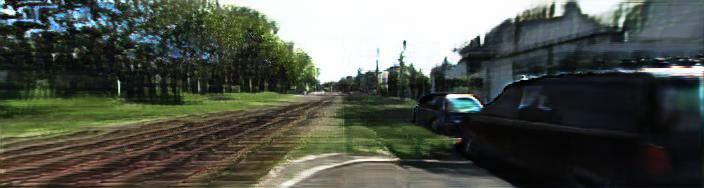}&
               \includegraphics[width=\mywidth\linewidth]{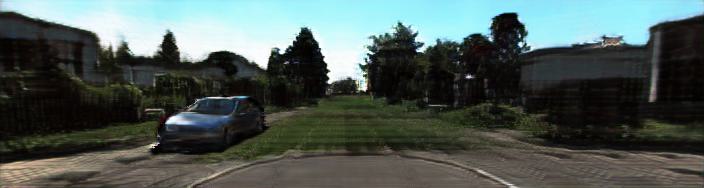}&
               \includegraphics[width=\mywidth\linewidth]{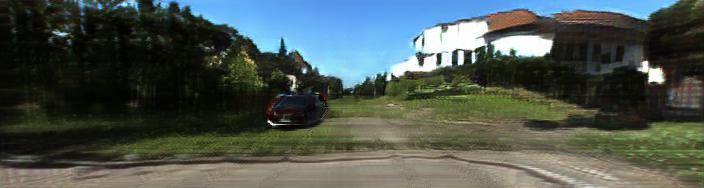}&
               \includegraphics[width=\mywidth\linewidth]{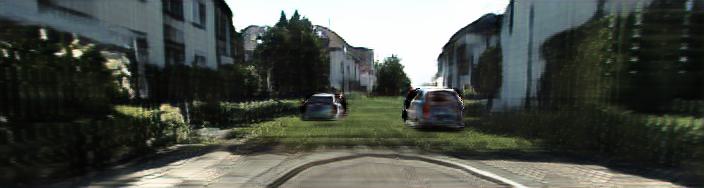}&
               \includegraphics[width=\mywidth\linewidth]{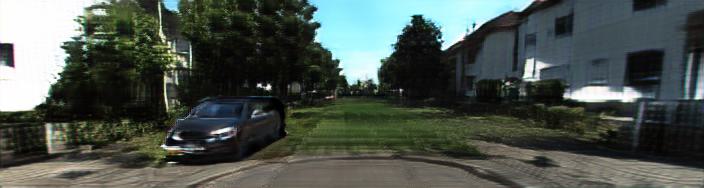} \vspace{\reduceheight}\\

               \includegraphics[width=\mywidth\linewidth]{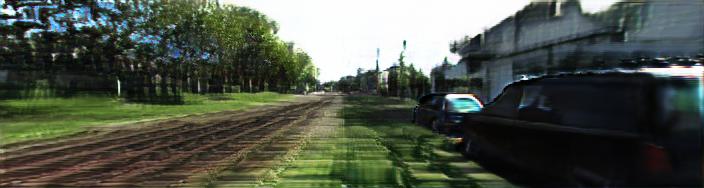}&
               \includegraphics[width=\mywidth\linewidth]{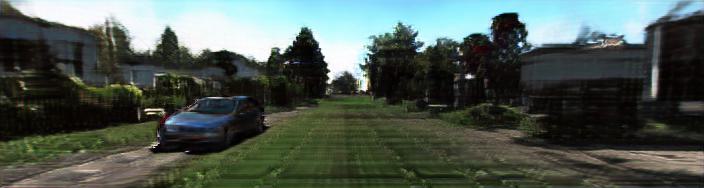}&
               \includegraphics[width=\mywidth\linewidth]{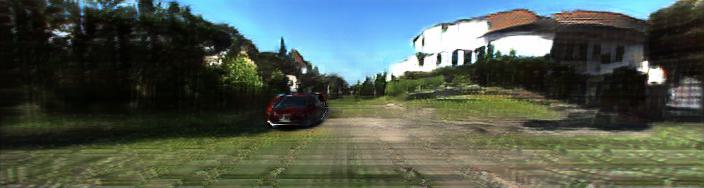}&
               \includegraphics[width=\mywidth\linewidth]{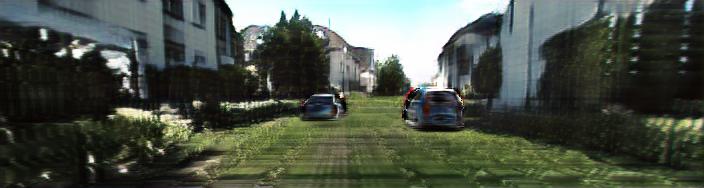}&
               \includegraphics[width=\mywidth\linewidth]{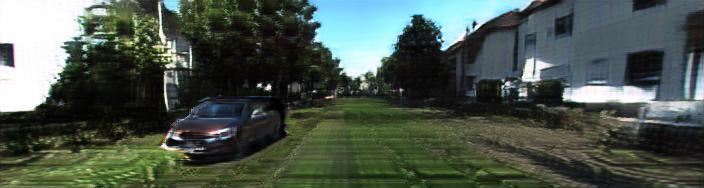} \vspace{\reduceheight}\\
         
              \end{tabular}\vspace{-0.1cm}}

     \subfloat[][Object editing\label{fig:object_editing}\vspace{-0.2cm}]{
     \begin{tabular}{ccccc}

          \includegraphics[width=\mywidth\linewidth]{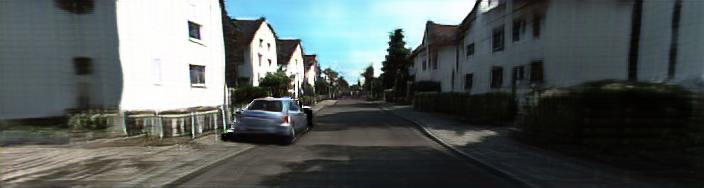}&
          \includegraphics[width=\mywidth\linewidth]{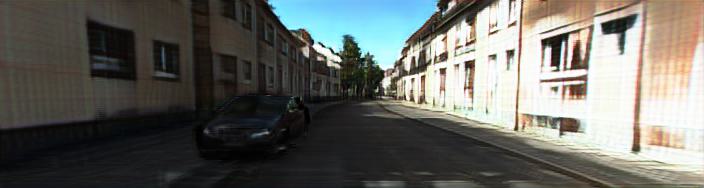}&
          \includegraphics[width=\mywidth\linewidth]{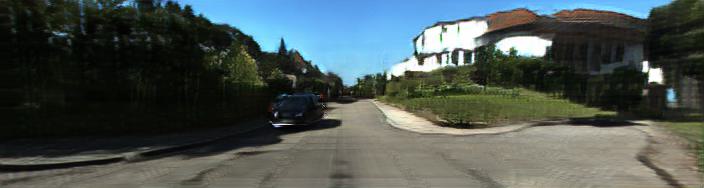}&
          \includegraphics[width=\mywidth\linewidth]{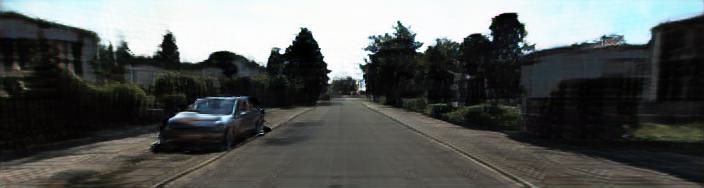}&
          \includegraphics[width=\mywidth\linewidth]{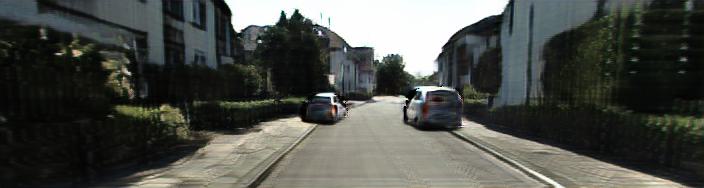} \vspace{\reduceheight}\\
          
          \includegraphics[width=\mywidth\linewidth]{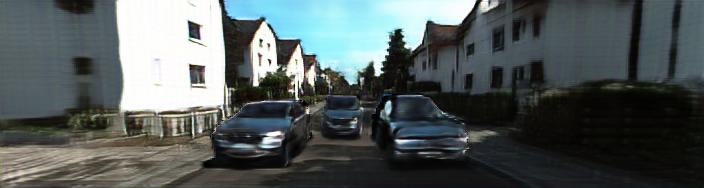}&
          \includegraphics[width=\mywidth\linewidth]{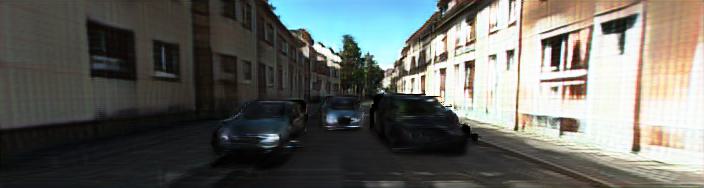}&
          \includegraphics[width=\mywidth\linewidth]{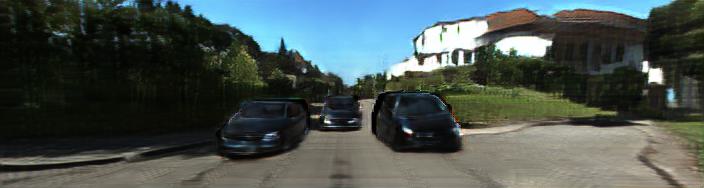}&
          \includegraphics[width=\mywidth\linewidth]{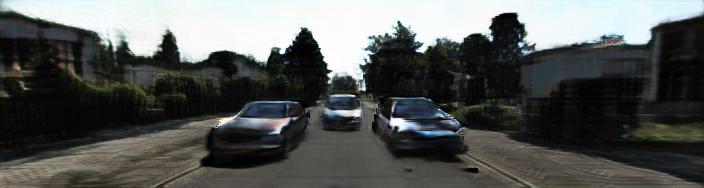}&
          \includegraphics[width=\mywidth\linewidth]{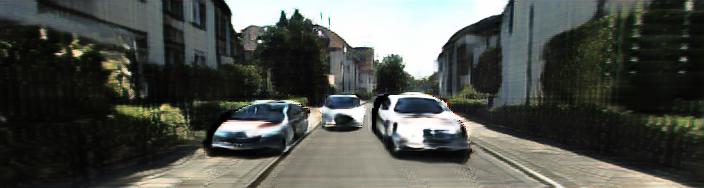} \vspace{\reduceheight}\\

          \includegraphics[width=\mywidth\linewidth]{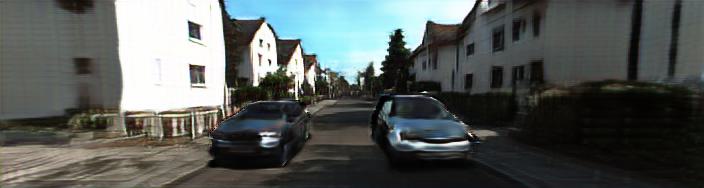}&
          \includegraphics[width=\mywidth\linewidth]{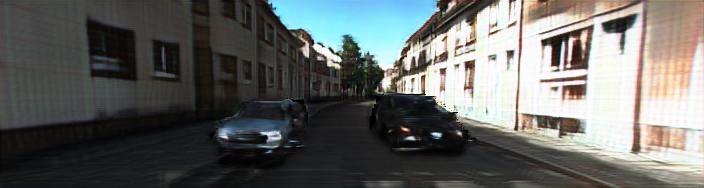}&
          \includegraphics[width=\mywidth\linewidth]{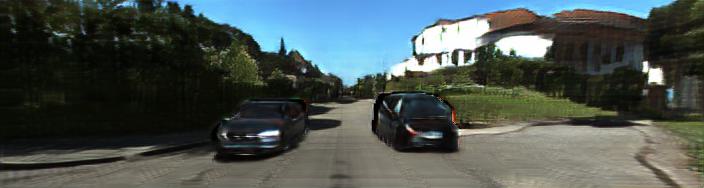}&
          \includegraphics[width=\mywidth\linewidth]{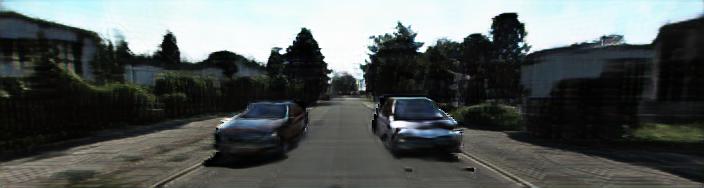}&
          \includegraphics[width=\mywidth\linewidth]{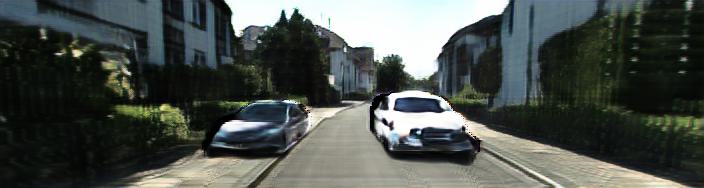} \vspace{\reduceheight}\\

          \includegraphics[width=\mywidth\linewidth]{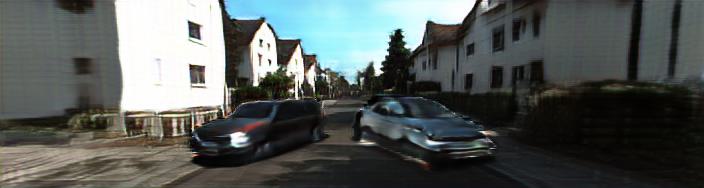}&
          \includegraphics[width=\mywidth\linewidth]{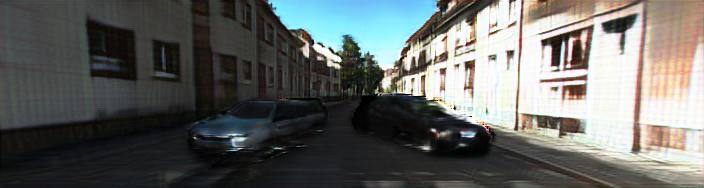}&
          \includegraphics[width=\mywidth\linewidth]{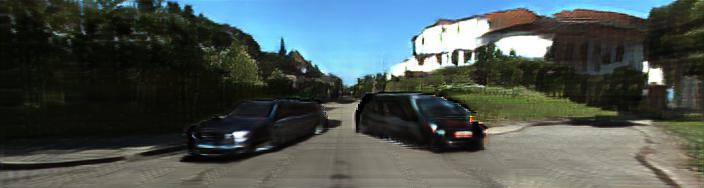}&
          \includegraphics[width=\mywidth\linewidth]{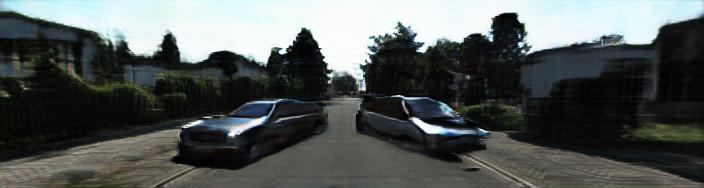}&
          \includegraphics[width=\mywidth\linewidth]{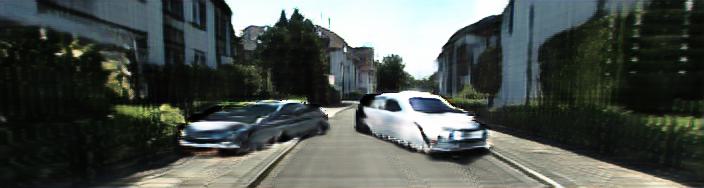} \vspace{\reduceheight}\\
          
          \includegraphics[width=\mywidth\linewidth]{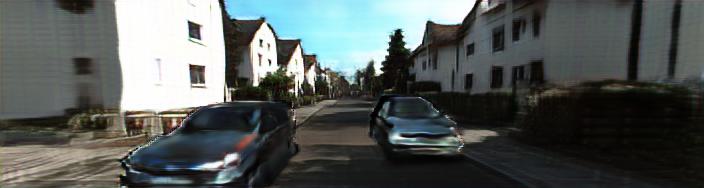}&
          \includegraphics[width=\mywidth\linewidth]{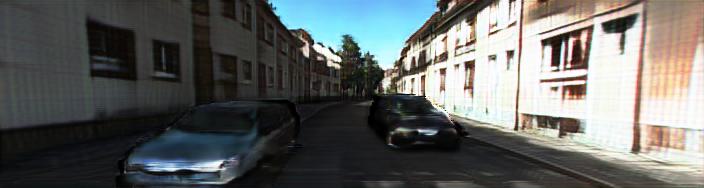}&
          \includegraphics[width=\mywidth\linewidth]{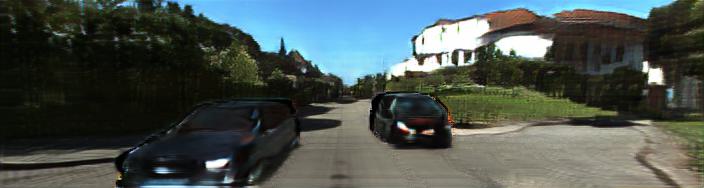}&
          \includegraphics[width=\mywidth\linewidth]{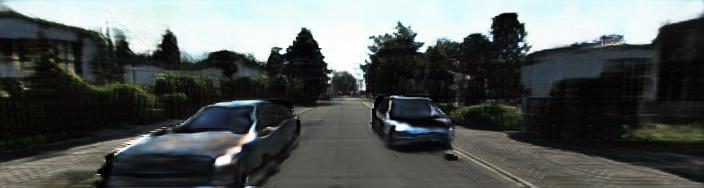}&
          \includegraphics[width=\mywidth\linewidth]{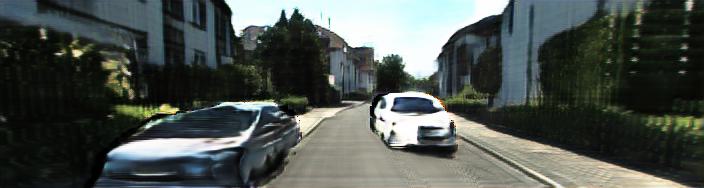} \vspace{\reduceheight}\\
          
          \includegraphics[width=\mywidth\linewidth]{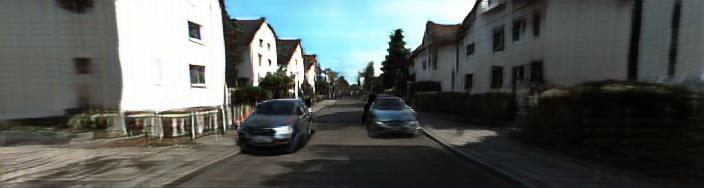}&
          \includegraphics[width=\mywidth\linewidth]{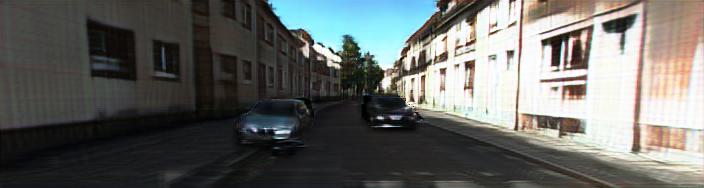}&
          \includegraphics[width=\mywidth\linewidth]{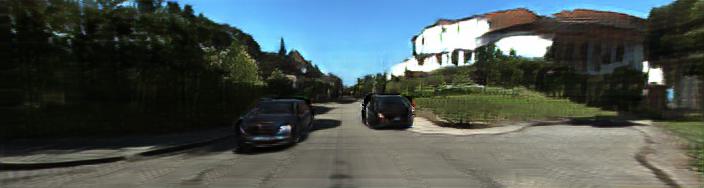}&
          \includegraphics[width=\mywidth\linewidth]{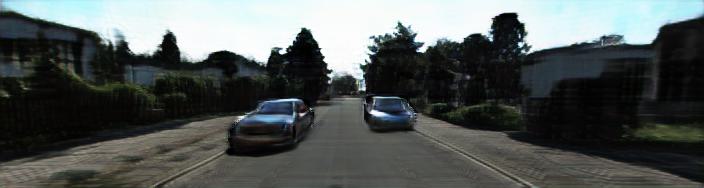}&
          \includegraphics[width=\mywidth\linewidth]{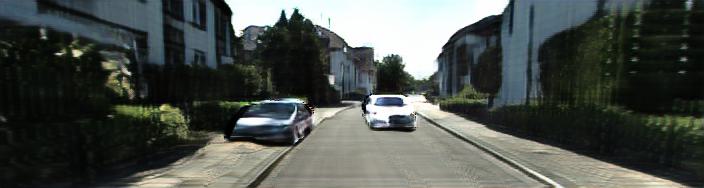} \vspace{\reduceheight}\\

          \end{tabular}\vspace{-0.1cm}}
     \caption{{\bf Additional Controllable Image Synthesis Results} on KITTI-360 dataset.}
     \label{fig:editing}
     \vspace{-0.3cm}
    \end{figure*}

%% file: gfx_supp/random.tex
\begin{figure*}[htbp!]
     \centering
     \setlength{\tabcolsep}{0pt}
     \def\mywidth{.2}
     \def\mywidthclevr{.1}
    \def\reduceheight{-3.5pt}
    \subfloat[][KITTI-360\label{fig:random_kitti360}\vspace{-0.2cm}]{
     \begin{tabular}{ccccc}

      \includegraphics[width=\mywidth\linewidth]{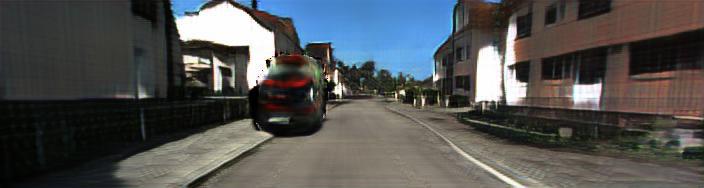}&
      \includegraphics[width=\mywidth\linewidth]{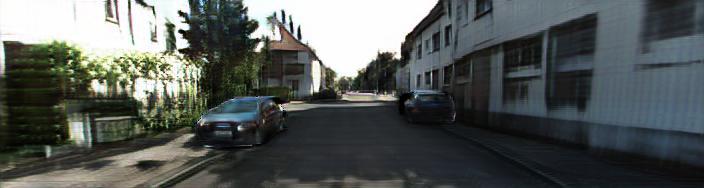}&
      \includegraphics[width=\mywidth\linewidth]{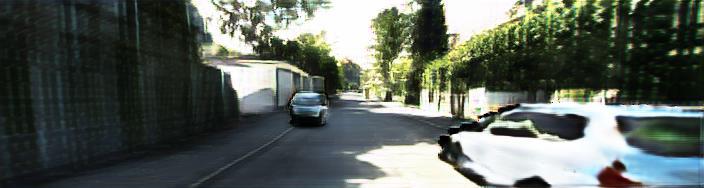}&
      \includegraphics[width=\mywidth\linewidth]{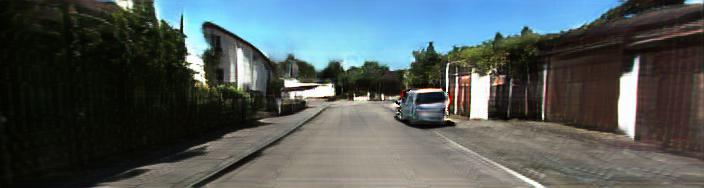}&
      \includegraphics[width=\mywidth\linewidth]{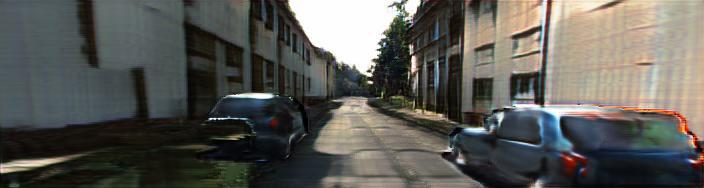}  \vspace{\reduceheight}\\

      \includegraphics[width=\mywidth\linewidth]{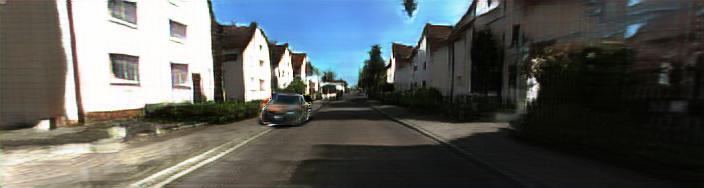}&
      \includegraphics[width=\mywidth\linewidth]{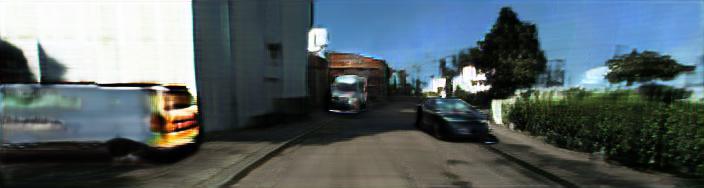}&
      \includegraphics[width=\mywidth\linewidth]{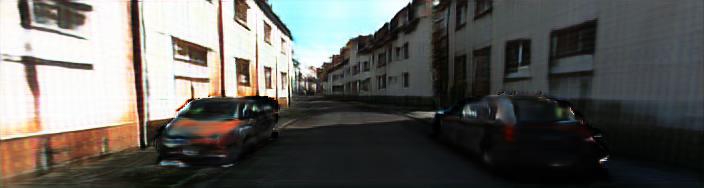}&
      \includegraphics[width=\mywidth\linewidth]{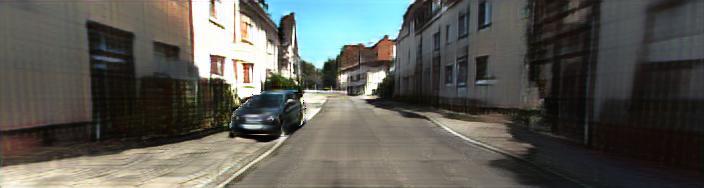}&
      \includegraphics[width=\mywidth\linewidth]{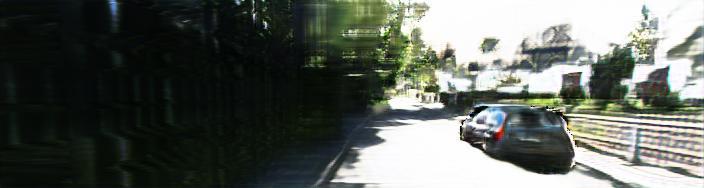}  \vspace{\reduceheight}\\

      \includegraphics[width=\mywidth\linewidth]{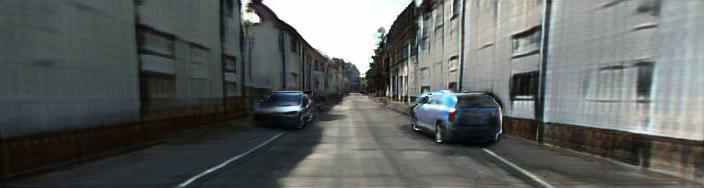}&
      \includegraphics[width=\mywidth\linewidth]{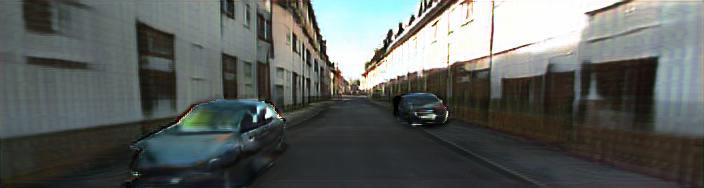}&
      \includegraphics[width=\mywidth\linewidth]{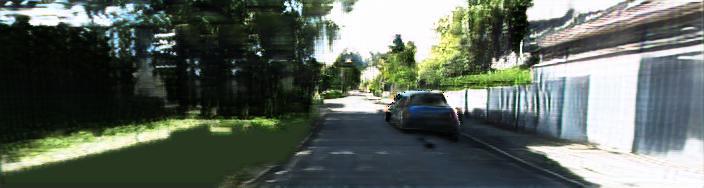}&
      \includegraphics[width=\mywidth\linewidth]{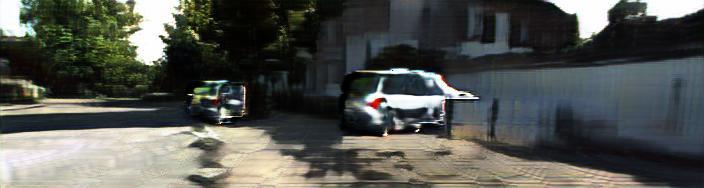}&
      \includegraphics[width=\mywidth\linewidth]{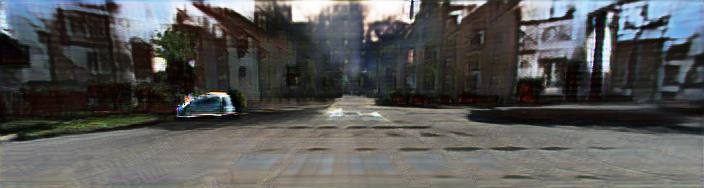}  \vspace{\reduceheight}\\

      \includegraphics[width=\mywidth\linewidth]{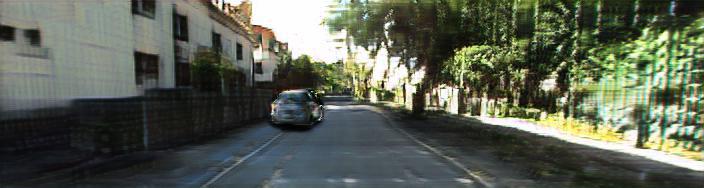}&
      \includegraphics[width=\mywidth\linewidth]{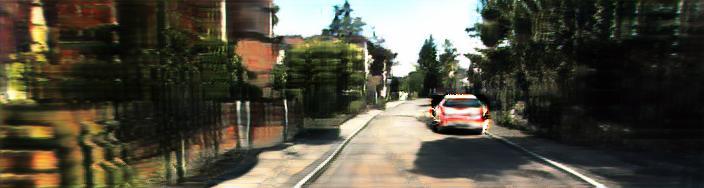}&
      \includegraphics[width=\mywidth\linewidth]{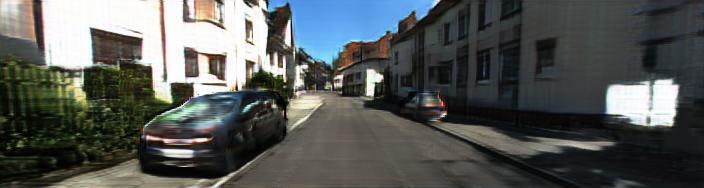}&
      \includegraphics[width=\mywidth\linewidth]{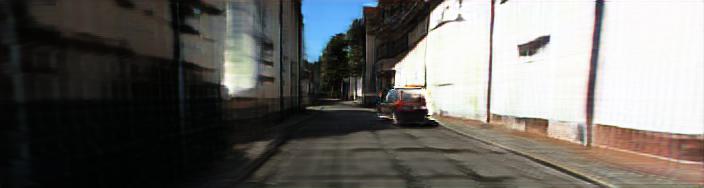}&
      \includegraphics[width=\mywidth\linewidth]{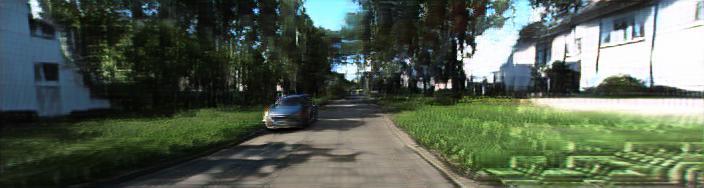}  \vspace{\reduceheight}\\

      \includegraphics[width=\mywidth\linewidth]{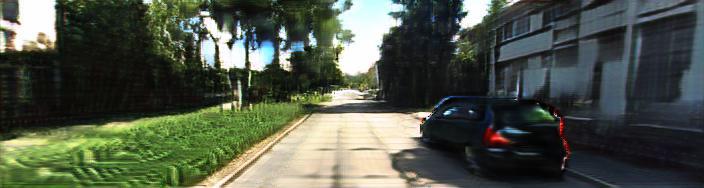}&
      \includegraphics[width=\mywidth\linewidth]{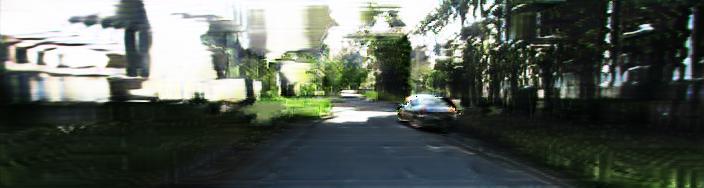}&
      \includegraphics[width=\mywidth\linewidth]{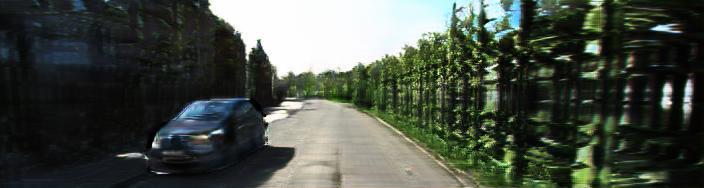}&
      \includegraphics[width=\mywidth\linewidth]{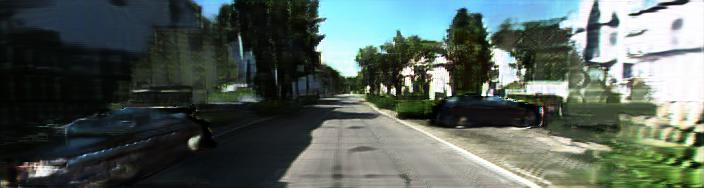}&
      \includegraphics[width=\mywidth\linewidth]{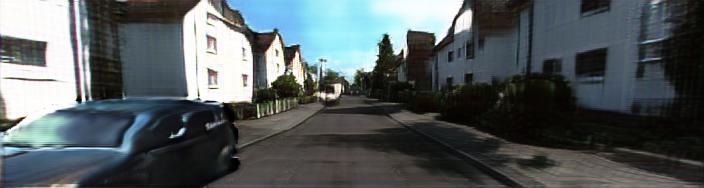}  \vspace{\reduceheight}\\

      \includegraphics[width=\mywidth\linewidth]{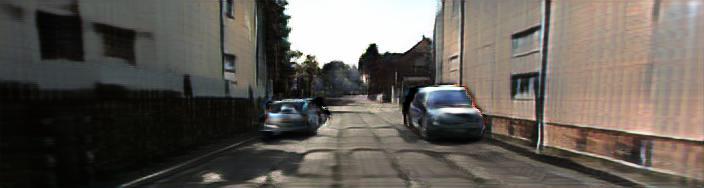}&
      \includegraphics[width=\mywidth\linewidth]{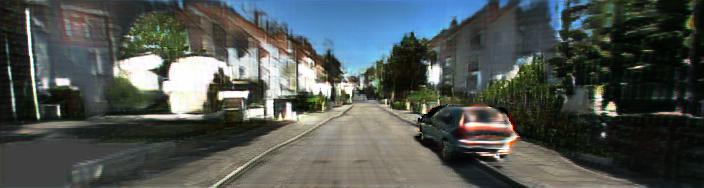}&
      \includegraphics[width=\mywidth\linewidth]{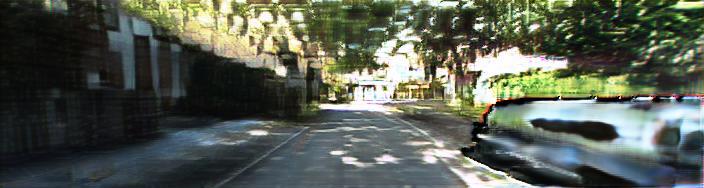}&
      \includegraphics[width=\mywidth\linewidth]{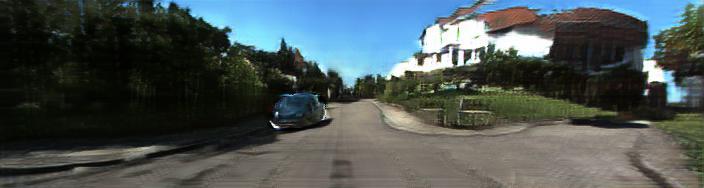}&
      \includegraphics[width=\mywidth\linewidth]{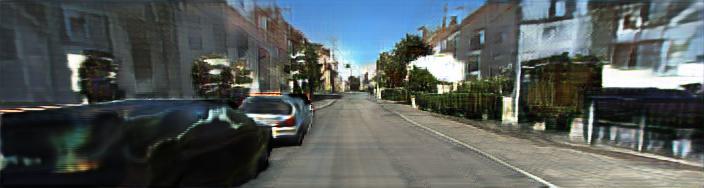}  \vspace{\reduceheight}\\

      \includegraphics[width=\mywidth\linewidth]{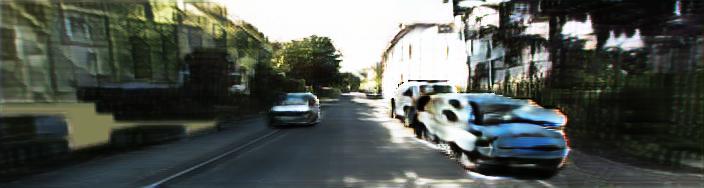}&
      \includegraphics[width=\mywidth\linewidth]{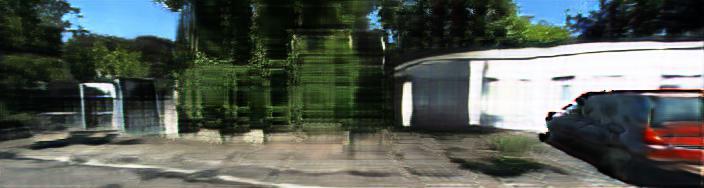}&
      \includegraphics[width=\mywidth\linewidth]{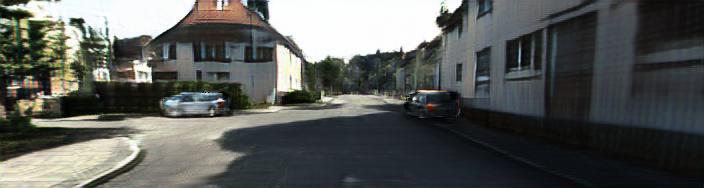}&
      \includegraphics[width=\mywidth\linewidth]{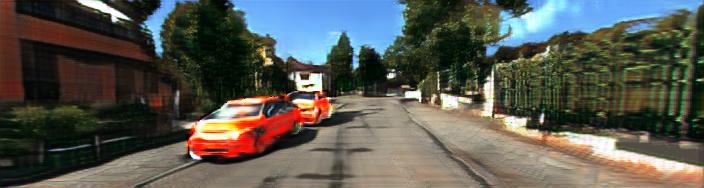}&
      \includegraphics[width=\mywidth\linewidth]{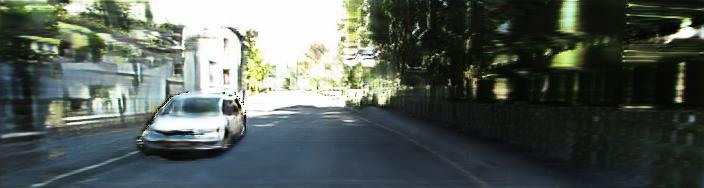}  \vspace{\reduceheight}\\

      \includegraphics[width=\mywidth\linewidth]{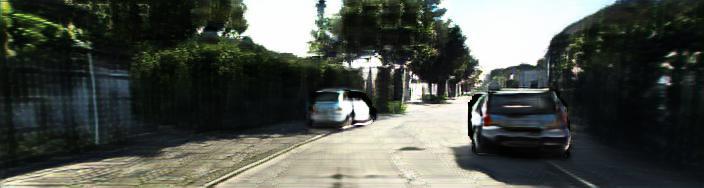}&
      \includegraphics[width=\mywidth\linewidth]{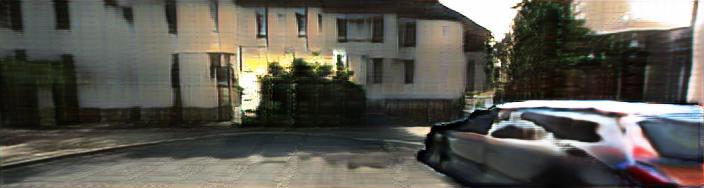}&
      \includegraphics[width=\mywidth\linewidth]{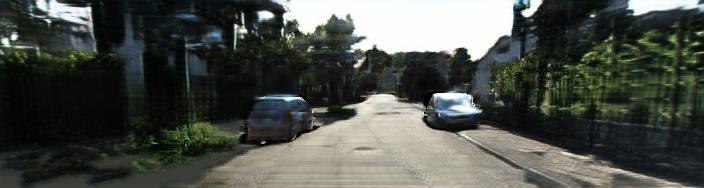}&
      \includegraphics[width=\mywidth\linewidth]{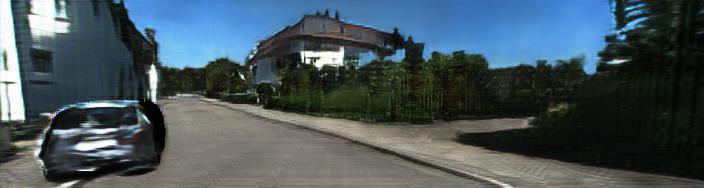}&
      \includegraphics[width=\mywidth\linewidth]{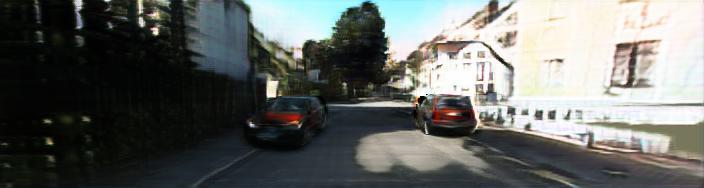}  \vspace{\reduceheight}\\

      \includegraphics[width=\mywidth\linewidth]{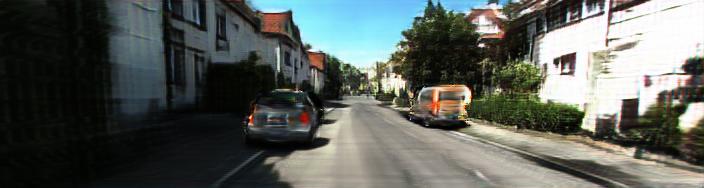}&
      \includegraphics[width=\mywidth\linewidth]{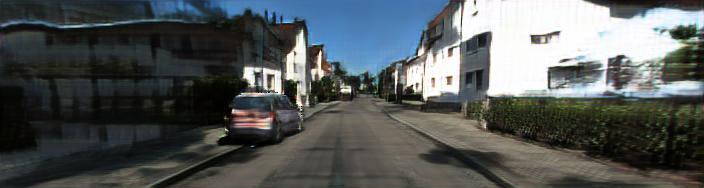}&
      \includegraphics[width=\mywidth\linewidth]{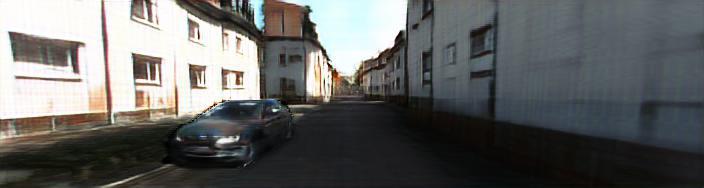}&
      \includegraphics[width=\mywidth\linewidth]{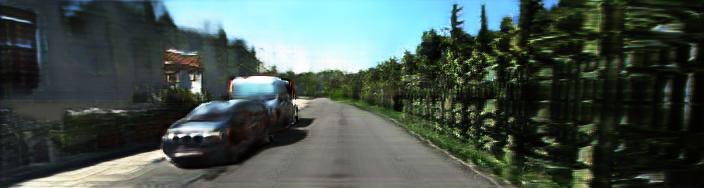}&
      \includegraphics[width=\mywidth\linewidth]{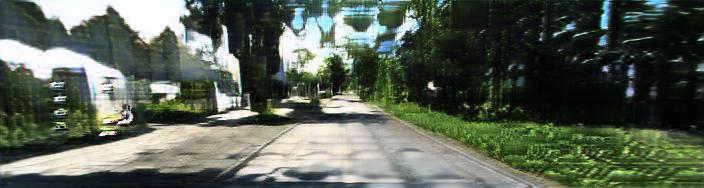}  \vspace{\reduceheight}\\
      \includegraphics[width=\mywidth\linewidth]{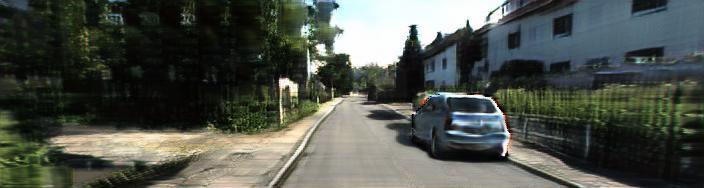}&
      \includegraphics[width=\mywidth\linewidth]{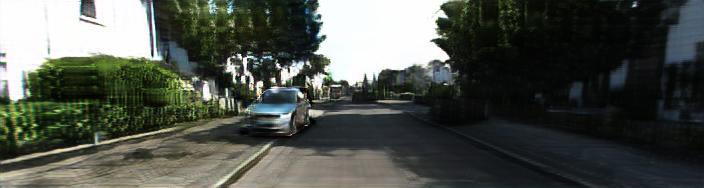}&
      \includegraphics[width=\mywidth\linewidth]{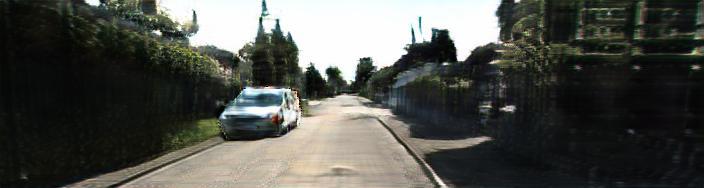}&
      \includegraphics[width=\mywidth\linewidth]{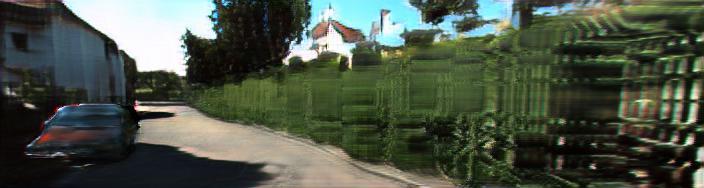}&
      \includegraphics[width=\mywidth\linewidth]{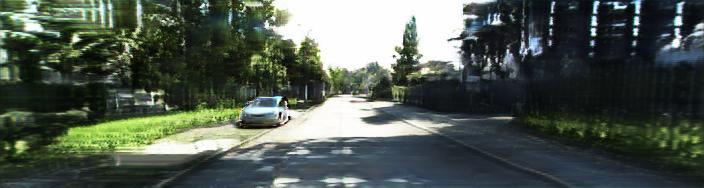}  \vspace{\reduceheight}\\
     \end{tabular}\vspace{-0.1cm}}

     \subfloat[][CLEVR-W\label{fig:random_clevr}\vspace{-0.2cm}]{
     \begin{tabular}{cccccccccc}
      \includegraphics[width=\mywidthclevr\linewidth]{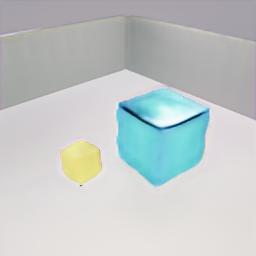} &
      \includegraphics[width=\mywidthclevr\linewidth]{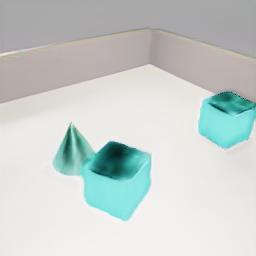} &
      \includegraphics[width=\mywidthclevr\linewidth]{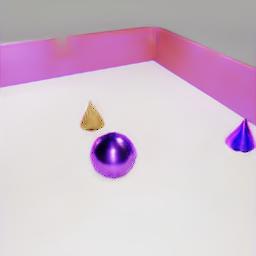} &
      \includegraphics[width=\mywidthclevr\linewidth]{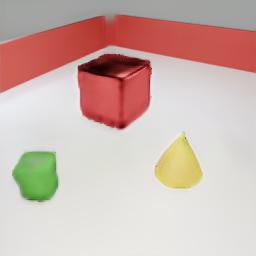} &
      \includegraphics[width=\mywidthclevr\linewidth]{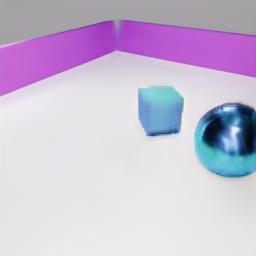} &
      \includegraphics[width=\mywidthclevr\linewidth]{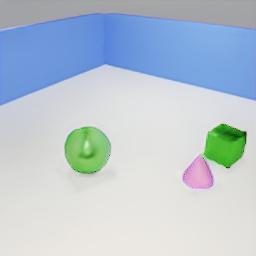} &      
      \includegraphics[width=\mywidthclevr\linewidth]{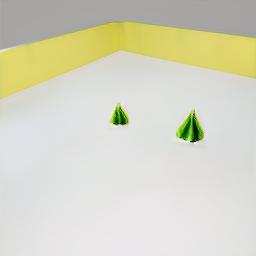} &
      \includegraphics[width=\mywidthclevr\linewidth]{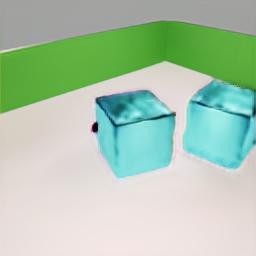} &
      \includegraphics[width=\mywidthclevr\linewidth]{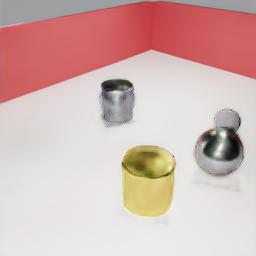} &      
      \includegraphics[width=\mywidthclevr\linewidth]{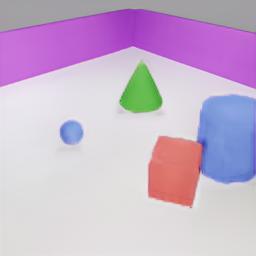}  \vspace{\reduceheight}\\
      \includegraphics[width=\mywidthclevr\linewidth]{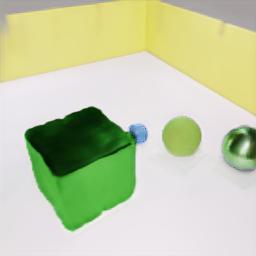} &
      \includegraphics[width=\mywidthclevr\linewidth]{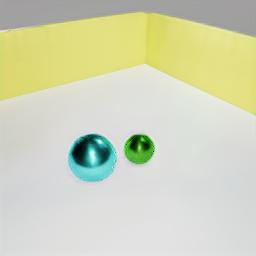} &
      \includegraphics[width=\mywidthclevr\linewidth]{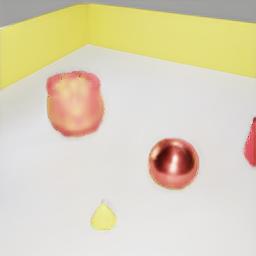} &
      \includegraphics[width=\mywidthclevr\linewidth]{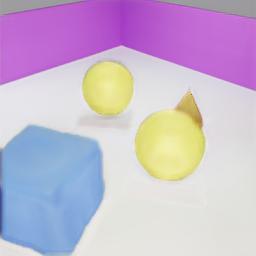} &
      \includegraphics[width=\mywidthclevr\linewidth]{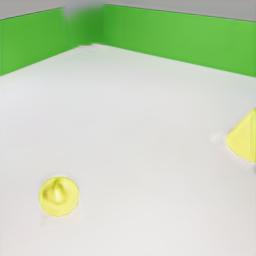} &
      \includegraphics[width=\mywidthclevr\linewidth]{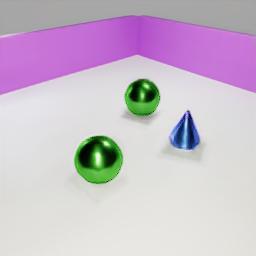} &      
      \includegraphics[width=\mywidthclevr\linewidth]{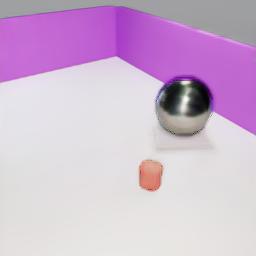} &
      \includegraphics[width=\mywidthclevr\linewidth]{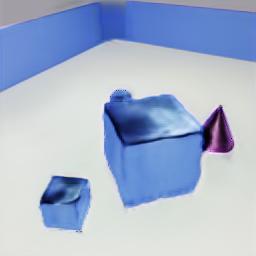} &
      \includegraphics[width=\mywidthclevr\linewidth]{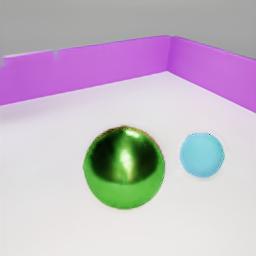} &      
      \includegraphics[width=\mywidthclevr\linewidth]{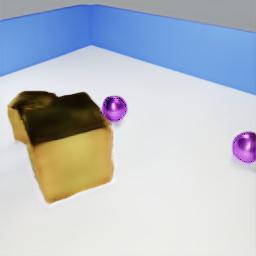}  \vspace{\reduceheight}\\
      \includegraphics[width=\mywidthclevr\linewidth]{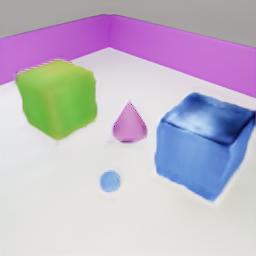} &
      \includegraphics[width=\mywidthclevr\linewidth]{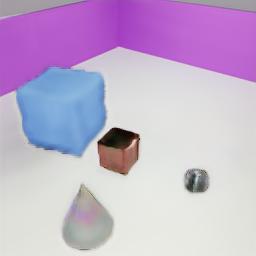} &
      \includegraphics[width=\mywidthclevr\linewidth]{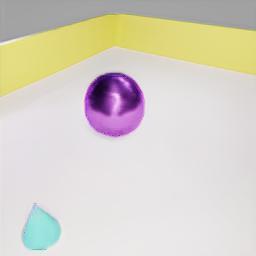} &
      \includegraphics[width=\mywidthclevr\linewidth]{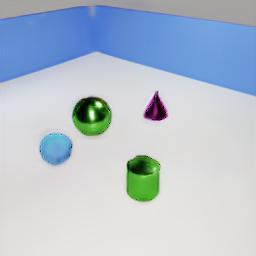} &
      \includegraphics[width=\mywidthclevr\linewidth]{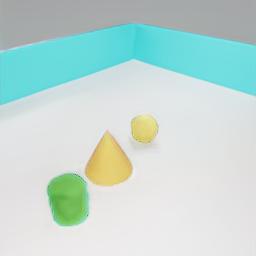} &
      \includegraphics[width=\mywidthclevr\linewidth]{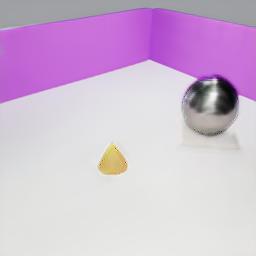} &      
      \includegraphics[width=\mywidthclevr\linewidth]{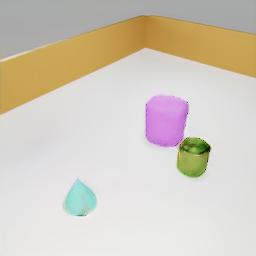} &
      \includegraphics[width=\mywidthclevr\linewidth]{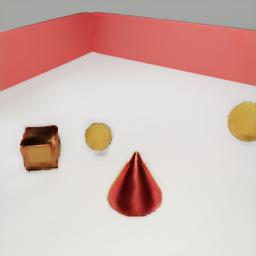} &
      \includegraphics[width=\mywidthclevr\linewidth]{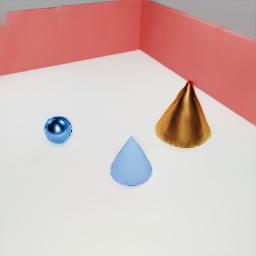} &      
      \includegraphics[width=\mywidthclevr\linewidth]{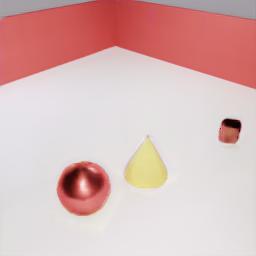}  \vspace{\reduceheight}\\
      \includegraphics[width=\mywidthclevr\linewidth]{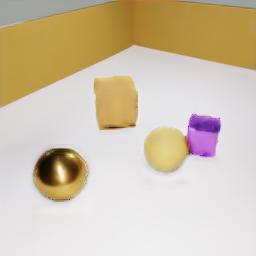} &
      \includegraphics[width=\mywidthclevr\linewidth]{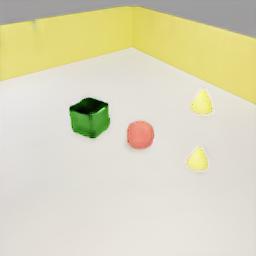} &
      \includegraphics[width=\mywidthclevr\linewidth]{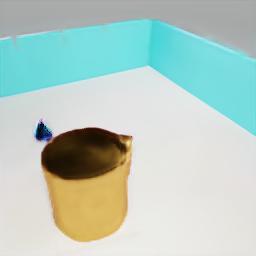} &
      \includegraphics[width=\mywidthclevr\linewidth]{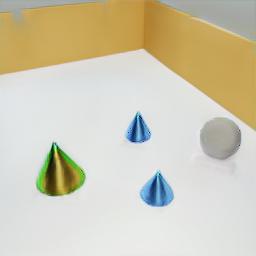} &
      \includegraphics[width=\mywidthclevr\linewidth]{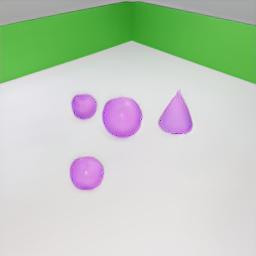} &
      \includegraphics[width=\mywidthclevr\linewidth]{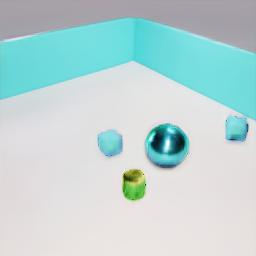} &      
      \includegraphics[width=\mywidthclevr\linewidth]{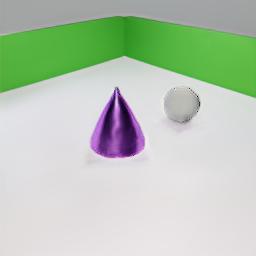} &
      \includegraphics[width=\mywidthclevr\linewidth]{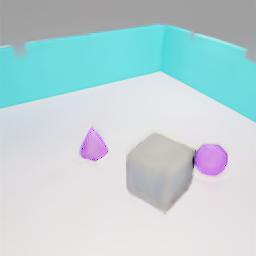} &
      \includegraphics[width=\mywidthclevr\linewidth]{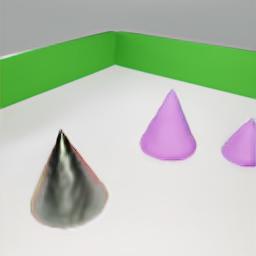} &      
      \includegraphics[width=\mywidthclevr\linewidth]{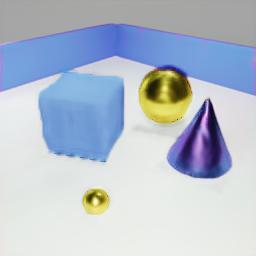}  \vspace{\reduceheight}\\
      \includegraphics[width=\mywidthclevr\linewidth]{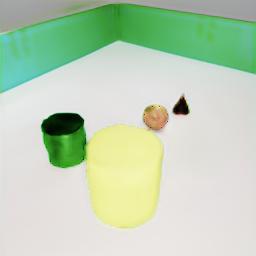} &
      \includegraphics[width=\mywidthclevr\linewidth]{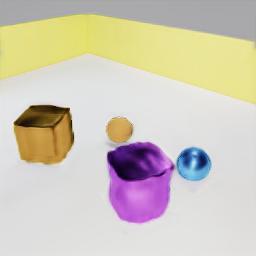} &
      \includegraphics[width=\mywidthclevr\linewidth]{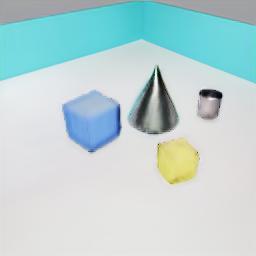} &
      \includegraphics[width=\mywidthclevr\linewidth]{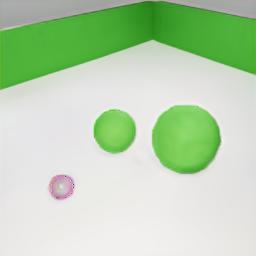} &
      \includegraphics[width=\mywidthclevr\linewidth]{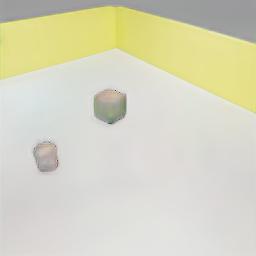} &
      \includegraphics[width=\mywidthclevr\linewidth]{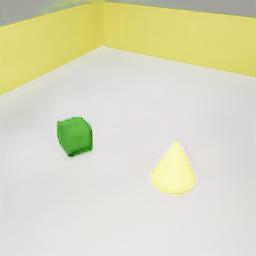} &      
      \includegraphics[width=\mywidthclevr\linewidth]{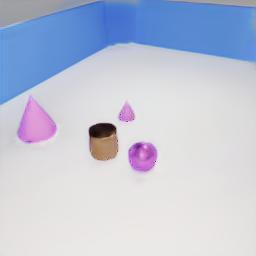} &
      \includegraphics[width=\mywidthclevr\linewidth]{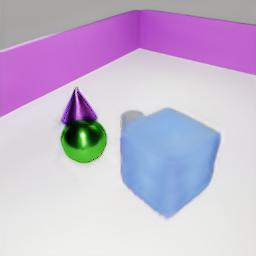} &
      \includegraphics[width=\mywidthclevr\linewidth]{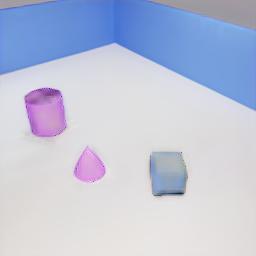} &      
      \includegraphics[width=\mywidthclevr\linewidth]{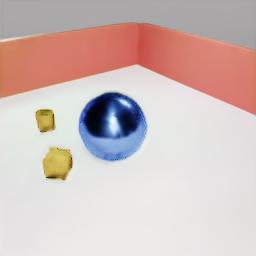} \vspace{\reduceheight}\\
      \includegraphics[width=\mywidthclevr\linewidth]{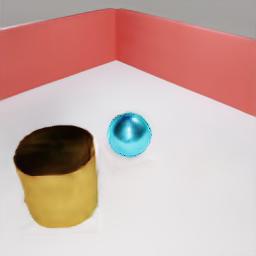} &
      \includegraphics[width=\mywidthclevr\linewidth]{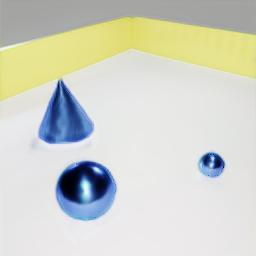} &
      \includegraphics[width=\mywidthclevr\linewidth]{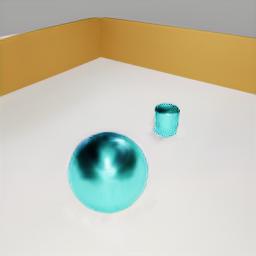} &
      \includegraphics[width=\mywidthclevr\linewidth]{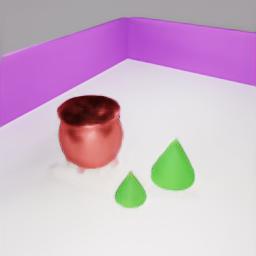} &
      \includegraphics[width=\mywidthclevr\linewidth]{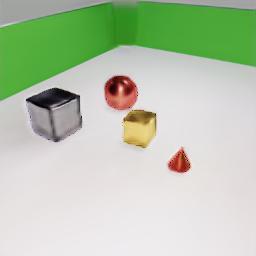} &
      \includegraphics[width=\mywidthclevr\linewidth]{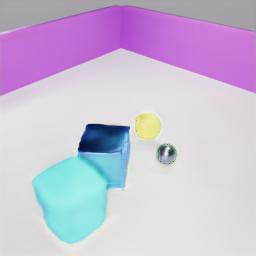} &      
      \includegraphics[width=\mywidthclevr\linewidth]{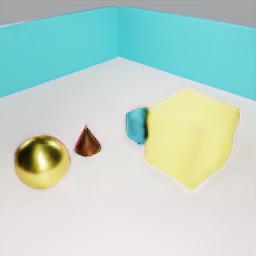} &
      \includegraphics[width=\mywidthclevr\linewidth]{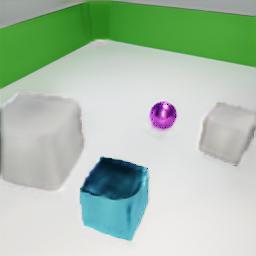} &
      \includegraphics[width=\mywidthclevr\linewidth]{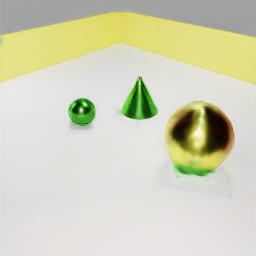} &      
      \includegraphics[width=\mywidthclevr\linewidth]{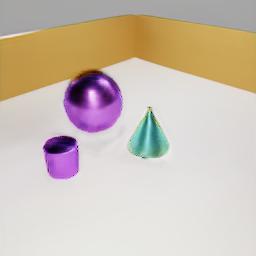} \vspace{\reduceheight}\\
     \end{tabular}\vspace{-0.1cm}}
     \caption{{\bf Uncurated Samples}. We show randomly sampled images of UrbanGIRAFFE.}
     \label{fig:random}
     \vspace{-0.3cm}
    \end{figure*}

%% file: main.bbl
\begin{thebibliography}{10}\itemsep=-1pt

\bibitem{Bautista2022NEURIPS}
Miguel~{\'{A}}ngel Bautista, Pengsheng Guo, Samira Abnar, Walter Talbott,
  Alexander Toshev, Zhuoyuan Chen, Laurent Dinh, Shuangfei Zhai, Hanlin Goh,
  Daniel Ulbricht, Afshin Dehghan, and Josh~M. Susskind.
\newblock {GAUDI:} {A} neural architect for immersive 3d scene generation.
\newblock {\em Advances in Neural Information Processing Systems (NeurIPS)},
  2022.

\bibitem{Bautista2022ARXIV}
Miguel~{\'{A}}ngel Bautista, Pengsheng Guo, Samira Abnar, Walter Talbott,
  Alexander Toshev, Zhuoyuan Chen, Laurent Dinh, Shuangfei Zhai, Hanlin Goh,
  Daniel Ulbricht, Afshin Dehghan, and Josh~M. Susskind.
\newblock {GAUDI:} {A} neural architect for immersive 3d scene generation.
\newblock {\em arXiv.org}, 2022.

\bibitem{Bergman2022ARXIV}
Alexander~W. Bergman, Petr Kellnhofer, Yifan Wang, Eric~R. Chan, David~B.
  Lindell, and Gordon Wetzstein.
\newblock Generative neural articulated radiance fields.
\newblock {\em arXiv.org}, 2022.

\bibitem{Binkowski2018ICLR}
Mikolaj Binkowski, Dougal~J. Sutherland, Michael Arbel, and Arthur Gretton.
\newblock Demystifying {MMD} gans.
\newblock In {\em Proc. of the International Conf. on Learning Representations
  (ICLR)}, 2018.

\bibitem{Cai2022CVPR}
Shengqu Cai, Anton Obukhov, Dengxin Dai, and Luc~Van Gool.
\newblock Pix2nerf: Unsupervised conditional
  {\textdollar}{\textbackslash}pi{\textdollar}-gan for single image to neural
  radiance fields translation.
\newblock In {\em Proc. IEEE Conf. on Computer Vision and Pattern Recognition
  (CVPR)}, 2022.

\bibitem{Cao2022CVPR}
Anh{-}Quan Cao and Raoul de Charette.
\newblock Monoscene: Monocular 3d semantic scene completion.
\newblock In {\em Proc. IEEE Conf. on Computer Vision and Pattern Recognition
  (CVPR)}, 2022.

\bibitem{Chan2022CVPR}
Eric~R. Chan, Connor~Z. Lin, Matthew~A. Chan, Koki Nagano, Boxiao Pan,
  Shalini~De Mello, Orazio Gallo, Leonidas~J. Guibas, Jonathan Tremblay, Sameh
  Khamis, Tero Karras, and Gordon Wetzstein.
\newblock Efficient geometry-aware 3d generative adversarial networks.
\newblock In {\em Proc. IEEE Conf. on Computer Vision and Pattern Recognition
  (CVPR)}, 2022.

\bibitem{Chan2021CVPR}
Eric~R. Chan, Marco Monteiro, Petr Kellnhofer, Jiajun Wu, and Gordon Wetzstein.
\newblock Pi-gan: Periodic implicit generative adversarial networks for
  3d-aware image synthesis.
\newblock In {\em Proc. IEEE Conf. on Computer Vision and Pattern Recognition
  (CVPR)}, 2021.

\bibitem{Chen2021ICCVmvs}
Anpei Chen, Zexiang Xu, Fuqiang Zhao, Xiaoshuai Zhang, Fanbo Xiang, Jingyi Yu,
  and Hao Su.
\newblock Mvsnerf: Fast generalizable radiance field reconstruction from
  multi-view stereo.
\newblock In {\em Proc. of the IEEE International Conf. on Computer Vision
  (ICCV)}, 2021.

\bibitem{Deng2022CVPR}
Yu Deng, Jiaolong Yang, Jianfeng Xiang, and Xin Tong.
\newblock {GRAM:} generative radiance manifolds for 3d-aware image generation.
\newblock In {\em Proc. IEEE Conf. on Computer Vision and Pattern Recognition
  (CVPR)}, 2022.

\bibitem{DeVries2021ICCV}
Terrance DeVries, Miguel~{\'{A}}ngel Bautista, Nitish Srivastava, Graham~W.
  Taylor, and Joshua~M. Susskind.
\newblock Unconstrained scene generation with locally conditioned radiance
  fields.
\newblock In {\em Proc. of the IEEE International Conf. on Computer Vision
  (ICCV)}, 2021.

\bibitem{Epstein2022ECCV}
Dave Epstein, Taesung Park, Richard Zhang, Eli Shechtman, and Alexei~A. Efros.
\newblock Blobgan: Spatially disentangled scene representations.
\newblock In Shai Avidan, Gabriel~J. Brostow, Moustapha Ciss{\'{e}},
  Giovanni~Maria Farinella, and Tal Hassner, editors, {\em Proc. of the
  European Conf. on Computer Vision (ECCV)}, 2022.

\bibitem{Fu2022THREEDV}
Xiao Fu, Shangzhan Zhang, Tianrun Chen, Yichong Lu, Lanyun Zhu, Xiaowei Zhou,
  Andreas Geiger, and Yiyi Liao.
\newblock Panoptic nerf: 3d-to-2d label transfer for panoptic urban scene
  segmentation.
\newblock 2022.

\bibitem{Gadde2021ICCV}
Raghudeep Gadde, Qianli Feng, and Aleix~M. Mart{\'{\i}}nez.
\newblock Detail me more: Improving gan's photo-realism of complex scenes.
\newblock In {\em Proc. of the IEEE International Conf. on Computer Vision
  (ICCV)}, 2021.

\bibitem{Goodfellow2014NIPS}
Ian~J. Goodfellow, Jean Pouget{-}Abadie, Mehdi Mirza, Bing Xu, David
  Warde{-}Farley, Sherjil Ozair, Aaron~C. Courville, and Yoshua Bengio.
\newblock Generative adversarial nets.
\newblock In {\em Advances in Neural Information Processing Systems (NIPS)},
  2014.

\bibitem{gu2021stylenerf}
Jiatao Gu, Lingjie Liu, Peng Wang, and Christian Theobalt.
\newblock Stylenerf: A style-based 3d-aware generator for high-resolution image
  synthesis.
\newblock In {\em Proc. of the International Conf. on Learning Representations
  (ICLR)}, 2022.

\bibitem{Hao2021ICCV}
Zekun Hao, Arun Mallya, Serge~J. Belongie, and Ming{-}Yu Liu.
\newblock Gancraft: Unsupervised 3d neural rendering of minecraft worlds.
\newblock In {\em Proc. of the IEEE International Conf. on Computer Vision
  (ICCV)}, 2021.

\bibitem{He2016CVPR}
Kaiming He, Xiangyu Zhang, Shaoqing Ren, and Jian Sun.
\newblock Deep residual learning for image recognition.
\newblock In {\em Proc. IEEE Conf. on Computer Vision and Pattern Recognition
  (CVPR)}, 2016.

\bibitem{He2021CVPR}
Sen He, Wentong Liao, Michael~Ying Yang, Yongxin Yang, Yi{-}Zhe Song, Bodo
  Rosenhahn, and Tao Xiang.
\newblock Context-aware layout to image generation with enhanced object
  appearance.
\newblock In {\em Proc. IEEE Conf. on Computer Vision and Pattern Recognition
  (CVPR)}, 2021.

\bibitem{Henzler2019ICCV}
Philipp Henzler, Niloy~J Mitra, , and Tobias Ritschel.
\newblock Escaping plato's cave: 3d shape from adversarial rendering.
\newblock In {\em Proc. of the IEEE International Conf. on Computer Vision
  (ICCV)}, 2019.

\bibitem{Heusel2017NIPS}
Martin Heusel, Hubert Ramsauer, Thomas Unterthiner, Bernhard Nessler, and Sepp
  Hochreiter.
\newblock Gans trained by a two time-scale update rule converge to a local nash
  equilibrium.
\newblock In {\em Advances in Neural Information Processing Systems (NIPS)},
  2017.

\bibitem{Hong2022ARXIV}
Fangzhou Hong, Zhaoxi Chen, Yushi Lan, Liang Pan, and Ziwei Liu.
\newblock {EVA3D:} compositional 3d human generation from 2d image collections.
\newblock {\em arXiv.org}, 2022.

\bibitem{Isola2017CVPR}
Phillip Isola, Jun{-}Yan Zhu, Tinghui Zhou, and Alexei~A. Efros.
\newblock Image-to-image translation with conditional adversarial networks.
\newblock In {\em Proc. IEEE Conf. on Computer Vision and Pattern Recognition
  (CVPR)}, 2017.

\bibitem{Johnson2016ECCV}
Justin Johnson, Alexandre Alahi, and Li Fei{-}Fei.
\newblock Perceptual losses for real-time style transfer and super-resolution.
\newblock In {\em Proc. of the European Conf. on Computer Vision (ECCV)}, 2016.

\bibitem{Johnson2017CVPR}
Justin Johnson, Bharath Hariharan, Laurens van~der Maaten, Li Fei-Fei, C
  Lawrence~Zitnick, and Ross Girshick.
\newblock Clevr: A diagnostic dataset for compositional language and elementary
  visual reasoning.
\newblock In {\em Proc. IEEE Conf. on Computer Vision and Pattern Recognition
  (CVPR)}, 2017.

\bibitem{Karras2020NeurIPS}
Tero Karras, Miika Aittala, Janne Hellsten, Samuli Laine, Jaakko Lehtinen, and
  Timo Aila.
\newblock Training generative adversarial networks with limited data.
\newblock In {\em Advances in Neural Information Processing Systems (NeurIPS)},
  2020.

\bibitem{Karras2021NIPS}
Tero Karras, Miika Aittala, Samuli Laine, Erik H{\"{a}}rk{\"{o}}nen, Janne
  Hellsten, Jaakko Lehtinen, and Timo Aila.
\newblock Alias-free generative adversarial networks.
\newblock In {\em NIPS}, 2021.

\bibitem{Karras2019CVPR}
Tero Karras, Samuli Laine, and Timo Aila.
\newblock A style-based generator architecture for generative adversarial
  networks.
\newblock In {\em Proc. IEEE Conf. on Computer Vision and Pattern Recognition
  (CVPR)}, 2019.

\bibitem{Karras2020CVPRa}
Tero Karras, Samuli Laine, Miika Aittala, Janne Hellsten, Jaakko Lehtinen, and
  Timo Aila.
\newblock Analyzing and improving the image quality of {StyleGAN}.
\newblock In {\em Proc. IEEE Conf. on Computer Vision and Pattern Recognition
  (CVPR)}, 2020.

\bibitem{Karras2020CVPR}
Tero Karras, Samuli Laine, Miika Aittala, Janne Hellsten, Jaakko Lehtinen, and
  Timo Aila.
\newblock Analyzing and improving the image quality of {StyleGAN}.
\newblock 2020.

\bibitem{Kundu2022CVPR}
Abhijit Kundu, Kyle Genova, Xiaoqi Yin, Alireza Fathi, Caroline Pantofaru,
  Leonidas~J. Guibas, Andrea Tagliasacchi, Frank Dellaert, and Thomas~A.
  Funkhouser.
\newblock Panoptic neural fields: {A} semantic object-aware neural scene
  representation.
\newblock In {\em Proc. IEEE Conf. on Computer Vision and Pattern Recognition
  (CVPR)}, 2022.

\bibitem{Lee2022ARXIV}
Minsoo Lee, Chaeyeon Chung, Hojun Cho, Min{-}Jung Kim, Sanghun Jung, Jaegul
  Choo, and Minhyuk Sung.
\newblock 3d-gif: 3d-controllable object generation via implicit factorized
  representations.
\newblock {\em arXiv.org}, 2022.

\bibitem{Liao2020CVPRa}
Yiyi Liao, Katja Schwarz, Lars~M. Mescheder, and Andreas Geiger.
\newblock Towards unsupervised learning of generative models for 3d
  controllable image synthesis.
\newblock {\em Proc. IEEE Conf. on Computer Vision and Pattern Recognition
  (CVPR)}, 2020.

\bibitem{Liao2022PAMI}
Yiyi Liao, Jun Xie, and Andreas Geiger.
\newblock Kitti-360: A novel dataset and benchmarks for urban scene
  understanding in 2d and 3d.
\newblock In {\em IEEE Trans. on Pattern Analysis and Machine Intelligence
  (PAMI)}, 2022.

\bibitem{Liu2020NIPS}
Lingjie Liu, Jiatao Gu, Kyaw~Zaw Lin, Tat{-}Seng Chua, and Christian Theobalt.
\newblock Neural sparse voxel fields.
\newblock In Hugo Larochelle, Marc'Aurelio Ranzato, Raia Hadsell,
  Maria{-}Florina Balcan, and Hsuan{-}Tien Lin, editors, {\em Advances in
  Neural Information Processing Systems (NIPS)}, 2020.

\bibitem{Mescheder2018ICML}
Lars Mescheder, Andreas Geiger, and Sebastian Nowozin.
\newblock Which training methods for gans do actually converge?
\newblock In {\em Proc. of the International Conf. on Machine learning (ICML)},
  2018.

\bibitem{Mescheder2019CVPR}
Lars Mescheder, Michael Oechsle, Michael Niemeyer, Sebastian Nowozin, and
  Andreas Geiger.
\newblock Occupancy networks: Learning 3d reconstruction in function space.
\newblock In {\em Proc. IEEE Conf. on Computer Vision and Pattern Recognition
  (CVPR)}, 2019.

\bibitem{Mildenhall2020ECCV}
Ben Mildenhall, Pratul~P Srinivasan, Matthew Tancik, Jonathan~T Barron, Ravi
  Ramamoorthi, and Ren Ng.
\newblock {NeRF}: Representing scenes as neural radiance fields for view
  synthesis.
\newblock In {\em Proc. of the European Conf. on Computer Vision (ECCV)}, 2020.

\bibitem{Muller2022CVPR}
Norman M{\"{u}}ller, Andrea Simonelli, Lorenzo Porzi, Samuel~Rota Bul{\`{o}},
  Matthias Nie{\ss}ner, and Peter Kontschieder.
\newblock Autorf: Learning 3d object radiance fields from single view
  observations.
\newblock In {\em Proc. IEEE Conf. on Computer Vision and Pattern Recognition
  (CVPR)}, 2022.

\bibitem{mueller2022instant}
Thomas M\"uller, Alex Evans, Christoph Schied, and Alexander Keller.
\newblock Instant neural graphics primitives with a multiresolution hash
  encoding.
\newblock {\em arXiv.org}, Jan. 2022.

\bibitem{Nguyen-Phuoc2019ICCV}
Thu Nguyen{-}Phuoc, Chuan Li, Lucas Theis, Christian Richardt, and Yong{-}Liang
  Yang.
\newblock Hologan: Unsupervised learning of 3d representations from natural
  images.
\newblock In {\em Proc. of the IEEE International Conf. on Computer Vision
  (ICCV)}, 2019.

\bibitem{Nguyen-Phuoc2020NEURIPS}
Thu Nguyen{-}Phuoc, Christian Richardt, Long Mai, Yong{-}Liang Yang, and Niloy
  Mitra.
\newblock Blockgan: Learning 3d object-aware scene representations from
  unlabelled images.
\newblock In {\em Advances in Neural Information Processing Systems (NeurIPS)},
  2020.

\bibitem{Niemeyer2020ARXIV}
Michael Niemeyer and Andreas Geiger.
\newblock Giraffe: Representing scenes as compositional generative neural
  feature fields.
\newblock In {\em arXiv.org}, volume 2011.12100, 2020.

\bibitem{Niemeyer2021CVPR}
Michael Niemeyer and Andreas Geiger.
\newblock {GIRAFFE:} representing scenes as compositional generative neural
  feature fields.
\newblock In {\em Proc. IEEE Conf. on Computer Vision and Pattern Recognition
  (CVPR)}, 2021.

\bibitem{Noguchi2022ECCV}
Atsuhiro Noguchi, Xiao Sun, Stephen Lin, and Tatsuya Harada.
\newblock Unsupervised learning of efficient geometry-aware neural articulated
  representations.
\newblock In {\em Proc. of the European Conf. on Computer Vision (ECCV)}, 2022.

\bibitem{Ost2021CVPR}
Julian Ost, Fahim Mannan, Nils Thuerey, Julian Knodt, and Felix Heide.
\newblock Neural scene graphs for dynamic scenes.
\newblock {\em Proc. IEEE Conf. on Computer Vision and Pattern Recognition
  (CVPR)}, 2021.

\bibitem{Park2019CVPR}
Jeong~Joon Park, Peter Florence, Julian Straub, Richard~A. Newcombe, and Steven
  Lovegrove.
\newblock Deepsdf: Learning continuous signed distance functions for shape
  representation.
\newblock In {\em Proc. IEEE Conf. on Computer Vision and Pattern Recognition
  (CVPR)}, 2019.

\bibitem{Park2019CVPRa}
Taesung Park, Ming-Yu Liu, Ting-Chun Wang, and Jun-Yan Zhu.
\newblock Semantic image synthesis with spatially-adaptive normalization.
\newblock In {\em Proc. IEEE Conf. on Computer Vision and Pattern Recognition
  (CVPR)}, 2019.

\bibitem{Reiser2021ICCV}
Christian Reiser, Songyou Peng, Yiyi Liao, and Andreas Geiger.
\newblock Kilonerf: Speeding up neural radiance fields with thousands of tiny
  mlps.
\newblock In {\em Proc. of the IEEE International Conf. on Computer Vision
  (ICCV)}, 2021.

\bibitem{Rematas2022CVPR}
Konstantinos Rematas, Andrew Liu, Pratul~P. Srinivasan, Jonathan~T. Barron,
  Andrea Tagliasacchi, Thomas~A. Funkhouser, and Vittorio Ferrari.
\newblock Urban radiance fields.
\newblock In {\em Proc. IEEE Conf. on Computer Vision and Pattern Recognition
  (CVPR)}, 2022.

\bibitem{SauerS022SIGGRAPH}
Axel Sauer, Katja Schwarz, and Andreas Geiger.
\newblock Stylegan-xl: Scaling stylegan to large diverse datasets.
\newblock In {\em ACM Trans. on Graphics}, 2022.

\bibitem{Schonfeld2021ICLR}
Edgar Sch{\"{o}}nfeld, Vadim Sushko, Dan Zhang, Juergen Gall, Bernt Schiele,
  and Anna Khoreva.
\newblock You only need adversarial supervision for semantic image synthesis.
\newblock In {\em Proc. of the International Conf. on Learning Representations
  (ICLR)}, 2021.

\bibitem{Schwarz2020NIPS}
Katja Schwarz, Yiyi Liao, Michael Niemeyer, and Andreas Geiger.
\newblock {GRAF:} generative radiance fields for 3d-aware image synthesis.
\newblock {\em Advances in Neural Information Processing Systems (NIPS)}, 2020.

\bibitem{Schwarz2022NEURIPS}
Katja Schwarz, Axel Sauer, Michael Niemeyer, Yiyi Liao, and Andreas Geiger.
\newblock Voxgraf: Fast 3d-aware image synthesis with sparse voxel grids.
\newblock {\em Advances in Neural Information Processing Systems (NeurIPS)},
  2022.

\bibitem{Sun2022TOG}
Jingxiang Sun, Xuan Wang, Yichun Shi, Lizhen Wang, Jue Wang, and Yebin Liu.
\newblock {IDE-3D:} interactive disentangled editing for high-resolution
  3d-aware portrait synthesis.
\newblock {\em ACM Trans. on Graphics}, 2022.

\bibitem{Sun2022CVPR}
Jingxiang Sun, Xuan Wang, Yong Zhang, Xiaoyu Li, Qi Zhang, Yebin Liu, and Jue
  Wang.
\newblock Fenerf: Face editing in neural radiance fields.
\newblock 2022.

\bibitem{Tan2022SIGGRAPH}
Feitong Tan, Sean Fanello, Abhimitra Meka, Sergio Orts{-}Escolano, Danhang
  Tang, Rohit Pandey, Jonathan Taylor, Ping Tan, and Yinda Zhang.
\newblock Volux-gan: {A} generative model for 3d face synthesis with {HDRI}
  relighting.
\newblock In {\em ACM Trans. on Graphics}, 2022.

\bibitem{Tancik2022CVPR}
Matthew Tancik, Vincent Casser, Xinchen Yan, Sabeek Pradhan, Ben~P. Mildenhall,
  Pratul~P. Srinivasan, Jonathan~T. Barron, and Henrik Kretzschmar.
\newblock Block-nerf: Scalable large scene neural view synthesis.
\newblock In {\em Proc. IEEE Conf. on Computer Vision and Pattern Recognition
  (CVPR)}, 2022.

\bibitem{Wang2021NIPS}
Peng Wang, Lingjie Liu, Yuan Liu, Christian Theobalt, Taku Komura, and Wenping
  Wang.
\newblock Neus: Learning neural implicit surfaces by volume rendering for
  multi-view reconstruction.
\newblock In Marc'Aurelio Ranzato, Alina Beygelzimer, Yann~N. Dauphin, Percy
  Liang, and Jennifer~Wortman Vaughan, editors, {\em Advances in Neural
  Information Processing Systems (NIPS)}, 2021.

\bibitem{Wang2021ibrnet}
Qianqian Wang, Zhicheng Wang, Kyle Genova, Pratul Srinivasan, Howard Zhou,
  Jonathan~T. Barron, Ricardo Martin-Brualla, Noah Snavely, and Thomas
  Funkhouser.
\newblock Ibrnet: Learning multi-view image-based rendering.
\newblock In {\em CVPR}, 2021.

\bibitem{Wang2022ARXIV}
Yiming Wang, Qin Han, Marc Habermann, Kostas Daniilidis, Christian Theobalt,
  and Lingjie Liu.
\newblock Neus2: Fast learning of neural implicit surfaces for multi-view
  reconstruction.
\newblock {\em arXiv.org}, 2022.

\bibitem{Xiangli2021ARXIV}
Yuanbo Xiangli, Linning Xu, Xingang Pan, Nanxuan Zhao, Anyi Rao, Christian
  Theobalt, Bo Dai, and Dahua Lin.
\newblock Citynerf: Building nerf at city scale.
\newblock {\em arXiv.org}, abs/2112.05504, 2021.

\bibitem{Xiangli2022ECCV}
Yuanbo Xiangli, Linning Xu, Xingang Pan, Nanxuan Zhao, Anyi Rao, Christian
  Theobalt, Bo Dai, and Dahua Lin.
\newblock Bungeenerf: Progressive neural radiance field for extreme multi-scale
  scene rendering.
\newblock In {\em Proc. of the European Conf. on Computer Vision (ECCV)}, 2022.

\bibitem{Xu2022ARXIV}
Yinghao Xu, Menglei Chai, Zifan Shi, Sida Peng, Ivan Skorokhodov, Aliaksandr
  Siarohin, Ceyuan Yang, Yujun Shen, Hsin{-}Ying Lee, Bolei Zhou, and Sergey
  Tulyakov.
\newblock Discoscene: Spatially disentangled generative radiance fields for
  controllable 3d-aware scene synthesis.
\newblock {\em arXiv.org}, 2022.

\bibitem{Xu2022CVPR}
Yinghao Xu, Sida Peng, Ceyuan Yang, Yujun Shen, and Bolei Zhou.
\newblock 3d-aware image synthesis via learning structural and textural
  representations.
\newblock In {\em Proc. IEEE Conf. on Computer Vision and Pattern Recognition
  (CVPR)}, 2022.

\bibitem{Xue2022CVPR}
Yang Xue, Yuheng Li, Krishna~Kumar Singh, and Yong~Jae Lee.
\newblock {GIRAFFE} {HD:} {A} high-resolution 3d-aware generative model.
\newblock In {\em Proc. IEEE Conf. on Computer Vision and Pattern Recognition
  (CVPR)}, 2022.

\bibitem{Yang2022CVPR}
Zuopeng Yang, Daqing Liu, Chaoyue Wang, Jie Yang, and Dacheng Tao.
\newblock Modeling image composition for complex scene generation.
\newblock In {\em Proc. IEEE Conf. on Computer Vision and Pattern Recognition
  (CVPR)}, 2022.

\bibitem{yu2021pixelnerf}
Alex Yu, Vickie Ye, Matthew Tancik, and Angjoo Kanazawa.
\newblock pixelnerf: Neural radiance fields from one or few images.
\newblock In {\em Proc. IEEE Conf. on Computer Vision and Pattern Recognition
  (CVPR)}, 2021.

\bibitem{Zhang2022ECCV}
Jianfeng Zhang, Zihang Jiang, Dingdong Yang, Hongyi Xu, Yichun Shi, Guoxian
  Song, Zhongcong Xu, Xinchao Wang, and Jiashi Feng.
\newblock Avatargen: {A} 3d generative model for animatable human avatars.
\newblock In Leonid Karlinsky, Tomer Michaeli, and Ko Nishino, editors, {\em
  Proc. of the European Conf. on Computer Vision (ECCV)}, 2022.

\bibitem{Zhao2020IJCV}
Bo Zhao, Weidong Yin, Lili Meng, and Leonid Sigal.
\newblock Layout2image: Image generation from layout.
\newblock {\em IJCV}, 2020.

\bibitem{Zhi2021ICCV}
Shuaifeng Zhi, Tristan Laidlow, Stefan Leutenegger, and Andrew~J Davison.
\newblock In-place scene labelling and understanding with implicit scene
  representation.
\newblock In {\em Proc. of the IEEE International Conf. on Computer Vision
  (ICCV)}, 2021.

\end{thebibliography}
